\pdfoutput=1
\documentclass{article}

\usepackage[table]{xcolor} % before icml2026 (which loads xcolor without options): \rowcolor in the tables
\usepackage{microtype}
\usepackage{graphicx}
\usepackage{placeins} % \FloatBarrier: appendix floats stay in their own section
\usepackage{booktabs}
\usepackage{array}
\usepackage{tabularx}
\usepackage{siunitx}  % S columns: numbers aligned on the decimal marker
\usepackage{multicol} % Appendices A and M: rendered episodes in columns
\usepackage{fvextra}  % Appendix M: prompt texts, long lines broken at spaces
\fvset{formatcom=\fontencoding{T1}\selectfont} % ASCII glyphs (~) in the prompt texts
\usepackage[skins,breakable]{tcolorbox} % Appendices A and M: boxes for prompts and model outputs

\usepackage{hyperref}

\usepackage[preprint]{icml2026}
\hypersetup{pdfsubject={arXiv preprint}} % after icml2026, which sets the subject to the ICML proceedings

\usepackage{amsmath}
\usepackage{amssymb}
\usepackage{mathtools}

\newif\ifshowpending \showpendingtrue
\usepackage{marginnote}
\usepackage[textsize=tiny]{todonotes}
\definecolor{rowgray}{rgb}{0.946,0.953,0.963}   % alternate rows and description bands
\definecolor{groupgray}{rgb}{0.835,0.858,0.895} % group rows
\definecolor{groupink}{rgb}{0.11,0.15,0.25}     % group labels
\definecolor{uncink}{gray}{0.36}                % uncertainties, intervals, n/a marks
\makeatletter
\newcommand{\uncsize}{\dimexpr\f@size pt-1pt\relax}
\newcommand{\uncolor}{.}
\newcommand{\uncfont}{\fontsize{\uncsize}{\f@baselineskip}\selectfont\color{\uncolor}}
\newcommand{\tablestyle}[1][1.18]{\footnotesize\setlength{\tabcolsep}{4pt}\renewcommand{\arraystretch}{#1}%
  \renewcommand{\uncsize}{\dimexpr\f@size pt-2pt\relax}\renewcommand{\uncolor}{uncink}}
\makeatother
\newcommand{\nicell}{\textcolor{uncink}{n.i.}}
\newcommand{\unc}[1]{\,{\uncfont(\ensuremath{\pm}#1)}}
\newcommand{\ci}[2]{\,{\uncfont[\num{#1}, \num{#2}]}}
\newcommand{\na}{\textcolor{uncink}{--}}
\makeatletter
\newcommand{\hd}[1]{#1}
\newcommand{\hdtwo}{\@ifstar{\hd@twoup}{\hd@two}}
\newcommand{\hd@two}[3][c]{\begingroup\edef\hd@ht{\the\ht\@arstrutbox}\renewcommand{\arraystretch}{0.95}%
  \begin{tabular}[b]{@{}#1@{}}\rule{\z@}{\hd@ht}\smash{#2}\\\smash{#3}\end{tabular}\endgroup}
\newcommand{\hd@twoup}[3][c]{\leavevmode\setbox\z@\hbox{\hd@two[#1]{#2}{#3}}\ht\z@\z@\box\z@}
\makeatother
\newcounter{tabcols}
\newcommand{\grouprow}[1]{\rowcolor{groupgray}\multicolumn{\value{tabcols}}{l}{\color{groupink}\bfseries #1}\\}
\newcommand{\groupsep}{\addlinespace[3pt]}
\newcommand{\entrysep}{\arrayrulecolor{black!45}\specialrule{0.3pt}{1.35pt}{1.35pt}\arrayrulecolor{black}}
\newlength{\entryA}\newlength{\entryB}\newlength{\entryW}
\newenvironment{entrytable}[1]{% #1: width of the name column, as a fraction of \columnwidth
  \setlength{\entryA}{#1\columnwidth}%
  \setlength{\entryW}{\dimexpr\columnwidth-2\tabcolsep\relax}%
  \setlength{\entryB}{\dimexpr\entryW-\entryA-2\tabcolsep\relax}%
  \begin{tabular}{>{\raggedright\arraybackslash}p{\entryA}>{\raggedright\arraybackslash}p{\entryB}}}
  {\end{tabular}}
\newcommand{\entryband}[1]{\rowcolor{rowgray}\multicolumn{2}{>{\raggedright\arraybackslash}p{\entryW}}{\rule{0pt}{2.45ex}#1\rule[-1.15ex]{0pt}{0pt}}\\}
\newcommand{\entrybandhead}[1]{\rowcolor{rowgray}\multicolumn{2}{l}{\itshape #1}\\}
\definecolor{clsnoise}{HTML}{697BC0}
\definecolor{clsexception}{HTML}{C295D8}
\definecolor{clsdrift}{HTML}{20A89B}
\definecolor{clsrule}{HTML}{DB7766}
\makeatletter
\newsavebox{\cls@box}
\newdimen\cls@em \newdimen\cls@pad \newdimen\cls@wd
\newcommand{\cls@chip}[3]{% #1 class, #2 text, #3: 1 with the tint, 0 the frame alone
  \leavevmode\begingroup
  \cls@em=\f@size\p@ \cls@pad=0.16\cls@em
  \sbox\cls@box{#2}\cls@wd=\wd\cls@box
  \hbox{\kern\cls@pad
    \rlap{\smash{\begin{pgfpicture}\pgfsetbaseline{\z@}%
      \pgfpathrectangle{\pgfpointorigin}{\pgfpointorigin}\pgfusepath{use as bounding box}%
      \pgfsetlinewidth{0.042\cls@em}\pgfsetcornersarced{\pgfqpoint{0.22\cls@em}{0.22\cls@em}}%
      \pgfsetstrokecolor{cls#1!92!black}%
      \pgfpathrectanglecorners{\pgfqpoint{\dimexpr0.021\cls@em-\cls@pad\relax}{-0.264\cls@em}}%
        {\pgfqpoint{\dimexpr\cls@wd+\cls@pad-0.021\cls@em\relax}{0.739\cls@em}}%
      \ifnum#3=\@ne \pgfsetfillcolor{cls#1!15}\pgfusepath{fill,stroke}\else\pgfusepath{stroke}\fi
    \end{pgfpicture}}}%
    \usebox\cls@box\kern\cls@pad}%
  \endgroup}
\DeclareRobustCommand{\cls}[2][]{\cls@chip{#2}{\if\relax\detokenize{#1}\relax#2\else#1\fi}{1}}
\DeclareRobustCommand{\clsring}[2][]{\cls@chip{#2}{\if\relax\detokenize{#1}\relax#2\else#1\fi}{0}}
\makeatother

\usepackage[varqu,varl]{zi4}
\fvset{breaksymbolleft={\color{black!40}\tiny\ensuremath{\hookrightarrow}},
  listparameters={\setlength{\topsep}{2pt}\setlength{\partopsep}{0pt}}}
\definecolor{promptbg}{HTML}{FCF9F4}
\definecolor{promptrule}{HTML}{C2AB86}
\definecolor{prompthead}{HTML}{F2E8D7}
\definecolor{promptink}{HTML}{6B5534}
\definecolor{outputbg}{HTML}{F0F4FA}
\definecolor{outputbar}{HTML}{4D5E99}
\definecolor{outputink}{HTML}{34406B}
\definecolor{outputrule}{HTML}{A9B2C1}
\newcommand{\boxlabel}[1]{\textls[70]{\textsc{#1}}}
\tcbset{quotedtext/.style={enhanced,breakable,center,width=0.94\linewidth,
  fontupper=\footnotesize,fonttitle=\footnotesize,coltitle=black!78,
  before upper={\parskip=0pt},before skip=10pt,after skip=10pt}}
\newtcolorbox{promptbox}[1][]{quotedtext,
  colback=promptbg,colframe=promptrule,boxrule=0.5pt,arc=1.5pt,left=8pt,right=8pt,top=5pt,bottom=5pt,
  colbacktitle=prompthead,toptitle=1.5pt,bottomtitle=1pt,lefttitle=8pt,titlerule=0.4pt,
  before title={{\color{promptink}\upshape\boxlabel{prompt}}\hspace{0.9em}\itshape},#1}
\newtcolorbox{outputbox}[1][]{quotedtext,
  colback=outputbg,frame hidden,sharp corners,boxrule=0pt,borderline west={2.5pt}{0pt}{outputbar},
  left=10pt,right=8pt,top=5pt,bottom=6pt,detach title,
  before upper={\tcbtitle\par\vspace{3pt}{\color{outputrule}\hrule height 0.7pt}\vspace{5pt}\parskip=0pt},
  before title={{\color{outputink}\upshape\boxlabel{model output}}\hspace{0.9em}\itshape},#1}

\icmltitlerunning{When Should an In-Context Learner Expand Its Hypothesis Space?}

\begin{document}

\twocolumn[
  \icmltitle{When Should an In-Context Learner Expand Its Hypothesis Space?}

  % Authors and affiliations are printed with [preprint] (as with [accepted]).
  \begin{icmlauthorlist}
    \icmlauthor{Weihan Li}{utokyo}
    \icmlauthor{Xinlei Chen}{hitsz}
    \icmlauthor{Junhao Wu}{rwth}
    \icmlauthor{Tianshi Zheng}{hkust}
  \end{icmlauthorlist}
  \icmlaffiliation{utokyo}{Department of Mechanical Engineering, The University of Tokyo, Tokyo, Japan}
  \icmlaffiliation{hitsz}{School of Mechanical Engineering and Automation, Harbin Institute of Technology, Shenzhen, China}
  \icmlaffiliation{rwth}{Faculty of Mechanical Engineering, RWTH Aachen University, Aachen, Germany}
  \icmlaffiliation{hkust}{Department of Computer Science and Engineering, HKUST, Hong Kong SAR, China}
  \icmlcorrespondingauthor{Weihan Li}{liweihan1107@g.ecc.u-tokyo.ac.jp}

  \icmlkeywords{in-context learning, hypothesis space, structural revision, Bayesian decision, language models}

  \vskip 0.3in
]

\printAffiliationsAndNotice{}

\begin{abstract}
Learning systems adapt quickly within a familiar family of models. The harder step is deciding, from a prediction failure, whether opening a richer family is worth its cost. We cast this as a costly sequential decision and introduce the Structural Revision Environment (SRE), which produces matched failures from noise, exceptions, change within the family or structure outside it, varies the price of expansion and the remaining horizon independently of the evidence, and admits an exact normative solution of the one-shot decision. The solution shows that revision is a value boundary, not an evidence threshold: except at the lowest prices, the richer family becomes the more probable explanation long before expanding becomes optimal. Small Transformers trained in SRE reproduce the expansion boundary from utility alone. Pretrained 7B and 8B language models fail at a definite point: their predictions carry a failure-sensitive signal that their decisions do not reflect. Told the gain of expanding, they weigh it against price and horizon only when they reason, and the one model that finishes its reasoning on the history alone still does not estimate that gain. Controlled post-training of the small Transformers moves the prior and the sharpness of predictions, and neither moves the criterion.
\end{abstract}

% ---- Main text (8-page limit) -----------------------------------------------
\section{Introduction}\label{sec:intro}

Learning systems adapt rapidly when new observations can be explained within a familiar family of models. A harder problem arises when the observations suggest that the family itself may be inadequate \citep{gelman2013philosophy,bobu2020quantifying}. Consider a learner that has seen many sequences generated by simple rules and now receives one in which a few values are off. The deviation might be noise, and the next observation of the same input will be normal. It might be a genuine quirk of one input, worth remembering and nothing more. Or it might be the first sign that this sequence follows a rule of a kind the learner is not currently considering. These possibilities call for qualitatively different responses: ignoring the deviation, recording an exception \citep{nosofsky1994rule}, or searching a richer family of rules. The last is by far the most expensive. It is also the only response that can exploit the missing structure. Yet the observations themselves do not immediately reveal which response is warranted.

\begin{figure*}[t]
  \centering
  \includegraphics[width=\textwidth]{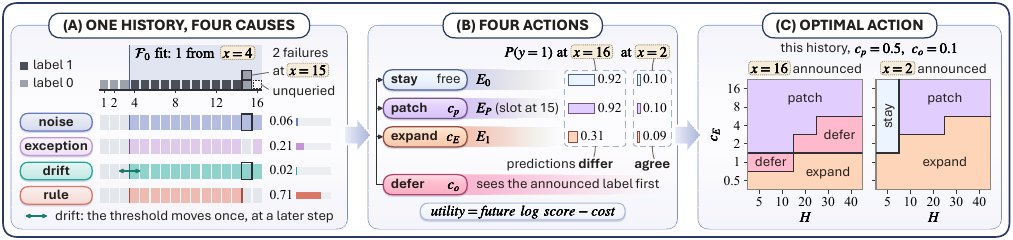}%
  \caption{\textbf{One history of SRE and its optimal actions.} (A) Sixteen observations fit a threshold at 4 except two labels at input 15, and input 16 is never queried. Each class explains them differently; the bars give the exact posterior of each class. (B) The actions and the predictors they select: $E_0$ holds the thresholds, $E_P$ adds a slot for input 15, and $E_1$ holds the intervals as well. At the announced query 16, $E_1$ parts from the other two, and at 2 all three agree. (C) The optimal one-shot action at each expansion price $c_E$ and remaining horizon $H$, with input 16 or input 2 announced as the next query.}
  \label{fig:overview}
\end{figure*}

A large theoretical literature studies in-context learning by asking how a learner infers a latent task, parameter, or algorithm from the examples available in its context. Transformers trained on mixtures of task families approximate Bayesian inference under their training distribution \citep{xie2022explanation,raventos2023pretraining,panwar2024incontext,wurgaft2025incontext}, average over or select among the families they know \citep{bai2023transformers,dudley2026incontext}, and language models have been tested against Bayesian references \citep{falck2024incontext,gupta2025enough} and on keeping enumerated hypotheses consistent with the evidence \citep{xu2026when}. These accounts characterize inference once the relevant alternatives are available to the learner's current computation. Our question begins one step earlier: \textbf{when making a richer alternative available is itself costly, when is that computation worth invoking?}

Prediction failure alone does not determine whether richer inference is warranted. A single extreme observation may be highly surprising while providing little evidence of missing structure \citep{liakoni2021learning}, whereas small deviations that recur systematically across related inputs can accumulate strong evidence that the current family is inadequate \citep{page1954continuous}. Even strong evidence does not determine the action. Opening a richer family incurs computation \citep{dasgupta2017where} whose benefit must be recovered through future predictions \citep{russell1991principles,poh1993reasoning}, so identical evidence can justify different decisions when the cost of computation or the remaining horizon changes. Evidence also leaves open whether a local repair suffices. An exception list can absorb any number of deviations, but its length grows with every new anomalous input, while a rule that explains them all has constant length and predicts the anomalies still to come \citep{grunwald2007minimum}. The problem is therefore a sequential decision over when richer inference is worth invoking, and over which kind of revision the evidence supports.

To separate structural evidence from the value of expansion, we introduce the \textbf{Structural Revision Environment (SRE)}, a controlled generative environment in which matched prediction failures arise from noise, local exceptions, within-family change, or structure outside the current family. The environment varies these sources of failure independently of the cost and future value of richer inference, and it admits exact Bayesian calculations and an exact normative solution of the one-shot revision decision. Figure~\ref{fig:overview} follows one of its histories.

The solution of SRE specifies when revision is worth its cost, and it is the reference against which we evaluate small Transformers trained from scratch and pretrained language models on three questions: whether they revise in response to evidence of structure or to surprise, whether their decisions follow the value of expansion, and whether the alternatives that eventually dominate were already influencing predictions before revision.

The solution gives a first result on its own: \textbf{revision is a value boundary and not an evidence threshold.} One history has different optimal actions under different prices, horizons and announced queries; above the lowest prices, the richer family becomes the more probable explanation long before expansion becomes optimal; and the boundary between local repair and expansion is set by the inputs a rule would predict and a repair cannot cover, so that the number of anomalies moves the boundary only through the region they imply.

Small Transformers trained in the environment reproduce the expansion boundary from utility alone, without action labels and without the posterior, so it is learnable without exact inference. Pretrained 7B and 8B language models break the chain at a definite point. Their predictions carry a failure-sensitive signal, with the shape of a learner that explains the failures inside its current family, and their decisions do not reflect it: with the history alone the decision is constant, at states where the reference expands and the models' own predictions have already moved. With the gain of expanding stated, they respond to it as a trigger responds to a statistic. In the three models tested with longer reasoning, the stated gain is weighed against the price and the horizon in the reference's proportions. In the one that finishes its reasoning on the history alone, the history still does not yield an estimate of what expansion would buy. The failure thus has two parts, \textbf{forming an estimate of the gain} and \textbf{pricing it}, and what distinguishes the models is the second.

Controlled post-training of the small Transformers separates three accounts of the late entry of a richer family after training: the data move the implied prior, the objective moves the sharpness of the predictions, and neither moves the decision criterion; the lost response remains decodable and is largely recoverable.

% Section 4 (Readouts without actions).
\section{The Revision Problem}\label{sec:problem}

\textbf{Setting.} A learner observes a stream of pairs $z_t=(x_t,y_t)$. Before each label is revealed it outputs a predictive distribution $q_t(\cdot\mid x_t,z_{<t})$ and receives the log score $\log q_t(y_t)$. The next query $x_{t+1}$ is announced one step ahead. A family is a set of rules, and a predictor is the Bayes predictive over a family: $E_0$ predicts with the current family $\mathcal F_0$ alone, and $E_1$ with a model that holds a richer family $\mathcal F_1\supset\mathcal F_0$ as well. Running $E_1$ costs computation. At every step the learner knows the remaining horizon $H$ and the prices of the structural actions defined below. Its utility over an episode of length $T$ is
\[
\mathcal U=\sum_{t\le T}\log q_t(y_t)\;-\;c_p\,n_{\mathrm{patch}}\;-\;c_E\,\mathbf 1[\mathrm{expanded}].
\]

\textbf{From failure to revision.} A prediction failure is the first link of a chain, and the revision decision depends on keeping the links apart. Whether the family is inadequate is a latent property of the process, its misspecification (Appendix~\ref{app:sre}), fixed within an episode and never observed. What the learner observes is \textbf{surprise}; what it can extract from the history is \textbf{evidence} that the family is inadequate; what it must estimate before acting is the \textbf{value} of acting on that evidence.

% Section 2: the three quantities of the chain from failure to revision. Entry table of main.tex
\begin{table}[!htb]
\centering\footnotesize\setlength{\tabcolsep}{3.5pt}\renewcommand{\arraystretch}{1.1}%
\begin{entrytable}{0.375}
\toprule
quantity & definition \\
\entrybandhead{when it grows}
\midrule
\textbf{Shannon surprise} $S_t$ & $-\log q_t(y_t\mid x_t)$ \\
\entryband{On any improbable label, even under a correct family}
\entrysep
\textbf{Structural evidence} $\mathcal E_t$ & $\sum_{s\le t}\log U(y_s\mid x_s,z_{<s})$\newline${}-\sup_{h\in\mathcal F_0}\log p_h(z_{\le t})$ \\
\entryband{When the residuals of $\mathcal F_0$ are compressible}
\entrysep
\textbf{Value of expansion}\newline$\Delta Q^{\ast}(s_t)$ & $Q^{\ast}(s_t,\mathrm{expand})$\newline${}-\max_{a\ne\mathrm{expand}}Q^{\ast}(s_t,a)$ \\
\entryband{With the horizon and the loss under $\mathcal F_0$; it falls with the price}
\bottomrule
\end{entrytable}
\end{table}

Here $U$ is a catch-all predictor that commits to no family: it keeps a smoothed count of the labels seen at each input. When $\mathcal F_0$ is correct, $e^{\mathcal E_t}$ is dominated by a nonnegative martingale, so $\Pr(\exists t: \mathcal E_t\ge\log\tfrac1\alpha)\le\alpha$ \citep{ville1939etude,ramdas2023gametheoretic}; when it is misspecified, $\mathcal E_t$ grows linearly.

\textbf{Actions.} Before each prediction the learner may change what it predicts with, by one of four actions. In the sequential protocol the learner acts at every step of an episode, and in the one-shot protocol it makes a single choice after a fixed history. Waiting has value only through what the next observation can change \citep{dixit1994investment}, and a state in which waiting is worth more than committing to any action now and than never revising is a defer state (Appendix~\ref{app:normative}).

\begin{table}[!htb]
\centering\footnotesize\setlength{\tabcolsep}{3.5pt}\renewcommand{\arraystretch}{1.1}%
\begin{entrytable}{0.175}
\toprule
action & operation and price \\
\entrybandhead{in the two protocols}
\midrule
\multicolumn{2}{c}{\raisebox{-3.5pt}{\includegraphics[width=\entryW]{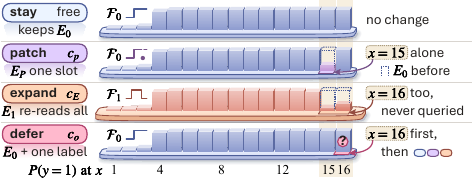}}}\\
\multicolumn{2}{l}{\rule{0pt}{2.6ex}\scriptsize\itshape Bars: the predictive after each action, on the history of Figure~\ref{fig:overview}.}\\
\midrule
\textbf{stay} & Keep $E_0$, whose change-point prior lets the rule in $\mathcal F_0$ move once. Free \\
\entryband{Sequential: taken when no revision is worth more than waiting.\newline One-shot: a commitment to $E_0$ for the remaining rounds}
\entrysep
\textbf{patch} & Give one input a slot that predicts its label from the labels observed there alone ($E_P$). $c_p$ per slot \\
\entryband{The same in both protocols}
\entrysep
\textbf{expand} & Switch to $E_1$, which holds $\mathcal F_1$ and re-reads the whole history. $c_E$ once, not reversed \\
\entryband{Sequential: when to switch is an optimal stopping problem.\newline One-shot: a commitment to $E_1$ for the remaining rounds}
\entrysep
\textbf{defer} & Pay $c_o$ to see the label of the announced query, then stay, patch or expand \\
\entryband{One-shot only. In the sequential protocol waiting is a stay whose value lies in the next observation}
\bottomrule
\end{entrytable}
\end{table}

\textbf{The normative problem.} The information state $s_t$ contains the history, the prices, the remaining horizon and any announced queries. The optimal values satisfy
\begin{gather*}
Q^{\ast}(s_t,a)=-C(a)+\mathbb E\big[\ell_{t+1}+V^{\ast}(s_{t+1})\mid s_t,a\big],\\
V^{\ast}(s)=\max_a Q^{\ast}(s,a),
\end{gather*}
with $\ell_{t+1}$ the next log score and $C(a)$ the price of $a$. The value of expansion is the expansion advantage $\Delta Q^{\ast}(s)=Q^{\ast}(s,\mathrm{expand})-\max_{a\ne\mathrm{expand}}Q^{\ast}(s,a)$, and expansion is optimal exactly when $\Delta Q^{\ast}(s)>0$. A one-step comparison gives the shape of the boundary. Let $p_t$ be the posterior probability that the richer family is needed, $\Delta$ the per-step loss of staying when it is needed, and $\delta$ the per-step loss of expanding when it is not. Expanding now rather than never is worth $H\{p_t\Delta-(1-p_t)\delta\}-c_E$, which is positive when $p_t$ exceeds a threshold that rises with $c_E/H$, the amortized price (Appendix~\ref{app:normative}). \textbf{The natural coordinates of the problem are therefore the belief in the richer family and the amortized price.}

\section{Structural Revision Environment}\label{sec:sre}

\textbf{Generative process.} Inputs are 16 ordered positions and labels are binary. In every episode each label is flipped independently with probability $\varepsilon=0.08$. The current family $\mathcal F_0$ contains the threshold rules $y=\mathbf 1[x\ge b]$. The richer family $\mathcal F_1$ contains the interval rules $y=\mathbf 1[l\le x\le u]$. Each episode belongs to one of four classes, and the class is never shown to the learner.

\begin{table}[!htb]
\centering\footnotesize\setlength{\tabcolsep}{3.5pt}\renewcommand{\arraystretch}{1.1}%
\setlength{\entryA}{0.18\columnwidth}%
\setlength{\entryW}{\dimexpr\columnwidth-2\tabcolsep\relax}%
\setlength{\entryB}{\dimexpr\entryW-\entryA-0.95in-4\tabcolsep\relax}%
\renewcommand{\entryband}[1]{\rowcolor{rowgray}\multicolumn{3}{>{\raggedright\arraybackslash}p{\entryW}}{\rule{0pt}{2.45ex}#1\rule[-1.15ex]{0pt}{0pt}}\\}%
\renewcommand{\entrybandhead}[2]{\rowcolor{rowgray}\multicolumn{3}{l}{\makebox[\dimexpr\entryA+2\tabcolsep+0.95in\relax][l]{\itshape #1\hfill\upshape\scriptsize\color{uncink}#2}}\\}%
\newcommand{\clsglyph}[1]{\raisebox{\dimexpr0.083in-\height\relax}[0.083in][\dimexpr\height-0.083in+2.5pt\relax]{\includegraphics{figures/cls_#1.pdf}}}%
% \isoflags{class}{0/1}{word}{0/1}{word}: two capsules of fixed widths, drawn behind their text like the chips of \cls
% (they take no vertical space): filled in the class tint with a tick for yes, white and grey with a cross for no.
\makeatletter
\newsavebox{\iso@box}\newdimen\iso@em\newdimen\iso@pad\newdimen\iso@wa\newdimen\iso@wb
\newcommand{\iso@font}{\fontsize{7.5}{9}\selectfont}%
\newcommand{\iso@cap}[4]{\begingroup\iso@em=\f@size\p@\iso@pad=0.36\iso@em
  \ifnum#3=\@ne\def\iso@mark{\checkmark}\def\iso@ink{black}\else\def\iso@mark{\times}\def\iso@ink{black!58}\fi
  \sbox\iso@box{\iso@font\color{\iso@ink}\makebox[#2]{$\iso@mark$\,#4}}%
  \hbox{\kern\iso@pad\rlap{\smash{\begin{pgfpicture}\pgfsetbaseline{\z@}%
    \pgfpathrectangle{\pgfpointorigin}{\pgfpointorigin}\pgfusepath{use as bounding box}%
    \pgfsetlinewidth{0.042\iso@em}\pgfsetcornersarced{\pgfqpoint{0.47\iso@em}{0.47\iso@em}}%
    \ifnum#3=\@ne\pgfsetstrokecolor{cls#1!92!black}\pgfsetfillcolor{cls#1!15}\else\pgfsetstrokecolor{black!30}\pgfsetfillcolor{white}\fi
    \pgfpathrectanglecorners{\pgfqpoint{-\iso@pad}{-0.21\iso@em}}{\pgfqpoint{\dimexpr\wd\iso@box+\iso@pad\relax}{0.73\iso@em}}%
    \pgfusepath{fill,stroke}\end{pgfpicture}}}\usebox\iso@box\kern\iso@pad}\endgroup}%
\settowidth{\iso@wa}{\iso@font$\checkmark$\,reproduces}\settowidth{\iso@wb}{\iso@font$\checkmark$\,spreads}%
\newcommand{\isoflags}[5]{\leavevmode\iso@cap{#1}{\iso@wa}{#2}{#3}\kern0.25em\iso@cap{#1}{\iso@wb}{#4}{#5}\kern0.5em\ignorespaces}%
\makeatother
\begin{tabular}{>{\raggedright\arraybackslash}p{\entryA}>{\raggedright\arraybackslash}p{0.95in}>{\raggedright\arraybackslash}p{\entryB}}
\toprule
class & label pattern & mechanism \\
\entrybandhead{what it isolates}{input\,$\rightarrow$}
\midrule
\textbf{\cls{noise}} & \clsglyph{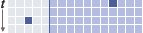} & \emph{A threshold rule. Failures come from label flips alone} \\
\entryband{\isoflags{noise}{0}{reproduces}{0}{spreads}Any revision is wasted}
\entrysep
\textbf{\cls{exception}} & \clsglyph{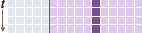} & \emph{A threshold rule, inverted at one or two inputs} \\
\entryband{\isoflags{exception}{1}{reproduces}{0}{spreads}One slot per input suffices, and expansion buys nothing more}
\entrysep
\textbf{\cls{drift}} & \clsglyph{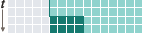} & \emph{The threshold moves once during the episode} \\
\entryband{\isoflags{drift}{1}{persists}{1}{spreads}The process stays inside $\mathcal F_0$}
\entrysep
\textbf{\cls{rule}} & \clsglyph{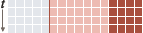} & \emph{An interval rule} \\
\entryband{\isoflags{rule}{1}{reproduces}{1}{spreads}Every input above $u$ conflicts with every threshold, including inputs not yet queried}
\bottomrule
\end{tabular}
\end{table}

An interval $[l,u]$ agrees with the threshold at $l$ up to $u$ and differs from it at every input above $u$. Seen from $\mathcal F_0$, a \cls{rule} episode is therefore an \cls{exception} episode whose exceptions fill the tail above $u$ and share one description.

\textbf{Matched failures.} At each step the query repeats an earlier input with probability $r=0.35$ and is a new input otherwise. The schedule is drawn independently of the class and the four classes share one noise process, so the same flips fall on the same queries under each of them.

\textbf{Predictors and protocols.} $E_0$ is the Bayes predictive over the threshold rules under the generative change-point prior \citep{adams2007bayesian}, so a \cls{drift} episode stays inside the family. $E_P$ adds a Beta-Bernoulli slot for each patched input. $E_1$ is the Bayes predictive under the full generative process, every class and every configuration. The prices, the horizon and the announced query vary independently of the history, and structural actions leave the queries and labels that follow unchanged. A sequential episode lasts 40 steps, and in the one-shot protocol the value of every action is computed from the exact posterior (Appendix~\ref{app:reference}).

\textbf{Reference policies.} The latent state takes 7,352 values, and the posterior is enumerated exactly. The reference policy maximizes expected utility under this posterior. It holds the same information as the learner and does not know the class of the episode. In the sequential protocol it is a two-step lookahead (Appendix~\ref{app:reference}). A second reference, the bounded reference policy, is restricted to count statistics that need no $\mathcal F_1$, such as the numbers of reproducible and non-reproducible anomalies, and it measures how much of the expansion decision can be made before expanding. Two triggers serve as baselines: the e-process trigger \citep{shin2024edetectors} expands when the structural evidence $\mathcal E_t$ crosses a threshold, fixed once or tuned in each cell of the price grid, and the surprise trigger uses $S_t$ instead.

\textbf{One history, several optimal actions.} Figure~\ref{fig:overview} follows one history. A threshold at 4 fits every observation except two labels at input 15, and input 16 is never queried. The posterior gives the \cls{rule} class 0.71 and the \cls{exception} class 0.21, and the two explanations part at input 16, where an interval ending at 14 predicts one label and an exception at 15 the other (Figure~\ref{fig:overview}B). Deferring buys the announced label before the choice. It is worth the improvement from choosing after seeing that label \citep{howard1966information}, which is zero when every explanation predicts the label alike (Appendix~\ref{app:identifiability}). At $c_E=4$ and $H=30$, with input 16 announced, waiting is optimal and worth 0.33 nats over the best immediate commitment; with input 2 announced, the value of waiting is exactly zero and the optimal action is to expand. With input 16 announced, raising $c_E$ from 0.5 to 16 moves the optimal action from expand through defer to patch, and raising $H$ from 5 to 40 moves it from patch to defer (Figure~\ref{fig:overview}C). A policy that uses only the history returns one action in all of these cases.

\textbf{The decision boundary.} Solved over histories with up to five anomalous inputs, SRE gives a phase diagram over the reference's belief in the richer family, its posterior log-odds of the \cls{rule} class, and the amortized price (Figure~\ref{fig:phase}, Appendix~\ref{app:same-history}). \textbf{Belief and decision come apart.} Measured in the number $m$ of distinct inputs with reproducible anomalies, the rule becomes more probable than an exception at $m^{\ast}_B=0.5$, so one reproducible anomaly at the top end already suffices. Above the lowest prices, expansion asks for more. A slot covers one input, and the inferred rule covers its region, every input above its upper end $u$, queried or not. Slots and the rule fit the observed anomalies equally well, so what expanding gains over patching comes from the rest of the region. Let $n$ be the number of inputs that the rule predicts and no slot covers, and $g$ the expected loss per step that each of them costs the patched predictor. Expanding beats patching when $n\,g\,H>c_E-c_p$, so the boundary lies at a fixed amortized price difference $(c_E-c_p)/H$ (Appendix~\ref{app:normative}). When the announced query is uninformative, expansion is optimal up to a difference of 0.20 with two distinct anomalous inputs and 0.56 with five, and the boundaries for $H=10,20,30$ agree to within a few hundredths. The number of anomalous inputs enters only through the region it implies: with two anomalous inputs, each observed twice, at the two ends of a region and the inputs between them never queried, the boundary rises from 0.14 for a region of two inputs to 0.68 for a region of five. Across SRE and two further description languages, this comparison predicts the boundary with a median ratio of 0.98 for two or more anomalous inputs, and where a rule's type fixes its region the boundary meets a ceiling (Appendix~\ref{app:languages}).

\section{Learners and Measurements}\label{sec:learners}

\begin{figure*}[t]
  \centering
  \includegraphics[width=\textwidth]{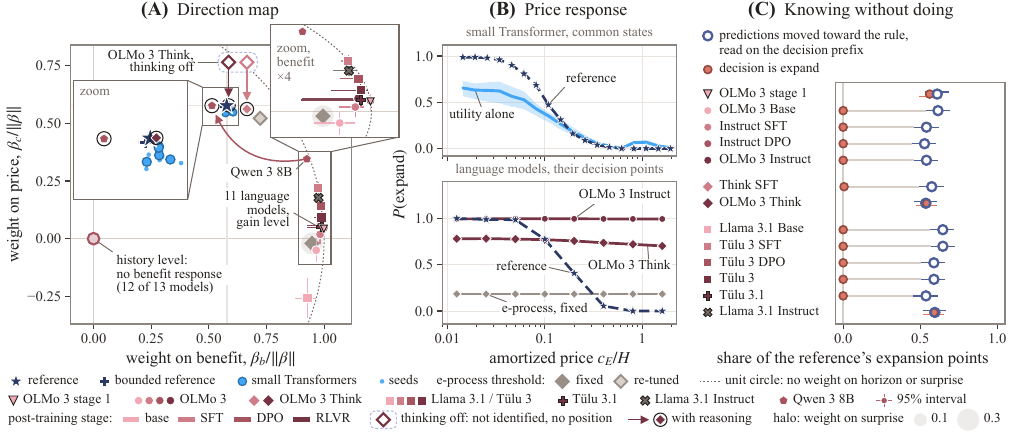}
  \caption{\textbf{Small Transformers weigh the price of expanding as the reference does; 7B and 8B language models value a stated gain only when they reason.} (A) The small Transformers (three seeds per training regime) and the bounded reference lie on the reference (left zoom); at the gain level, eleven language models lie in the fixed-threshold corner with the e-process trigger (right zoom); with reasoning, three models move onto the reference's direction while their slopes grow by at least an order of magnitude. (B) At rule log-odds 0 to 2, the small Transformer trained on utility alone (mean over three seeds, band the range) expands less as the amortized price rises, as the reference does, and the fixed trigger and, with the gain stated, OLMo 3 Instruct and OLMo 3 Think are nearly flat. (C) At the decision points where the reference expands, the share at which each language model's label predictions have moved toward the rule, and the share at which its decision is expand, with 95\% intervals.}
  \label{fig:model-map}
\end{figure*}

\textbf{Small Transformers.} They are meta-trained on episodes drawn from SRE, so their training distribution is known exactly. An \emph{implicit} learner predicts the next label and takes no structural action. We train it at three base rates of \cls{rule} episodes: the generative 0.25, a rare 0.02, which shows how much lateness a small prior alone produces, and 0, so that it never sees a rule. An \emph{explicit} learner chooses a structural action at every step, and the exact predictors of Section~\ref{sec:sre} produce the predictions. Its errors are therefore errors of decision. Its input is the observations, the announced query, the prices and the remaining horizon. We train it either to imitate the reference policy or to maximize utility directly, which the exogeneity of the observations makes exact. A diagnostic variant also receives the posterior of the reference policy and has only the decision left to learn.
\textbf{Language models.} We use three lineages of 7 or 8 billion parameters that release a checkpoint after every stage of post-training: OLMo 3 Instruct and OLMo 3 Think \citep{olmo2025olmo3}, which share a base model and apply supervised fine-tuning (SFT), preference tuning (DPO; \citealp{rafailov2023direct}) and reinforcement learning with verifiable rewards (RLVR), and T{\"u}lu 3 \citep{lambert2024tulu3}, which applies the same sequence of stages to Llama 3.1 \citep{grattafiori2024llama3}. Their fourteen checkpoints are read in the implicit protocol and thirteen of them in the explicit protocol (Table~\ref{tab:llms}), and Qwen 3 8B \citep{yang2025qwen3} is added in the conditions with reasoning (Appendix~\ref{app:reasoning}). In the explicit protocol, which is one-shot, the probabilities of the four actions are read from the log-probabilities of their continuations. The prompt is given at three levels of input, which remove the links of the chain of Section~\ref{sec:problem} one at a time. In the implicit protocol the models predict the labels of the rendered episode in three contexts, alone or after eight demonstration episodes drawn from threshold rules or from the generative mixture, and their predictions are read as those of the small Transformers (Appendix~\ref{app:prompts}).

\begin{table}[!htb]
\centering\footnotesize\setlength{\tabcolsep}{3.5pt}\renewcommand{\arraystretch}{1.1}%
\begin{entrytable}{0.15}
\toprule
level & prompt \\
\entrybandhead{left to the model}
\midrule
\textbf{history} & The observations, the prices, the rounds left and the announced query \\
\entryband{Find the failures, estimate what expanding would gain, and price the gain against $c_E$ and the rounds left}
\entrysep
\textbf{failure} & The same, with a summary of the observations $E_0$ fails on and of the inputs at which a failure repeats \\
\entryband{Estimate the gain and price it}
\entrysep
\textbf{gain} & The same, with the reference's expected gain per round from expanding now \\
\entryband{Price the gain}
\bottomrule
\end{entrytable}
\end{table}

\textbf{Decision regression.} Every learner and baseline is read on one set of states, every prefix of the evaluation episodes crossed with every price cell and episode length, by a logistic regression of its expansion decision on the reference's own quantities, with linear predictor $\beta_b\log b+\beta_H\log H-\beta_c\log(c_E+O)+\gamma S_{t-1}$, where $b$ is the benefit of expanding per remaining round, which the gain level states, $H$ the horizon, $c_E+O$ the effective price, $c_E$ plus the value $O$ of the best alternative to expanding, and $S_{t-1}$ the surprise of the last observation (Appendix~\ref{app:statistics}). The reference expands when $bH>c_E+O$, so for it $\beta_b=\beta_H=\beta_c$ and $\gamma=0$. We report each learner by the direction of its coefficient vector, the weight on the benefit $\beta_b/\|\beta\|$ against the weight on the price $\beta_c/\|\beta\|$, with the weight on surprise as a halo (Figure~\ref{fig:model-map}A): a learner that weighs the benefit against the price as the reference does lies on the diagonal, a trigger blind to the price on the horizontal axis, and a decision that responds to neither at the origin. Where $\beta_b$ is identified (Appendix~\ref{app:reasoning}) we also report $\beta_c/\beta_b$ and $\beta_H/\beta_b$, its price sensitivity and horizon sensitivity, which are 1 for the reference, and $\gamma/\beta_b$, its surprise leakage, which is 0. The identification rests on histories built in pairs (Appendix~\ref{app:same-history}).

\textbf{Readouts without actions.} The \emph{matched-pair $\kappa$} of a learner is the share of the Bayes shift that its predictions make between two histories $D$ and $D'$ that have the same likelihood under every hypothesis in $\mathcal F_0$ and differ in their evidence for the richer family, read at a probe input $x^{\ast}$ on which the two families disagree, with $y^{\ast}$ the label the richer family predicts there:
\[
\kappa=\frac{\operatorname{logit}q(y^{\ast}\mid x^{\ast},D)-\operatorname{logit}q(y^{\ast}\mid x^{\ast},D')}{\operatorname{logit}p^{\mathrm B}(y^{\ast}\mid x^{\ast},D)-\operatorname{logit}p^{\mathrm B}(y^{\ast}\mid x^{\ast},D')},
\]
where $p^{\mathrm B}$ is the Bayes predictive under the generative prior. In $D$ the $m$ anomalous inputs lie above $u$, where one interval explains them, and in $D'$ they lie where no interval does (Appendix~\ref{app:sre}). A Bayes predictor that gives the richer family a smaller prior $\pi$ has a curve $\kappa(m;\pi)$ of its own, with $m$ as in Section~\ref{sec:sre}; it starts below 1 and rises as evidence accumulates, so a learner's curve is read against this family of curves. A learner on one of them averages under that prior, and a learner flat and below every curve gives structural evidence less weight than any Bayes predictor does. $\kappa$ gives the size of the response to structural evidence, and a second readout gives its shape: a Bayes predictor confines the response to the inputs above $u$, with a step at $u$ that a small prior lowers and does not remove. On a \cls{rule} episode a learner \emph{lifts} at the first step at which its mean predicted probability of label 1 over the unqueried inputs above $u$ falls below one half.

\section{Results}\label{sec:results}

\subsection{Small Transformers reproduce the expansion boundary}\label{sec:value}

Explicit small Transformers lose 0.04 to 0.09 nats per episode to the reference policy over the grid of prices and horizons. The bounded reference loses 0.24, the e-process trigger tuned anew in every cell 0.42, and the same trigger with one threshold 1.19 (Appendix~\ref{app:table1}, Table~\ref{tab:regret}). The bounded reference has price sensitivity 1.00 and horizon sensitivity 1.03: count statistics suffice to follow the value of expansion, and what it loses against the reference, it loses in evidence. Within the range of prices seen in training, price sensitivity is 0.91 to 0.96, horizon sensitivity 0.95 to 1.07 and surprise leakage at most 0.02 (Figure~\ref{fig:model-map}).
\emph{Utility suffices.} The learner trained on utility alone never sees an action of the reference policy and comes closer to the reference than any trigger. Giving a learner the posterior of the reference changes neither its loss nor its sensitivities, so for these learners forming the evidence is not what limits the decision (Appendix~\ref{app:learned-policies}). Patching and waiting are learned less well: under utility alone, 0.09 of \cls{exception} episodes are patched without expanding, against 0.41 for the reference.

\begin{figure*}[t]
  \centering
  \includegraphics[width=\textwidth]{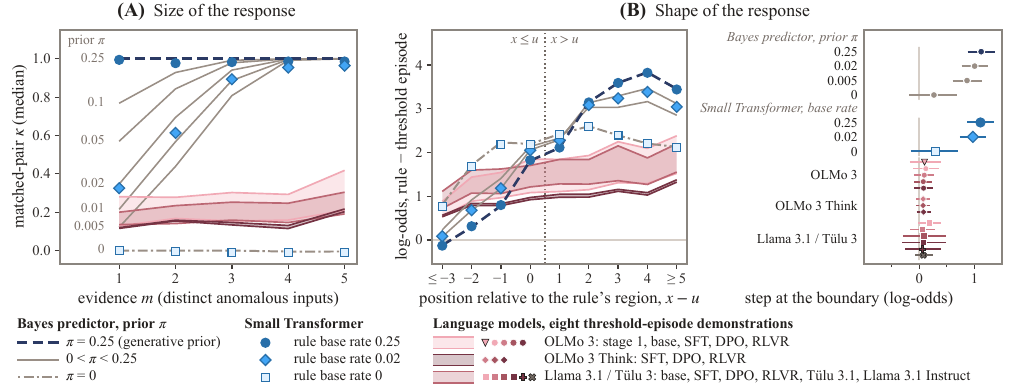}%
  \caption{\textbf{What responds to a failure.} (A) Matched-pair $\kappa$, the share of the Bayes shift toward the interval rule that a learner's predictions make between the two histories of a pair (Section~\ref{sec:learners}), against the number of distinct anomalous inputs. Bands: the 14 language models of three lineages, each band the range within a lineage. (B) The shift in log-odds between a \cls{rule} episode and its matched threshold episode at inputs placed by their distance from the edge of the rule's region, on 110 such pairs of episodes. Right: the step at $u$, the coefficient of $\mathbf 1[x>u]$ in the regression of Appendix~\ref{app:active-space}, with 95\% bootstrap intervals over episodes.}
  \label{fig:active-space}
\end{figure*}
\subsection{Language models decide without the evidence}\label{sec:llm-decision}

The three levels locate where the chain breaks. At the history and failure levels the models do not turn the failures into an estimate of what expanding would gain, and at the gain level they do not price the estimate unless they reason.

\emph{At the history level, the decision is constant.} In the four stages of OLMo 3 Instruct the probability of expanding differs between the four classes by at most 0.01, it is the same on a \cls{rule} history and on its matched threshold history (same queries and flips, a threshold at the rule's lower end), and the coefficient on the benefit of expanding is zero; the probability of deferring moves by at most 0.03 when the announced query separates the competing explanations, against 0.29 for the reference. The other lineages and stages behave in the same way, and the null response is robust across renderings and action orders, with small exceptions detailed in Appendix~\ref{app:prompts}. Marking the failures in the prompt, the failure level, leaves the decision constant as well: the response to the benefit stays below 0.06 wherever it is identified. Stating the generative process in full, the four classes with their base rates and the families the two predictors assume, raises the probability of expanding in most models and produces no positive response to a \cls{rule} history over its matched threshold history (Appendix~\ref{app:controls}).

\emph{A failure-sensitive signal is present in the predictions, and the decision does not reflect it.} Read on the same decision-context prefix, with the decision cue replaced by a label question, the predictions have moved toward the labels the interval rule predicts at 0.53 to 0.65 of the points where the reference expands, in every model, while the decision on that prefix is expand at none of them in ten of the thirteen models and at a share that does not follow the prediction in the other three (Figure~\ref{fig:model-map}C, Appendix~\ref{app:controls}).

\emph{With the gain stated, models respond to it alone; with room to reason, they weigh it.} When the expected gain per round from expanding now is written into the prompt, the probability of expanding rises with it in every model. In eleven of them it depends on neither the price nor the rounds left: price sensitivity lies between $-0.28$ and 0.22 and horizon sensitivity between $-0.29$ and 0.13. On the map of Figure~\ref{fig:model-map} these models fall in the corner of the triggers with a fixed threshold: a response to one number, without valuation. The two models of the OLMo 3 Think lineage expand on nearly every point, with responses to the gain, the price and the rounds left of 0.05 to 0.13 each, too small for their proportion to be identified (Appendix~\ref{app:reasoning}). At 32 billion parameters the history-level decision still does not separate the classes, and the stated gain is weighed in part without reasoning, with responses far smaller than the reference's (Appendix~\ref{app:reasoning-32b}). Three models were allowed to reason before answering, Qwen 3 8B until it finished and the two OLMo 3 Think models for 8,192 tokens. All three then weigh the stated gain against the price and the rounds left in the reference's proportions, with price sensitivities of 1.12, 0.99 and 0.85 and horizon sensitivities of 1.24, 0.97 and 0.74, and their responses grow by at least an order of magnitude. At the history level, where only Qwen 3 8B finishes its reasoning within the cap, reasoning moves probability on \cls{rule} histories toward patching and deferring, raises the probability of expanding over the matched threshold history by at most 0.06, and does not make it follow the price: \textbf{the history is not turned into an estimate of what expansion would buy} (Appendix~\ref{app:reasoning}).

\subsection{What responds to a failure}\label{sec:respond}

\emph{Language models respond with a fraction of the Bayes shift at every level of evidence.} In all 14 models of the three lineages (Figure~\ref{fig:active-space}A), $\kappa$ with three anomalous inputs lies between 0.09 and 0.43 across the three contexts, where every Bayes predictor with a prior of at least 0.005 is above 0.80, and none of the 126 curves, one per model, context and rendering, follows a member of the family indexed by the prior, so the weak response is not the response of a learner that holds the richer family with a small prior.

\emph{The response has the shape of a learner without the richer family.} Two shapes are possible at the edge of the rule's region: a step at $u$, or a response that decays with the distance from the observed anomalies and crosses $u$ without a step. A Bayes predictor that holds the richer family confines its response to the inputs that the rule covers: at the upper end of the interval its log-odds step by about one unit for any prior from 0.005 to 0.25, and at inputs well below $u$ the \cls{rule} episode and its matched threshold episode are predicted alike. The small Transformers step like the Bayes predictors of their priors. A Bayes predictor with a prior of zero also responds to the anomalies, and strongly. It explains them inside the current family, by a threshold that has moved, so its response has no step, crosses the edge of the region and persists well below $u$. The learner that never saw a rule does the same. The language models show no step, with every interval including zero, and they shift their predictions at inputs well below $u$ by 0.5 to 2.6 in log-odds across the three contexts (Figure~\ref{fig:active-space}B, Appendix~\ref{app:active-space}): \textbf{they respond to the failures and not to the rule that would explain them.} They differ from the predictor with a prior of zero in one respect. That predictor responds to the presence of anomalies and not to their arrangement, so its $\kappa$ is zero, and theirs is positive. The surprise leakage of their prediction updates (Appendix~\ref{app:active-space}) is 0.10 to 0.19, against at most 0.02 for any Bayes predictor with a misspecified prior.

\emph{Stages.} No stage of post-training changes the shape of the response, and reinforcement learning with verifiable rewards changes none of the predictive readouts (Appendix~\ref{app:posttraining}).

\subsection{Post-training moves the prior and the sharpness, and neither moves the criterion}\label{sec:posttrain}

We post-trained the implicit small Transformers of the generative base rate for 20,000 training steps in a two-by-two design. The data either kept \cls{rule} episodes at the generative base rate or removed them, and the objective was either the log score or the expected 0/1 reward, which pays the probability placed on the observed label and is maximized by predictions at the extremes. Readouts are taken against the Bayes predictor of the prior the learner holds after post-training, with $\kappa$ computed so that a sharpening of the predictions cannot enter it (Appendix~\ref{app:posttraining}).

\emph{The data set the prior.} Removing \cls{rule} episodes lowers the implied base rate of the richer family from 0.25 to $10^{-7.4}$, nearly seven decades in 20,000 steps, and the objective leaves it unchanged (Appendix~\ref{app:posttraining}). The learner trained without rules lifts seven episode steps later than the Bayes predictor of the generative prior and no later than that of the prior it now holds, and its $\kappa$ stays on that predictor's curve, within 0.03 at one, three and five anomalous inputs. \textbf{Its lateness is a prior.}

\emph{The objective sets the sharpness.} The reward objective multiplies the scale of the log-odds by eight to ten and leaves the base rate where it was. With \cls{rule} episodes in the data, the response to evidence is unchanged once the sharpening is removed, and lifts come 0.2 episode steps earlier than under the learner's own prior.
\emph{The signal remains decodable and the behavior is largely recoverable.} The rule log-odds remain decodable from the residual stream after both changes, so what post-training removes from the behavior it does not remove from the representation \citep{li2026erased}. Two thousand steps of the original data and objective restore the prior, the $\kappa$ curve and the latency within 200 to 600 steps.

\emph{No criterion shift.} A lift later than the learner's own prior predicts, with the evidence still decodable, did not appear under either the data or the objective.

\section{Related Work}\label{sec:related}

\textbf{The value of computation.} Rational metareasoning treats computation as an action chosen for its expected value \citep{russell1991principles,poh1993reasoning,desabbata2024rational}. \citet{hammar2026optimal} cast self-refinement as optimal stopping and obtain a threshold that does not depend on the remaining rounds, and reasoning models respond weakly to point values when they allocate a shared token budget \citep{fan2026thinking}. In these works computation improves the output of a fixed model. Our costly action changes which hypothesis family is available, and its return accrues over the remaining horizon. \citet{jhawar2026when} bound the error of any selector of an adaptation action when the evidence has the same law under deployments with different best actions, with a reference action that knows the deployment. Our reference policies hold the information of the learner.

\textbf{Creating and expanding hypotheses.} In nonparametric accounts of latent causes, creating a new cause is part of posterior inference \citep{gershman2010context,collins2013cognitive}. PROBE \citep{collins2012reasoning} stays with the current task set, switches to a monitored one, or creates a new one, and its creation criterion is a fixed satisficing threshold. Systems that enlarge a hypothesis pool likewise act when a statistic of fit crosses a fixed threshold, such as a per-form likelihood \citep{paramore2026learning} or a median residual \citep{murphy2026model}, or when a goal cannot be reached under the current model \citep{xiao2026hypothesis}. The distinction between inference within a framework, local repair of the framework and replacement of its basis is familiar from the history of science \citep{kuhn1962structure,lakatos1970falsification,gelman2013philosophy}; Appendix~\ref{app:framing} gives its computational reading.

\section{Discussion}\label{sec:discussion}

\textbf{Detection and decision are separable, and the language models separate them.} Their predictions move with the anomalies at every stage, as those of a learner that explains them inside the current family, while at the history level their decisions do not follow the evidence that moves those predictions. Told the gain of expanding, most respond to it as a trigger responds to a statistic, and the three given room to reason weigh it as the reference does. What reasoning supplies is valuation, the second half of the decision. The first half, turning failures into an estimate of what expansion would buy, does not appear in the models without reasoning or in the one that finishes its reasoning from the history alone. Without reasoning it is absent even when the prompt lists the failures and where they repeat, the counts from which the bounded reference follows price and horizon without $\mathcal F_1$ (Section~\ref{sec:value}). The models lack the computation that turns these counts into a value. For systems that escalate to costlier computation this locates the gap: reasoning can price a gain once it is given, and what remains to be built is the estimate. Late use of a richer family does not by itself indicate a biased criterion: the small Transformer that lost the family lifts when its implied prior predicts, so attributing lateness requires the curves indexed by the prior as a control.

\textbf{Limitations.} SRE is deliberately small, with sixteen inputs, binary labels and one richer family, so that the posterior can be enumerated exactly. The sequential reference is a two-step lookahead, and beyond SRE the boundary is solved in only two further description languages (Appendix~\ref{app:languages}). The learner is told the price and the horizon, which a deployed system must estimate. Language-model decisions are read from one-shot continuation probabilities (Appendix~\ref{app:prompts}), mostly at 7B and 8B, and the history-level result under reasoning rests on the one model that finishes within the token cap.

\phantomsection\label{end-of-main}
% ---- End of main text -------------------------------------------------------

% Required by the ICML 2026 template: unnumbered, before the references, outside the page limit.
\section*{Impact Statement}
This work studies how in-context learners decide when to revise their hypothesis space, using a small synthetic environment, small models trained from scratch and publicly released language models. Its findings concern the measurement of decision making under computational cost and carry no direct societal risk beyond those of the models studied.

\bibliography{references,references_appendix}
\bibliographystyle{icml2026}

\newpage
\appendix
\onecolumn
\renewcommand{\topfraction}{0.85}\renewcommand{\textfraction}{0.1}\setcounter{topnumber}{3}\setcounter{totalnumber}{4}

\section{The Structural Revision Environment}\label{app:sre}
% Appendix A: the Structural Revision Environment in full.

This appendix gives the environment of Section~\ref{sec:sre} in full: the generative process, the three predictors, the prices, the paired and designed histories, the text the language models read, and the two further worlds of Appendix~\ref{app:languages}.

\textbf{Generative process.} An episode has $T$ steps $t=0,\dots,T-1$ over the inputs $x\in\{1,\dots,16\}$ with binary labels; $T=40$ unless stated, and the remaining horizon at step $t$ is $H_t=T-t$. The class is drawn with probabilities 0.40, 0.20, 0.15 and 0.25 for \cls{noise}, \cls{exception}, \cls{drift} and \cls{rule}, and then:
\begin{itemize}\setlength\itemsep{0pt}
\item \cls{noise}: a threshold rule $y=\mathbf 1[x\ge b]$ with $b$ uniform on $\{1,\dots,16\}$;
\item \cls{exception}: a threshold rule whose label is inverted on a set $S$; every input belongs to $S$ with probability $\rho=0.06$, independently, conditioned on $1\le|S|\le2$, so that $|S|=1$ with probability 0.676 and $|S|=2$ with probability 0.324;
\item \cls{drift}: the threshold moves from $b$ to $b'\ne b$, uniform on the other fifteen values, at a step $\tau$ uniform on $\{\lfloor T/4\rfloor,\dots,\lfloor 3T/4\rfloor\}$ (10 to 30 at $T=40$), the first step that uses $b'$;
\item \cls{rule}: an interval $y=\mathbf 1[l\le x\le u]$ with $(l,u)$ uniform on the 120 pairs $1\le l\le u\le 15$, so that no interval equals a threshold.
\end{itemize}
In every class each observed label is flipped independently with probability $\varepsilon=0.08$. A family $\mathcal F_0$ is misspecified for the process when $\liminf_t \tfrac1t\,\mathbb E\big[\log p^{\ast}(z_{\le t})-\sup_{h\in\mathcal F_0}\log p_h(z_{\le t})\big]>0$, a relation between the family and the process that is fixed within an episode and never observed. The latent configurations number 7,352: 16 thresholds, 5,040 drifts, 2,176 exceptions and 120 intervals; configurations of different classes with the same label table, such as an exception at $x=b$ and the threshold at $b+1$, are kept apart. The queries come from their own random stream, independent of the class. The first query is uniform; afterwards a query repeats a uniformly chosen input that was queried before with probability $r=0.35$ and is otherwise a uniformly chosen input that was not, and it repeats whenever all sixteen inputs have been queried. At step $t$ the next query $x_{t+1}$ is announced.

\textbf{Predictors.} $E_0$ is the exact Bayesian predictive of the \cls{noise} and \cls{drift} classes with their prior renormalized: a posterior over thresholds that allows one change of threshold under the generative change-point prior (a change with probability $0.15/0.55$, at a step uniform on the drift window). It is the current family with its change-point tracking, and it is what the learner predicts with before any structural action.

$E_P$ is $E_0$ on the inputs that are not patched, combined with an independent Beta(1,1)--Bernoulli parameter for each patched input, which predicts that input's label from the labels observed there and nowhere else (one half before any observation). At most two inputs can be patched, each once, and $E_0$ no longer conditions on the patched inputs.

$E_1$ is the exact Bayesian predictive under the full generative model, all four classes and all 7,352 configurations, conditioned on the whole history at the moment it is switched on. Expansion is absorbing: from then on $E_1$ predicts every label, and the patch slots are dropped.

\textbf{Prices and cells.} The expansion price $c_E$ takes the values 0.5, 1, 2, 4, 8 and 16 and the patch price $c_p$ the values 0.1, 0.25, 0.5, 1 and 2, with defaults 4 and 0.5; deferring costs $c_o=0.1$ in the one-shot protocol. In the sequential protocol the horizon varies through the episode length $T\in\{16,24,32,40\}$, so the price grid has $6\times5\times4=120$ cells. The explicit learners are trained on the cells of $(c_E,H)$ outside a held-out set, the cells with $(i+j)\bmod 4=3$ on the lattice of the six values of $c_E$ and $H\in\{8,16,24,32,40\}$, and on intervals whose width is not 3, 7 or 11. The common states of the decision regression add $c_E\in\{0.25,32\}$ at every $c_p$, 40 price cells in all, and episodes of length 48. The one-shot solutions use $H\in\{10,20,30\}$, 41 log-spaced values of $c_E$ in $[0.25,16]$ and $c_p\in\{0.25,0.5,1\}$, and for the history of Figure~\ref{fig:overview} the horizons 5 to 40 and the prices 0.5 to 16 of its panel C; the phase diagrams of Figures~\ref{fig:phase} and~\ref{fig:app-F} are drawn from a coarser grid, 25 values of $c_E$ and the 11 histories of the phase family below with at most five anomalous inputs, each observed once or twice (Figure~\ref{fig:phase} shows those observed once).

\textbf{Matched and null-equivalent pairs.} Matched failures come from groups of four episodes, one per class, that share the query schedule and the label flips: the schedule and the noise are drawn once and applied to one latent truth per class. The null-equivalent pairs behind the matched-pair $\kappa$ are built at each evidence level $m=1,\dots,5$. A threshold $b$ and an interval top $u$ are drawn ($u$ uniform on $\{2,\dots,15-m\}$, $b$ uniform on $\{2,\dots,u\}$). Both histories observe $b$ with label 1 and $b-1$ with label 0 two or three times each, which pins the threshold, and further inputs at or below $u$ once or twice each with their threshold labels. $D$ then inverts the labels of $m$ distinct inputs above $u$, which an interval $[b,u']$ explains, while $D'$ inverts those of $m$ inputs in $\{b+1,\dots,u\}\cup\{1,\dots,b-2\}$, which no interval explains; corresponding inputs are observed equally often (one to three times), and the observations are shuffled into one order shared by both histories. The probe $x^\ast$ is an unqueried input above $u$. A candidate is kept if the exact predictives at $x^\ast$ differ between $D$ and $D'$ by less than 0.05 in logit under $E_0$ and by more than 0.5 under $E_1$; 400 pairs are kept at every level.

\textbf{Designed states.} The phase family fixes a threshold at 4 and places $m=0,\dots,7$ distinct anomalous inputs at the top of the range, each observed one to three times (22 histories); the same-history design has two anomalous labels at input 15 and input 16 never queried. Each history is presented in two orders, its last observation anomalous or normal, under an announced query on which the competing explanations disagree or one on which they agree (input 2), without slots and at $H\in\{10,20,30\}$: 270 designed states, each read at the 40 price cells. The count and diversity histories place the same number of anomalous observations on one input or spread them over several, and the region-size histories put two anomalous inputs at the two ends of a region of two to seven inputs that are otherwise unqueried. The waiting set pairs 182 state-cells at which the reference waits under an informative announcement and commits under a repeat of an anomalous input with 182 controls matched on the advantage of committing.

\textbf{What the language models read.} In the implicit protocol an episode is a header line and one observation per line; the two label symbols are drawn per group of episodes and never named. The block below is the first \cls{rule} episode of the test set (interval $[6,9]$, label 1 shown as X) in rendering R1 and context C0 of Appendix~\ref{app:prompts}, set here in four columns:
\begin{promptbox}[width=0.6\linewidth,title={implicit protocol: a rule episode in R1 and C0}]
\setlength{\multicolsep}{2pt}%
\begin{Verbatim}[fontsize=\footnotesize]
Each line gives a position and its label.
\end{Verbatim}
\begin{multicols}{4}
\begin{Verbatim}[fontsize=\footnotesize]
1 -> V
1 -> V
14 -> V
7 -> X
2 -> V
1 -> V
1 -> V
4 -> V
3 -> V
12 -> V
2 -> V
11 -> V
6 -> X
10 -> V
12 -> V
4 -> V
8 -> X
15 -> V
13 -> V
1 -> V
1 -> V
5 -> V
10 -> V
16 -> V
9 -> X
12 -> V
9 -> X
9 -> X
3 -> X
10 -> V
12 -> V
16 -> V
4 -> V
11 -> V
2 -> V
7 -> X
3 -> V
14 -> V
4 -> V
2 -> V
\end{Verbatim}
\end{multicols}
\end{promptbox}
The prompts of the explicit protocol are given in Appendix~\ref{app:prompts}.

\textbf{Feature world.} The sixteen inputs are the four-bit vectors $(x_1,\dots,x_4)$, labels are binary, and noise flips a label with probability 0.08. The current family reads one feature, $y=x_i$ or $y=\neg x_i$ (eight hypotheses). The richer family contains the Boolean functions of two features that depend on both: the 12 exclusive-or functions, negations included, the 24 conjunctions and the 24 disjunctions of two literals, 60 rules in all. A conjunction or a disjunction differs from its two nearest current hypotheses on four inputs each, an exclusive-or from every current hypothesis on eight. Classes and base rates, exceptions, drift (a change of feature hypothesis), the schedule, the announcement and $T=40$ are those of SRE; the latent configurations number 2,332.

\textbf{Shift world.} The inputs $x\in\{0,\dots,15\}$ carry labels in $\mathbb Z_{16}$. The current family contains the sixteen cyclic shifts $y=x+b\bmod 16$; a rule adds a deviation $c\in\{\pm1,\pm2,\pm4\}$ on the inputs that satisfy one of twenty predicates (the fifteen thresholds $x\ge a$, parity and the four residues modulo 4), $y=x+b+c\,\mathbf 1[x\in R]\bmod 16$, which gives 1,920 rules. Noise moves a label with probability 0.08 by $\pm d$, with $d\in\{1,\dots,8\}$ geometric of ratio one half and a uniform sign, so that the surprise of a deviation grows linearly with its size; an exception deviates reproducibly by one of the six rule deviations. The latent configurations number 77,632; the 384 rule configurations that coincide with exception configurations are kept.

% Appendix B: exact and bounded reference policies; exogeneity and the counterfactual utility.
\FloatBarrier
\section{Exact and bounded reference policies}\label{app:reference}

This appendix constructs the two reference policies of Section~\ref{sec:sre}, against which Section~\ref{sec:results} and Tables~\ref{tab:regret} and~\ref{tab:regret-rest} measure every learner, and derives the exact counterfactual utility on which regret and utility training rest.

\textbf{Assumptions.} The construction uses three properties of SRE. (i) The latent state takes 7,352 values at $T=40$: 16 thresholds, 2,176 pairs of a threshold and an exception set, 5,040 pairs of thresholds with a change point, and 120 intervals. The generative process, with $\varepsilon$, $\rho$, the class prior and the prior of the change point, is known to the reference. (ii) Structural actions leave the queries and labels that follow unchanged. (iii) Expansion is absorbing, and at most two inputs are patched, each once, as many as the largest exception set of the generative process.

\textbf{The reference policy.} Given the history, the posterior over the latent state is enumerated exactly. It gives the predictive of $E_1$, the law of every future label and with it the expected score of every predictor. In the one-shot protocol a single decision is taken, and the value of an action is the expected score of the predictor it selects over the $H$ scored steps minus its price, computed by Monte Carlo over future queries and labels with common random numbers across actions and prices (standard errors below 0.03 nats in the example of Section~\ref{sec:sre}). In the sequential protocol the reference takes the exact expectation over the labels of the current query $x_t$ and of the announced query $x_{t+1}$, with stay, patch and expand available at both steps, and values each leaf of this two-step tree by the best of four continuations, estimated on 256 rollouts from the posterior. The first never revises and the second expands at the leaf. The third expands 3, 8 or 16 steps after the leaf, with the stopping rule fitted by least-squares regression of the realized gain on the class posterior and the share of unqueried inputs \citep{longstaff2001valuing}, cross-fitted so that its value is that of a feasible rule. The fourth, when a slot is free, patches the first input that is queried again after at least half of its observations have been anomalous, and is valued as the expected gain of that patch minus $c_p$ times its probability. The fourth gives a free slot a value. Without it, staying is valued as never patching, and where expansion is expensive and patches are cheap the lookahead spends its slots on inputs not yet seen to fail. In the sequential protocol a patch is a commitment too, and its optimal time is the next query of an input that has failed.

\textbf{Calibration.} At $T=12$, with the query schedule given to both, the optimal policy is computed by exact dynamic programming and compared with the two-step lookahead with exact leaves, both evaluated exactly under each class, on 24 schedules at each of six price cells ($c_E/T\in\{0.08,0.5\}$, $c_p\in\{0.1,0.25,0.5\}$). The lookahead comes within 0.002 to 0.015 nats per episode of the optimum and takes the optimal action in 98.5\% to 99.9\% of the optimal policy's decision states, weighted by how often that policy visits them. With Monte Carlo leaves and the fitted stopping rule, as in the experiments, the reference loses at most 0.008 nats per decision on the states the optimal policy visits and takes its action on 96\% to 99\% of them. At $T=40$ no optimum is available, and the reference is exact in its posterior and approximate in its control. A variant that knows the class of the episode is above it in each of the 120 price cells of the grid, by 0.01 to 0.67 nats per episode, as an upper bound must be.

\textbf{The bounded reference policy.} Its state at step $t$ holds statistics computed from $E_0$ and from counting alone. It counts the distinct inputs queried and the inputs whose anomalies are single (one observation, anomalous), reproducible (at least two observations, a majority anomalous) or non-reproducible, an observation being anomalous when the smoothed posterior of $E_0$ gives its label a probability below one half. A refit flag records whether $E_0$ puts more than half of its mass on a change of threshold that has already occurred. The state also holds the category of the current query (new, clean, single, reproducible or non-reproducible), the number of slots in use and whether the current query is patched, and it does not include the announced query. Because the observations are exogenous, the joint law of these statistics and of the future scores of every predictor does not depend on the policy, and it is estimated from 200,000 episodes simulated from the generative prior at each episode length. The expansion decision is solved by backward induction over $t$: in state $s$ the bounded policy expands when
\begin{equation}\label{eq:B-bounded}
\mathbb E\big[\,W_t-c_E-\big(\ell_t-c_p\mathbf 1[\mathrm{patch}_t]+V_{t+1}\big)\,\big|\,s_t=s\,\big]>0,
\end{equation}
where $W_t$ is the realized score of $E_1$ from $t$ to the end, $\ell_t$ the log score of step $t$ without expansion and $V_{t+1}$ the realized value, on the same simulated path, of the policy already solved for the later steps. The average over the paths in state $s$ marginalizes over the posterior given $s$, so no Markov property of the count state is needed. This is the least-squares estimator of \citet{longstaff2001valuing} with the count state as a tabular basis. Patching follows the best of three fixed rules, chosen for each price cell by cross-fitting: never patch, patch a reproducible anomaly, or patch a single or reproducible one. The gap between the two references therefore bounds the price of boundedness from above.

\textbf{The price of boundedness.} The reference can test whether distinct anomalies are predicted by one rule, and the bounded policy can only count them. A heuristic comparison locates the difference. For the reference, each further reproducible anomalous input that a rule predicts adds $\log\{(1-\rho)/\rho\}$ to the configuration odds of the rule against the exception list, whatever the size of the region (Equation~\eqref{eq:C-config}). A policy that sees only whether inputs are anomalous can use only their rate: under a rule a share $q_R=(16-u)/16$ of the inputs is anomalous and under the exception list a share of about $\rho$, so an anomaly adds about $\log(q_R/\rho)$. The heuristic compresses position and label into one indicator, and the bounded policy works on the simulated laws of its state instead, so it gives the shape of the gap and no numbers. When the region is small, $q_R$ approaches $\rho$ and counts cannot separate a rule from one or two exceptions. The missing evidence turns into delay, whose cost is the log loss not avoided in the meantime, so the gap vanishes where both policies expand at once or neither expands and is largest in between. Measured, it averages 0.24 nats per episode over the 120 price cells, small where expansion is cheap and largest at intermediate prices (Appendix~\ref{app:same-history}). At the default cell the bounded reference expands one step later at the median, expands in 27\% of \cls{exception} episodes against the reference's 13\% and loses 1.48 nats per \cls{rule} episode (Tables~\ref{tab:regret} and~\ref{tab:regret-rest}), and in no cell is it above the reference by more than two standard errors. Its decisions follow price and horizon as the reference's do (Appendix~\ref{app:statistics}). What it lacks is the evidence that separates a small anomalous region above $u$ from one or two exceptions, the part of the expansion decision that requires $\mathcal F_1$ to be instantiated.

\textbf{Exogeneity and the counterfactual utility.} Fix an episode $\omega=(x_1,y_1,\dots,x_T,y_T)$. By (ii), given the prefix and the past actions, the slot state and the predictor in use are determined, and so is the next prediction. By induction over $t$, a non-anticipating policy $\mu$, with any internal randomness fixed, has the utility of Section~\ref{sec:problem}
\begin{equation}\label{eq:B-utility}
\mathcal U_\mu(\omega)=\sum_{t=1}^{T}\log q^{\mu}_t(y_t\mid x_t)-c_p\,N^{\mu}_{P}(\omega)-c_E\,\mathbf 1[\sigma^{\mu}_E\le T],
\end{equation}
where $q^\mu_t$ is the predictive of the predictor that $\mu$ uses at step $t$ ($E_0$, $E_P$ with its slots, or $E_1$ from its expansion time $\sigma^\mu_E$ on) and $N^\mu_P$ its number of patches. In the one-shot protocol defer subtracts $c_o$. Every policy is therefore scored on the same episodes without importance weights, and the regret of Appendix~\ref{app:table1} is exact episode by episode.

\textbf{Regret decomposition.} Let $F_t(s;\omega)$ be the utility collected from step $t$ on by the reference started in information state $s$ on the suffix of $\omega$, let $s^\mu_t$ be the state that $\mu$ reaches and $a^\mu_t$ its action there, and let $\ell^\mu_t(\omega)$ be its log score at step $t$ and $C(a^\mu_t)$ the price of that action. The disagreement cost $d_t=F_t(s^\mu_t;\omega)-\ell^\mu_t(\omega)+C(a^\mu_t)-F_{t+1}(s^\mu_{t+1};\omega)$ is zero whenever $\mu$ takes the reference's action in $s^\mu_t$, since both then collect the same reward and reach the same state. With $F_1(s_1;\omega)=\mathcal U_{\mathrm{ref}}(\omega)$ and $F_{T+1}=0$ the sum telescopes:
\begin{equation}\label{eq:B-decomposition}
\mathcal U_{\mathrm{ref}}(\omega)-\mathcal U_\mu(\omega)=\sum_{t=1}^{T}d_t=R_{\mathrm{FE}}+R_{\mathrm{FL}}+R_{\mathrm{LAT}}+R_{\mathrm{PT}},\qquad R_j=\sum_{t}\mathbf 1[\chi_t=j]\,d_t ,
\end{equation}
where $\chi_t$ classifies the disagreement at step $t$: false expansion (FE) when $\mu$ expands and the reference stays, false lift (FL) when $\mu$ expands and the reference patches, latency (LAT) when the reference expands and $\mu$ does not, and patch-type error (PT) when one of the two patches the current input and the other stays. FE and FL are the expansions made where $\Delta Q^\ast<0$ (Appendix~\ref{app:table1}), split by the alternative the reference prefers, and in a \cls{rule} episode $R_{\mathrm{LAT}}$ is the cost of excess latency. Exogeneity makes the total exact, and the classification is one exhaustive convention for dividing it. A term can be negative on a single episode. With the optimal policy as reference, $\mathbb E[d_t\mid s^\mu_t,a^\mu_t]=V^\ast(s^\mu_t)-Q^\ast(s^\mu_t,a^\mu_t)\ge0$ and every term has a nonnegative expectation. With the approximate reference this holds up to its approximation error. Tables~\ref{tab:regret} and~\ref{tab:regret-rest} report the regret by class and, as counterparts of the four terms, rates by class: the shares of \cls{noise}, \cls{drift} and \cls{exception} episodes in which a policy expands, its excess latency in \cls{rule} episodes and the share of episodes in which its type of revision differs from the reference's.

\textbf{Targets for utility training.} The same identity gives unbiased targets for learning the revision decision from utility alone. The realized return of expanding at step $t$, $Y^{E}_t=-c_E+\sum_{s=t}^{T}\log E_{1,s}(y_s\mid x_s)$ with $E_{1,s}$ the predictive of $E_1$ at step $s$, is available on every episode, and $\mathbb E[Y^{E}_t\mid s_t]=Q(s_t,\mathrm{expand})$ at every step reached by a non-anticipating rule. For a continuation policy $\nu$, the step reward of staying or patching plus the realized utility of $\nu$ from $s_{t+1}$ is an unbiased target for $Q^\nu(s_t,a)$. The learners of Section~\ref{sec:learners} trained on utility regress on these targets with $\nu$ their own greedy continuation, so each update evaluates the current greedy policy and the greedy choice improves it. With a sufficient state and exact fits, a fixed point of this scheme satisfies the Bellman equation of the stopping problem \citep{tsitsiklis2001regression}. A fit under one fixed $\nu$ gives $Q^\nu$, which equals $Q^\ast$ only when $\nu$ is optimal.

\textbf{Fixed thresholds as fixed prices.} The one-step rule of Section~\ref{sec:problem} expands when $v>\lambda$, with $v=(\Delta+\delta)p-\delta$ the value of expanding per remaining step and $\lambda=c_E/H$. A trigger that expands when a statistic monotone in $p$ crosses a fixed level $p_0$ applies the same rule with $\lambda$ fixed at $\lambda_0=(\Delta+\delta)p_0-\delta$, and its one-step regret per remaining step is $R/H=|v-\lambda|\,\mathbf 1\big[\mathbf 1[v>\lambda_0]\neq\mathbf 1[v>\lambda]\big]$. Where the amortized price exceeds $\lambda_0$, the trigger expands in the states with $\lambda_0<v<\lambda$ and its loss grows with the price. Where the price falls below $\lambda_0$, it waits in the states with $\lambda<v<\lambda_0$ and its loss grows as the price falls. Against a patch the amortized price difference $(c_E-c_p)/H$ replaces $\lambda$. The reading needs the statistic to be monotone in $p$ within the state, which a residual, a fitted likelihood or the surprise of one observation need not be. The e-process trigger tuned at the default cell loses 1.19 nats per episode on average over the 120 cells of the grid, against 0.42 with its threshold tuned again in each cell (Table~\ref{tab:regret}), most of it at high amortized prices, where it keeps expanding.

% Appendix C (derivations): collapse, one-step threshold and monotonicity, belief crossing,
% decision boundary, catch-all evidence, Occam and catch-up.
\FloatBarrier
\section{Normative fragments}\label{app:normative-derivations}\label{app:normative}

This appendix states the patch-or-expand comparison and derives it, together with the collapse to model averaging, the one-step threshold and the guarantees of the structural evidence stated in Section~\ref{sec:problem}. Logarithms are natural. We write $P^\ast_t(\cdot\mid x)$ for the predictive of $E_1$, the posterior predictive of the full generative model, and $E_{a,t}$ for that of the predictor selected by action $a$. The prior masses of the \cls{rule} and \cls{exception} classes are $\pi_{\mathrm{rule}}=0.25$ and $\pi_{\mathrm{exc}}=0.20$.

\textbf{Collapse to model averaging.} When every family is instantiated at every step and no structural action has a price, the predictor that maximizes expected utility is the Bayesian model average $P^\ast_t(y\mid x)=\sum_\theta p_\theta(y\mid x,D_t)\Pr(\theta\mid D_t)$ over the full latent space, with $D_t=z_{<t}$. The log score is strictly proper \citep{gneiting2007strictly}: for every predictive $q$,
\begin{equation}\label{eq:C-proper}
\mathbb E_{Y\sim P^\ast_t(\cdot\mid x)}\big[\log P^\ast_t(Y\mid x)-\log q(Y\mid x)\big]=\operatorname{KL}\big(P^\ast_t(\cdot\mid x)\,\Vert\,q\big)\ge0,
\end{equation}
with equality only at $q=P^\ast_t(\cdot\mid x)$. Since actions do not change the queries and labels, Equation~\eqref{eq:C-proper} holds at every step of every future history, and summing it gives the claim. At zero prices nothing offsets this advantage, so free expansion weakly dominates staying with $E_0$. The predictors still predict differently and only the decision disappears, so a collapse control compares a learner's predictions with the model average. At the lowest amortized price of the grid the reference comes close to the collapse and expands in almost every \cls{noise} episode, mostly at the first step (Appendix~\ref{app:same-history}).

\textbf{One-step threshold.} With $p=p_t$, $\Delta$ and $\delta$ as in Section~\ref{sec:problem} and $\Delta+\delta>0$, expanding for the remaining $H$ steps rather than staying is worth $H\{p\Delta-(1-p)\delta\}-c_E$, which is positive exactly when $p>(c_E/H+\delta)/(\Delta+\delta)$. At fixed $\Delta$ and $\delta$ the threshold $p^\ast$ has $\partial p^\ast/\partial c_E=1/\{H(\Delta+\delta)\}>0$ and $\partial p^\ast/\partial H=-c_E/\{H^2(\Delta+\delta)\}\le0$, and it depends on price and horizon only through $c_E/H$. The base rate enters through the posterior, $\operatorname{logit}p=\log\{\pi_{\mathrm{rule}}/(1-\pi_{\mathrm{rule}})\}+\Lambda(D)$ with $\Lambda(D)$ the log-likelihood ratio of the history for the \cls{rule} class against the others, so a higher base rate lowers the evidence that the threshold requires. Belief in the richer family and amortized price are therefore the natural coordinates, and reading the comparison as one threshold across states requires $\Delta$ and $\delta$ not to vary with $p$.

\textbf{Price and horizon in the sequential problem.} Two monotonicity statements hold without any ordering of the evidence. For the price, let $\mu$ be a policy that does not expand now, $A_\mu$ its value before expansion prices and $e_\mu$ its probability of expanding later, and let $A_E$ be the value of expanding now before its price. Relative to $\mu$, expanding now is worth
\begin{equation}\label{eq:C-price}
(A_E-c_E)-(A_\mu-c_E\,e_\mu)=A_E-A_\mu-c_E(1-e_\mu),
\end{equation}
which does not increase with $c_E$. The optimal advantage $\Delta Q^\ast$ is the infimum of Equation~\eqref{eq:C-price} over $\mu$ and does not increase either, so the expansion region shrinks as $c_E$ rises. For the horizon, Equation~\eqref{eq:C-proper} turns the maximization of utility into the minimization of the expected score regret against the model average plus prices. Let $W_H(s)$ be the minimal such loss over $H$ remaining steps and $d_a(s)\ge0$ the expected one-step divergence of a non-expanding action $a$. Expansion costs exactly $c_E$ from then on, so
\begin{equation}\label{eq:C-horizon}
W_0=0,\qquad W_H(s)=\min\{c_E,\,\widetilde W_H(s)\},\qquad \widetilde W_H(s)=\min_{a\in\{\mathrm{stay},\,\mathrm{patch}\}}\Big\{C(a)+d_a(s)+\mathbb E\big[W_{H-1}(s')\mid s,a\big]\Big\},
\end{equation}
with $C(\mathrm{stay})=0$ and $C(\mathrm{patch})=c_p$ as in Section~\ref{sec:problem}. Induction from $W_0=0$ gives $W_H\ge W_{H-1}$ and $\widetilde W_H\ge\widetilde W_{H-1}$, so the expansion region $\{s:\widetilde W_H(s)>c_E\}$ grows with $H$. On any slice of states along which the expansion region is an upper set in the evidence, the evidence threshold therefore rises with $c_E$ and falls with $H$. The horizon statement extends the horizon with beliefs and transition kernel held fixed. In SRE a change of episode length also moves the window of the change point.

\textbf{Evidence and base rate in the bounded policy.} For the bounded reference, monotonicity in the anomaly evidence and in the base rate of rules would follow from the comparative statics of the Bellman recursion if its stage values were monotone and its transition laws preserved the order \citep{topkis1998supermodularity}. Its laws are estimated by simulation (Appendix~\ref{app:reference}), so the conclusions are checked on its solved table. At fixed step, distinct inputs, non-reproducible anomalies, refit flag, query category and slots, expansion at one level of anomaly evidence implies expansion one unit higher (one more single anomaly, one more reproducible anomaly, or a single anomaly that reproduces) in all but 1.8\% of the comparable pairs of visited states (2.6\% for the last). Over 90\% of the violations involve a state visited fewer than 20 times in the simulation, and among pairs of states visited at least 50 times they fall to 0.15\% to 0.18\% (0.92\% for the last). Raising the base rate of rules from 0.1 to 0.25 and to 0.4, the other classes rescaled, enlarges the expansion region: among states visited at least 50 times, those that begin to expand outnumber those that stop by factors of 13 to 148. Both monotonicities hold up to the sampling noise of the table.

\textbf{Patch or expand.} Let $m$ be the number of distinct inputs at which reproducible anomalies have been observed. An exception list explains them at a description length that grows linearly in $m$, and it predicts nothing at inputs not yet queried. A rule in $\mathcal F_1$ explains them at constant length and predicts the anomalies still to come. Belief and decision come apart. The point at which the rule becomes more probable, $m^{\ast}_B$, is the $m$ at which the posterior favors the rule over the exception list,
\[
m^{\ast}_B\approx\frac{\Delta L-\log(\pi_{\mathrm{rule}}/\pi_{\mathrm{exc}})}{\log(K-1)+\log(1/\rho)},
\]
where $\Delta L$ is the description length that the rule adds to the hypothesis in $\mathcal F_0$ it replaces, $\pi_{\mathrm{rule}}/\pi_{\mathrm{exc}}$ the prior odds of \cls{rule} against \cls{exception}, $K-1$ the number of labels an exception can take and $\rho$ the prior probability that an input is an exception. The decision follows a different comparison. A patch and a rule fit the observed anomalies equally well, and they differ at the inputs that the rule predicts and no slot covers. Let $n$ be the number of such inputs and $g$ the expected loss per step that one of them causes, which is its share of the queries times the divergence between the two predictions at that input. With the rule established, expansion is optimal when
\[
n\,g\,H>c_E-c_p ,
\]
so the price enters as the amortized price difference $(c_E-c_p)/H$. Diversity enters this comparison in two ways. It raises the posterior of the rule, and this saturates after two or three distinct anomalous inputs. It can also enlarge the region that the rule is inferred to cover, and with it $n$. Expansion becomes optimal at $m^{\ast}_D$, the smallest $m$ at which it beats every other action. Repetitions of one anomalous input separate noise from structure. Once an anomaly has reproduced, further repetitions leave the boundary where it is.

\textbf{When the rule becomes more probable.} Consider a description language in which a rule explains $m$ anomalies at a constant incremental description length $\Delta L$, and an exception list encodes each anomalous input separately, at $\log(1/\rho)$ for being an exception and $\log(K-1)$ for its label. For two explanations of the same anomalies with equal likelihood,
\begin{equation}\label{eq:C-mdl}
\log\frac{\Pr(\mathrm{rule}\mid D)}{\Pr(\mathrm{exc}\mid D)}\simeq\log\frac{\pi_{\mathrm{rule}}}{\pi_{\mathrm{exc}}}-\Delta L+m\Big\{\log(K-1)+\log\frac1\rho\Big\},
\end{equation}
which crosses zero at the $m^\ast_B$ above. Description length enters as a negative log prior weight \citep{rissanen1978modeling,grunwald2007minimum}, and the approximation neglects the normalization of the exception prior, factors of order $\rho$ and the residual likelihood. With a non-uniform distribution of exception labels, the negative log probability of the observed label replaces $\log(K-1)$.

In SRE the computation is exact. The interval's lower end takes the place of the threshold, so $\Delta L=\log(120/16)$. One interval has prior mass $\pi_{\mathrm{rule}}/120$, and one threshold with an exception set of size $m$ has mass $\pi_{\mathrm{exc}}\rho^m(1-\rho)^{16-m}/(16Z)$, where $Z=\sum_{j=1}^{2}\binom{16}{j}\rho^j(1-\rho)^{16-j}=0.5611$ normalizes the prior over $1\le|S|\le2$. For $m=1,2$ the log-odds of the two configurations are
\begin{equation}\label{eq:C-config}
\ell^{\mathrm{cfg}}(m)=\log\frac{\pi_{\mathrm{rule}}}{\pi_{\mathrm{exc}}}-\log\frac{120}{16}+C_{16}+m\log\frac{1-\rho}{\rho}=-1.3795+2.7515\,m,\qquad C_{16}=\log Z-16\log(1-\rho)=0.4122,
\end{equation}
whose linear interpolation crosses zero at $m^\ast_B=0.5014$, against 0.6369 from Equation~\eqref{eq:C-mdl}. Of the gap of 0.135, the normalization term $C_{16}$ lowers $m^\ast_B$ by 0.150 and the factor $\log(1-\rho)$ in the slope raises it by 0.014. No exception configuration has $m\ge3$, so Equation~\eqref{eq:C-config} is a prior odds only up to $m=2$ and a linear extrapolation beyond. The posterior odds of the two classes add a term $\zeta(D)$ for the likelihood differences and the weighted number of compatible configurations: in the history of Section~\ref{sec:sre}, with $m=1$, $\ell^{\mathrm{cfg}}(1)=1.37$ while the posterior gives $\log(0.710/0.213)=1.20$.

The same algebra gives $m^\ast_B$ in the other two worlds (Appendix~\ref{app:languages}). In the feature world the full library, the and-rules and the xor-rules are 7.5, 3 and 1.5 times as large as the eight single-feature rules, which gives the exact values 0.50, 0.17 and $-0.08$. In the shift world $\Delta L=\log120$ (20 predicates and 6 deviations for each shift) and an exception takes one of $K-1=6$ deviations ($\pm1$, $\pm2$, $\pm4$), so the exact slope is $\log6+\log\{(1-\rho)/\rho\}=4.543$ and $m^\ast_B=(\log120-\log1.25-C_{16})/4.543=0.914$, against 0.991 in closed form. The gap of 0.08 has the same two sources, the normalization term lowering $m^\ast_B$ by 0.09 and $\log(1-\rho)$ in the slope raising it by 0.01.

\textbf{Decision boundary.} Let $a$ be the committing action that competes with expansion in the one-shot protocol and $c_{\mathrm{alt}}=C(a)$ its price. Summing Equation~\eqref{eq:C-proper} over the $H$ scored steps,
\begin{equation}\label{eq:C-gain}
Q(\mathrm{expand})-Q(a)=G_{t,H}(a)-(c_E-c_{\mathrm{alt}}),\qquad G_{t,H}(a)=\mathbb E_t\sum_{s=t}^{t+H-1}\operatorname{KL}\big(P^\ast_s(\cdot\mid X_s)\,\Vert\,E_{a,s}(\cdot\mid X_s)\big),
\end{equation}
with both predictors updating as the protocol prescribes. Freeze both predictions at $t$ and suppose that they differ only on the set $\mathcal N$ of inputs that the rule predicts and the competitor does not cover. Let $q_x=H^{-1}\mathbb E_t\sum_s\mathbf 1[X_s=x]$ be the share of the remaining queries that falls on $x$, computed from the query schedule of Section~\ref{sec:sre} with its repeat probability $r$, and $g_x=q_x\operatorname{KL}(P^\ast_t\Vert E_{a,t})_x$. Expansion beats $a$ when
\begin{equation}\label{eq:C-boundary}
H\sum_{x\in\mathcal N}g_x>c_E-c_{\mathrm{alt}},
\end{equation}
which with $n=|\mathcal N|$ and $g$ the mean of the $g_x$ is the comparison above. Expansion is optimal when it beats every other action, waiting included. The two predictions also differ, with different confidence, where both explain the anomalies: a Beta(1,1) slot that has seen an anomalous label twice predicts it with probability 0.75, a saturated rule with $1-\varepsilon=0.92$, a divergence of 0.097 nats per query. With $R_{\mathrm{fit}}\ge0$ for these terms outside $\mathcal N$, $\Gamma_1$ for the expected improvement of $E_1$ over its frozen prediction within the horizon and $\Gamma_a$ for that of the competitor,
\begin{equation}\label{eq:C-corrections}
\frac{G_{t,H}(a)}{H}=\sum_{x\in\mathcal N}g_x+\frac{R_{\mathrm{fit}}+\Gamma_1-\Gamma_a}{H}.
\end{equation}
Three corrections separate Equation~\eqref{eq:C-boundary} from the exact comparison. (i) Relative to an established rule, an unsaturated posterior lowers the price that expansion can repay. If the non-rule predictive at $x$ is the competitor's $A_x$, then $P^\ast=pL_x+(1-p)A_x$, and convexity gives $\operatorname{KL}(P^\ast\Vert A_x)\le p\operatorname{KL}(L_x\Vert A_x)$. Near $p=1$ the reduction is $(1-p)\sum_xq_x\{\operatorname{KL}(L_x\Vert A_x)+\operatorname{KL}(A_x\Vert L_x)\}$, and a $g_x$ computed from the posterior predictive already contains it. (ii) The competitor keeps learning within the horizon, and $\Gamma_a$ lowers the boundary when it is positive, which for a misspecified competitor it need not be. (iii) $E_1$ keeps sharpening its estimate of the region. For the Bayes predictor $\Gamma_1=\mathbb E_t\sum_s\operatorname{KL}(P^\ast_s\Vert P^\ast_t)_{X_s}\ge0$, the information that the intervening labels carry about each future one, and it raises the boundary. The deviations measured over the three worlds have these signs (Appendix~\ref{app:languages}). The solved boundary price is a median of 0.90 times the prediction of Equation~\eqref{eq:C-boundary} where the region covers half of the inputs and the competitor learns, 1.20 times with a single anomalous input, whose region $E_1$ keeps tightening, and 1.02 times with at least two anomalous inputs and at most six inputs in $\mathcal N$ (10th to 90th percentile 0.99 to 1.21).

Defer is defined by values. Let $V_{\mathrm{never}}(s)$ be the value of never revising and $V^{a}_{\mathrm{now}}(s)$ the value of committing to action $a$ immediately. A state $s$ is a defer state when $V^{\ast}(s)>\max\{V_{\mathrm{never}}(s),\max_a V^{a}_{\mathrm{now}}(s)\}$. In a stay state, further monitoring has no value.

\textbf{Four consequences.} (1) In SRE the region is the $m$ observed top inputs and the unqueried inputs between them and the highest normal input, 0.7 on average under the posterior, and a slot covers one input, so $|\mathcal N|\approx m$ and each further anomalous input adds a step. At $H=20$, Equation~\eqref{eq:C-boundary} places the boundaries for $m=2$ to 5 at $(c_E-c_p)/H=0.16$, 0.30, 0.43 and 0.55, with $g_x$ rising from 0.08 to 0.11 as the posterior sharpens, and the solved boundaries lie slightly higher, at 0.20, 0.34, 0.45 and 0.56. (2) The price of the competitor enters as $c_{\mathrm{alt}}$, so against a patch the boundary is fixed in $(c_E-c_p)/H$ and against staying in $c_E/H$. In SRE the mean spread of $m^\ast_D$ across $H=10,20,30$ at matched price is 0.21 in $(c_E-c_p)/H$ and 0.49 in $c_E/H$. Where the region covers half of the inputs, a slot is too small to compete and the order reverses: 0.12 in $c_E/H$ against 0.23 for the xor-rules of the feature world, 0.16 against 0.27 for the parity rules of the shift world. (3) Where the type of rule fixes the region, $|\mathcal N|$ does not grow with $m$, and $m$ acts only through the posterior, which saturates at $m=2$. Then $\sum_{x\in\mathcal N}g_x$ is a ceiling on the price per remaining step above which no $m$ makes expansion optimal: 0.40 to 0.47 for the and-rules of the feature world and 0.80 to 1.01 for the residue rules of the shift world ($|\mathcal N|=3$), 1.9 to 2.1 for its parity rules ($|\mathcal N|=7$). (4) $E_1$ contains the current family, so as $c_E\to0$ expansion loses nothing but its price. Wherever $P^\ast_t$ already differs from the competitor at inputs with query mass, a low enough price makes expansion optimal without any anomaly, and $m^\ast_D=0<m^\ast_B$. At the lowest price of the grid, $c_E=0.25$, $m^\ast_D$ is 0 to 0.4 in the feature and shift worlds, below their $m^\ast_B$ of 0.50 and 0.91, and 0 to 0.65 in SRE. Only once the price exceeds a small level does $m^\ast_D$ lie beyond $m^\ast_B$.

\textbf{Structural evidence.} The catch-all predictor of Section~\ref{sec:problem} is the per-input Krichevsky--Trofimov estimator \citep{krichevsky1981performance}, $U(y\mid x,D)=(n_{x,y}+1/2)/(n_x+K/2)$, where $n_{x,y}$ counts the labels $y$ observed at $x$. It differs from the Beta(1,1) slot of $E_P$. The supremum in $\mathcal E_t$ runs over the 16 thresholds with noise level $\varepsilon$, and the query factors cancel. For every $h_0\in\mathcal F_0$, $\Pr_{h_0}(\exists t:\mathcal E_t\ge\log(1/\alpha))\le\alpha$. Under $h_0$ the likelihood ratio $M_t=\prod_{s\le t}U(Y_s\mid X_s,z_{<s})/p_{h_0}(Y_s\mid X_s)$ is a nonnegative martingale with $M_0=1$, because $U$ is predictable and the queries do not depend on the labels. Since $\sup_h p_h\ge p_{h_0}$, $e^{\mathcal E_t}\le M_t$, and Ville's inequality \citep{ville1939etude} for $M_t$ gives the bound. The process $e^{\mathcal E_t}$ need not be a supermartingale itself, and domination by a test martingale suffices \citep{shafer2011test,ramdas2023gametheoretic}. The null is the static threshold family, so a \cls{drift} episode lies outside it unless the paths with a change point enter the supremum. If the label law $P_x$ at each input is stationary and the query frequencies converge to $q_x$, the logarithmic regret of the estimator at each input and the law of large numbers over the finite family give
\begin{equation}\label{eq:C-rate}
\frac{\mathcal E_t}{t}\longrightarrow\min_{h\in\mathcal F_0}\sum_x q_x\operatorname{KL}\big(P_x\,\Vert\,p_h(\cdot\mid x)\big)\quad\text{almost surely},
\end{equation}
the per-step divergence that $\mathcal F_0$ cannot remove. It is positive in a \cls{rule} episode and zero when the misspecification lies only at inputs that are never queried. The e-process trigger (Appendix~\ref{app:prompts}) thus bounds false expansion under the null at all times and responds to none of $c_E$, $c_p$ and $H$.

\emph{Remark.} The structural evidence $\mathcal E_t$ of Section~\ref{sec:problem} is the cumulative log-likelihood ratio between the catch-all predictor and the best hypothesis in $\mathcal F_0$.
Its one-step form coincides with the Bayes Factor Surprise of \citet{liakoni2021learning}, with the catch-all predictor in place of the prior predictive. A single extreme label raises $S_t$ and leaves $\mathcal E_t$ almost unchanged. A deviation that recurs at the same input, or across inputs related by a rule, raises $\mathcal E_t$ at every recurrence.

\textbf{Normative delay.} Slow concentration of the posterior is not evidence of a structural delay. In a regular parametric model the log marginal likelihood ratio is $\ell_1(\hat\theta_1)-\ell_0(\hat\theta_0)-(\Delta d/2)\log n+O(1)$ \citep{schwarz1978estimating}, whose last term is an Occam penalty that the richer model must first overcome, and in prediction a richer model that has become the better predictor must still recover the losses of its early predictions. \citet{vanerven2012catching} analyze this catch-up phenomenon and the switch distribution, which mixes over switching times so that a switch does not carry the full early loss of the new model, with faster convergence for the model classes and risks they study. In SRE the library is finite and prior and likelihoods are exact, so the penalty is the prior term $\log(120/16)$ of Equation~\eqref{eq:C-config} in place of $(\Delta d/2)\log n$, and the model average holds the richer family from the first step yet needs time to concentrate on it. Latency is therefore measured against a reference with the same prior, noise, prices, horizon and information (Appendix~\ref{app:table1}). A delay counted from the first anomaly, or from the step at which the posterior crosses one half, would include normative learning and optimal waiting.

\FloatBarrier
\section{Further facts of the normative solution}\label{app:same-history}
\begin{figure}[htb]
  \centering
  \includegraphics[width=0.5\textwidth]{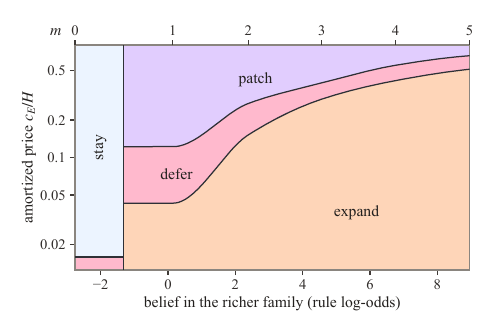}
  \caption{\textbf{The normative phase diagram of SRE.} The optimal one-shot action over the reference's posterior log-odds of the \cls{rule} class and the amortized price $c_E/H$, on histories with a threshold at 4 and $m=0$ to 5 anomalous inputs at the top of the range, each observed once ($H=20$, $c_p=0.5$, $c_o=0.1$, informative announcement). The dark lines are the boundaries between the regions of the optimal action. The boundaries quoted in Section~\ref{sec:sre} are those of an uninformative announcement, in $(c_E-c_p)/H$.}
  \label{fig:phase}
\end{figure}

This appendix collects further facts of the normative solution of Section~\ref{sec:sre}, starting with its phase diagram (Figure~\ref{fig:phase}) and its boundaries in the amortized price difference at each number of anomalous inputs.

When the announced query is informative, waiting takes the upper part of the expansion region at each number of anomalous inputs, and immediate expansion ends 0.04 to 0.07 below the boundaries of Section~\ref{sec:sre}. An anomaly observed twice counts as reproduced, and a third observation moves $m^\ast_D$ (Appendix~\ref{app:normative}) by less than 0.08 at nine prices in ten. An anomaly observed once leaves label noise as an explanation and raises $m^\ast_D$ slightly.

\emph{The reference often waits, and sometimes never expands.} At the default prices it expands in 75\% of \cls{rule} episodes, after waiting a median of 5 steps from the first informative observation. When a single input lies above $u$, a \cls{rule} episode is observationally equivalent to an exception, and the reference policy expands in 22\% of such episodes and patches in most of the others.

\emph{Expansion inside the family can be optimal.} The full model contains the current family, so expanding loses only the price. At an amortized price of 0.0125 the reference policy expands in 98\% of \cls{noise} episodes, most of them at the first step. At 0.025 the share is 31\%, at 0.1 it is 7\%, and at 0.4 it is zero. Class labels therefore do not define expansion errors, and we define them by $\Delta Q^{\ast}$.

\emph{Counts carry most of the decision.} The bounded reference policy, obtained by backward induction over its count statistics, expands one step later than the full reference at the median and loses 0.49 nats per episode at the default prices, with 1.5 nats lost in \cls{rule} episodes. Its loss is 0.01 nats at the lowest amortized price and peaks at about 0.6 nats at intermediate prices. The triggers that respond to a single statistic lose more: 0.68 nats for the e-process, 0.81 for a likelihood threshold, 1.6 for an anomaly count and 2.0 for a surprise threshold, against 2.2 for never revising. Their thresholds were tuned at the default prices. Carried to the other prices and horizons of the grid, the e-process threshold loses 1.2 nats on average, and 0.4 when it is tuned again in every cell. A fixed threshold is the optimal rule for one amortized price.

Besides the decision regression of Section~\ref{sec:learners}, the learners are read on designed histories.

\textbf{Designed contrasts.} The regression summarizes, and the identification rests on histories built in pairs. Matched failures present the same failed predictions under different classes. Histories with the same number of anomalous observations place them on one input or spread them over several, which separates count from diversity. One history is presented under different prices and horizons, and under announced queries on which the competing explanations disagree or agree.

\FloatBarrier
\section{The boundary across description languages}\label{app:languages}
The facts of Section~\ref{sec:sre} were obtained in one description language. We solved two further worlds on the same skeleton. In the feature world the current family contains the rules that read one of four binary features, and the richer family contains the Boolean functions of two features. In the shift world labels take 16 ordered values, the current family contains the cyclic shifts, and a rule adds a deviation on the inputs that satisfy a predicate. Three facts carry over unchanged. An announced query on which the competing explanations disagree makes waiting optimal, and an uninformative one makes its value exactly zero. Raising the price moves the optimal action from expand through defer to patch. A third observation of an anomalous input leaves $m^\ast_D$ where it was.

The rise of the boundary with the number of anomalous inputs (Section~\ref{sec:sre}) does not carry over, and the comparison of Appendix~\ref{app:normative} says why. In the two new worlds the type of the rule fixes the region it covers, so additional anomalous inputs raise the posterior and leave $n$ unchanged. The boundary rises through two or three inputs and then meets a ceiling at $n\,g$, above which no number of anomalous inputs makes expansion optimal. Where the region covers half of the inputs, a slot is too small to compete, expansion is compared with staying, and the horizons collapse in $c_E/H$. Computed from the exact predictive distributions, the comparison predicts the boundary to within a median ratio of 0.98, solved over predicted, over 90 histories at three horizons in the three worlds (10th to 90th percentile 0.78 to 1.06). In the shift world, three designs with the same total magnitude of deviation have three different optimal actions. A single large deviation that does not reproduce is left alone at every price, a deviation that reproduces at one input is patched, and deviations that share a predicate are expanded. The single deviation is the most surprising observation in the three designs.

% Appendix F: identifiability, delay and the value of waiting.
\FloatBarrier
\section{Identifiability, delay and the value of waiting}\label{app:identifiability}

This appendix derives the two limits on waiting behind the defer regions of Figures~\ref{fig:overview}C and~\ref{fig:phase}, a lower bound on the delay with which patch and expansion can be told apart and the value of the announced query in the one-shot protocol, and checks them on the solved environment.

\textbf{Identifiability and delay.} Let $L$ and $P$ be two specified worlds, a rule configuration and an exception configuration, that give the same law to every observation except the informative ones. A policy stops at a time $\sigma$ with a binary choice. Let $A$ be the event that it chooses expansion, with $\Pr_P(A)\le\alpha$, $\Pr_L(A)\ge1-\beta$ and $\alpha+\beta<1$. If, given any history, the expected divergence of the law of the next observation under $L$ from its law under $P$ is at most $\varrho I$, then
\begin{equation}\label{eq:F-delay}
\mathbb E_L[\sigma]\ge\frac{\operatorname{kl}(1-\beta,\alpha)}{\varrho I},\qquad \operatorname{kl}(a,b)=a\log\frac ab+(1-a)\log\frac{1-a}{1-b}.
\end{equation}
Informative observations that arrive independently with probability $\varrho$ and carry divergence $I$ each are the special case. The bound is a sequential change of measure \citep[Lemma~1]{kaufmann2016complexity}. By the chain rule, the divergence between the laws of the stopped data under $L$ and under $P$ is $\mathbb E_L\sum_{s\le\sigma}\operatorname{KL}\big(p_L(\cdot\mid Z_{<s})\,\Vert\,p_P(\cdot\mid Z_{<s})\big)$, which the per-step bound caps at $\varrho I\,\mathbb E_L[\sigma]$. The data-processing inequality applied to the event $A$ bounds it from below by $\operatorname{kl}(\Pr_L(A),\Pr_P(A))\ge\operatorname{kl}(1-\beta,\alpha)$, where the last step uses $\Pr_L(A)\ge1-\beta>\alpha\ge\Pr_P(A)$. With announced or non-stationary queries the cumulative conditional divergence takes the place of $\varrho I\,\mathbb E_L[\sigma]$. In SRE an informative observation is the first query of an input above $u$, where the rule and the matching exception explanation disagree. It carries $I=(1-2\varepsilon)\log\{(1-\varepsilon)/\varepsilon\}=2.05$ nats, and the repeat probability $r$ of the schedule sets its rate $\varrho$.

When $\varrho=0$ the two worlds cannot be told apart, and no policy meets $\alpha+\beta<1$ however long the history. A rule whose region is the single input 16 is such a case. It is observationally equivalent to an exception at 16, and the reference expands in 22\% of these episodes and patches in most of the others (Appendix~\ref{app:same-history}). The bound constrains only the decisions that must tell the worlds apart. When the price makes one action optimal in both worlds, that action is right without identification, and this is where prices and horizon enter.

\textbf{The value of the announced query.} In the one-shot protocol the announced query is an unscored probe whose label every predictor receives, the predictor of an immediate action included, and defer moves the decision to after the probe. Let $v_a(Y)$ be the value of committing to $a\in\{\mathrm{stay},\mathrm{patch},\mathrm{expand}\}$ after the probe label $Y$, price included. Then $V_{\mathrm{now}}=\max_a\mathbb E[v_a(Y)]$, $V_{\mathrm{defer}}=-c_o+\mathbb E[\max_a v_a(Y)]$, and the value of information \citep{howard1966information} is
\begin{equation}\label{eq:F-voi}
\mathrm{VOI}=V_{\mathrm{defer}}-V_{\mathrm{now}}+c_o=\mathbb E\big[\max_a v_a(Y)\big]-\max_a\mathbb E\big[v_a(Y)\big]\ge0,
\end{equation}
with equality exactly when one action is optimal for every label of positive probability. Defer, the one-shot form of a defer state (Appendix~\ref{app:normative}), is optimal exactly when $\mathrm{VOI}>c_o$. For two worlds with no other latent variable relevant to the decision, a probe on which both predict the same label distribution has likelihood ratio one and leaves every difference between actions unchanged, so its VOI is zero. Equal predictions of the two families do not suffice in general, because the label can still inform a parameter within a family and change the ranking of a third action. In the history of Section~\ref{sec:sre}, at $c_E=4$, $H=30$, $c_p=0.5$ and $c_o=0.1$ with input 16 announced, the values are stay $-15.62$, patch $-12.86$, expand $-13.32$ and defer $-12.53$. The best immediate action is patch, so $\mathrm{VOI}=-12.53+12.86+0.10=0.43$ and waiting is worth 0.33 net of $c_o$. With input 2 announced, defer equals expand minus $c_o$ exactly and VOI is zero.

\textbf{Relation to the identifiability of adaptation.} \citet{jhawar2026when} show that when two deployments give the evidence the same law and have different best adaptation actions, no selector chooses reliably. Their evidence is a static batch of unlabelled samples, and their reference action knows the deployment. The bound of Equation~\eqref{eq:F-delay} keeps their criterion with evidence that arrives sequentially: the rate $\varrho$ of informative observations sets the delay, the reference holds the learner's information, and prices and horizon decide whether the two worlds must be told apart at all.

\begin{figure}[htb]
  \centering
  \includegraphics[width=\textwidth]{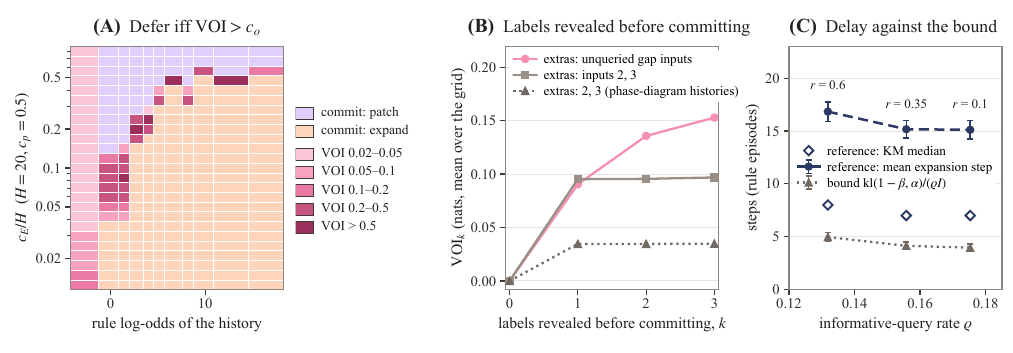}
  \caption{\textbf{The value of waiting.} (A) The one-shot grid of Figure~\ref{fig:phase} ($H=20$, $c_p=0.5$, informative announcement): cells where the announced label has value are shaded by its VOI, the others by the action committed to. At any $c_o$ the reference defers exactly where VOI exceeds $c_o$ (F1). (B) Mean VOI over the grid against the number $k$ of announced labels revealed before committing, for informative extra queries (the other unqueried inputs) and uninformative ones (inputs 2 and 3), the latter also on the histories of Figure~\ref{fig:phase}, which leave one input unqueried (F2). (C) The reference's mean and Kaplan--Meier median expansion step in \cls{rule} episodes and the bound of Equation~\eqref{eq:F-delay}, against the rate $\varrho$ of informative queries at repeat probabilities $r=0.1$, 0.35 and 0.6 (F3). Bars: one bootstrap standard error over episodes.}
  \label{fig:app-F}
\end{figure}

\textbf{Experiments.} Three experiments check these statements on the solved environment (Figure~\ref{fig:app-F}). F1 sweeps $c_o$ from 0 to 1 on the one-shot grid of Figure~\ref{fig:phase} (11 designed histories, 25 prices) and on the history of Section~\ref{sec:sre} ($H$ from 5 to 40), under an informative and an uninformative announcement. The cells where defer is optimal coincide with $\{\mathrm{VOI}>c_o\}$ at every $c_o$. Under the informative announcement the defer share of the grid falls from 19.6\% at $c_o=0$ to 9.1\% at $c_o=0.1$ and to zero at $c_o=1$. Under the uninformative one VOI is zero to numerical precision on the grid and below 0.005 nats on the single history, and no cell defers at a positive $c_o$. With the history fixed, the announced query alone opens and closes the defer region. F2 announces the next three queries as unscored probes, on histories with three unqueried inputs below the anomalous block, and lets the reference reveal the first $k$ labels at a price $k\,c_o$ before committing. When the extra probes are the other unqueried inputs, a second label adds 0.045 nats of VOI on average and a third 0.017. When they are inputs 2 and 3, they add less than 0.005 nats in every cell except those with five anomalous inputs, where these inputs are no longer neutral. The value of waiting grows only through queries that can still separate rule from exception.

F3 runs the sequential reference at $c_E=4$, $c_p=0.5$ and $T=40$ on 1,200 episodes for each repeat probability $r\in\{0.1,0.35,0.6\}$, with latent states and label noise held fixed so that only the schedule changes. The rate $\varrho$ is measured in \cls{rule} episodes before the expansion step, and $\alpha$ and $\beta$ are the reference's own error rates, its prior-weighted rate of expansion in the other classes and its rate of non-expansion in \cls{rule} episodes. Raising $r$ from 0.1 to 0.6 lowers $\varrho$ from 0.175 to 0.132 and raises the bound from 4.0 to 5.0 steps. The reference's share of defer states rises from 28.0\% to 31.2\% (4.5 standard errors) and its mean expansion step from 15.1 to 16.8 (1.4 standard errors), both in the direction of the bound. The level is far from the bound: the mean expansion step is 3.4 to 3.8 times the bound and its Kaplan--Meier median 1.6 to 1.8 times, and the latency counted from the first informative query has a median of 5 steps at every $r$. The reference maximizes utility, and its waiting is set by the prices and the horizon as much as by identifiability. The bound orders the regimes and does not predict the level.

\FloatBarrier
\section{Active-set transitions}\label{app:active-space}
An implicit learner has no action to regress. Its active set is read from $\kappa$ on null-equivalent pairs, against the family of curves indexed by the prior. A learner that stays below every curve able to account for its later behavior, and then rises, has expanded its active set. Its lift on a \cls{rule} episode is the first step at which its mean predicted probability of label 1 over the unqueried inputs above $u$ falls below one half (Section~\ref{sec:learners}). On \cls{rule} episodes and on threshold episodes matched to them in boundary and schedule, we take the difference in the learner's log-odds at unqueried inputs as a function of their position relative to the upper end $u$ of the interval. A response that decays with the distance from the observed anomalies and crosses $u$ without a step is local smoothing, and a response of equal size at all inputs is a shift in label frequency. The step at the boundary is the coefficient of $\mathbf 1[x>u]$ when this difference is regressed on the distance to the nearest reproduced anomalous input and on $\mathbf 1[x>u]$, over $u-2\le x\le u+3$. The response well below $u$ is the mean of this difference at the unqueried inputs at least three below $u$ and not below $l$, $l\le x\le u-3$, where the interval and the threshold at $l$ predict the same label. In Figure~\ref{fig:active-space} each language model is the mean over its three renderings, read with eight threshold-episode demonstrations in the context. For language models $\kappa$ is reported together with the calibration slope $s_0$, the slope of the model's log-odds on those of $E_0$ in \cls{noise} episodes, which separates a general flattening of predictions from a reduced weight on structural evidence. Latency is the first step at which the learner's predictions at unqueried inputs above $u$ have moved toward the rule, in excess of the same step for $E_1$. Surprise leakage is read from the updates: the change in the learner's log-odds after each observation is regressed on the change in the Bayes predictor's log-odds and on the surprise of that observation, and the Bayes predictor has leakage 0.

\textbf{Small Transformers are Bayes predictors under their training prior}, as memory-based meta-learning predicts \citep{ortega2019metalearning,mikulik2020metatrained}. The implicit learner trained at the generative base rate has $\kappa$ between 0.98 and 0.99 at every level of evidence, and its loss exceeds that of the Bayes predictor by 0.0005 nats per step. The learner trained with \cls{rule} episodes at a base rate of 0.02 starts at $\kappa=0.33$ with one anomalous input and reaches 0.95 with four. The Bayes predictor with a prior of 0.02 starts at 0.35 and reaches 1.00. The two agree in excess loss, in latency and in the share of \cls{rule} episodes in which the predictions never move toward the rule. The learner that never saw a rule has $\kappa=0$ and matches the Bayes predictor with a prior of zero in the same readouts. A curve of $\kappa$ that rises with evidence is therefore what a small prior produces, and by itself it is no sign that a family entered the computation late. Surprise leakage is within 0.007 of zero in all three learners.

The shift of the language models in Section~\ref{sec:respond} varies with the context as follows.
With demonstrations in the context the shift decays with the distance from the nearest anomalous input and passes the end of the interval smoothly, which is local smoothing. With the test episode alone it has nearly the same size at near and distant inputs, one model excepted, which is a shift in the frequency of the labels. In this setting the predictions also move toward the rule in 22\% to 40\% of the boundary-matched threshold episodes, against 2\% for the Bayes predictor.

\textbf{Demonstrations act through the statistics of positions and labels.} In the OLMo 3 Instruct chain, when the demonstrations share the label symbols of the test episode, the response at the inputs where the demonstrations almost always showed one label is half of the response elsewhere (0.13 to 0.36 against 0.47 to 0.67, as the slope of the model's shift on the Bayes shift in \cls{rule} episodes); in the T{\"u}lu 3 chain the response at those inputs falls to zero or reverses after preference tuning ($-0.55$ to 0.39). Moving the demonstrations to other positions removes most of the suppression, and giving them their own symbols removes it entirely.

\emph{The stages of OLMo 3 stay below every Bayes curve.} Read for each stage, context and rendering of OLMo 3, the matched-pair $\kappa$ of Section~\ref{sec:respond} with three anomalous inputs lies between 0.12 and 0.40 across the four stages and the contexts (Figure~\ref{fig:active-space}A), and in no rendering does it exceed 0.51. Every Bayes predictor with a prior of at least 0.005 is above 0.80 there.

\begin{figure}[h]
  \centering
  \includegraphics[width=\textwidth]{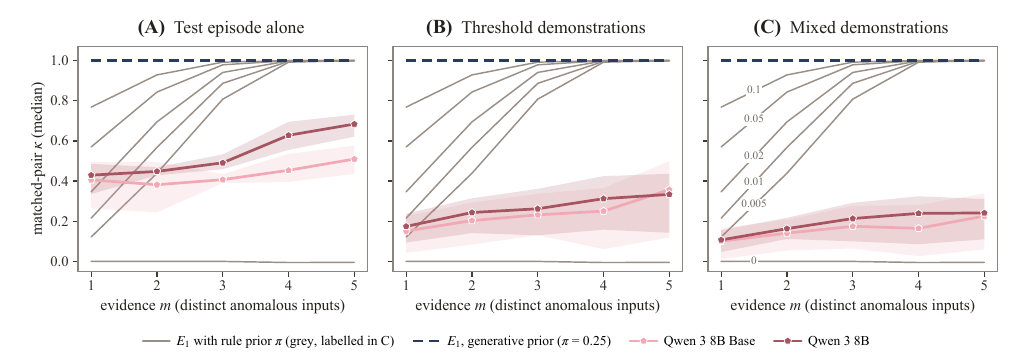}
  \caption{\textbf{The Qwen 3 pair against the family indexed by the prior.} Matched-pair $\kappa$ of Qwen 3 8B Base and Qwen 3 8B against the evidence level $m$, with the test episode alone (A), after eight threshold-episode demonstrations (B) and after eight demonstrations from the generative mixture (C), next to $E_1$ under \cls{rule} priors from 0 to 0.25 (grey lines for 0 to 0.1, labelled in C; navy and dashed for the generative prior 0.25). Lines are medians over the null-equivalent pairs averaged over the three renderings, and bands the range over the renderings.}
  \label{fig:pairk-qwen3}
\end{figure}

\textbf{The Qwen 3 pair.} Qwen 3 8B Base and its distilled final model, Qwen 3 8B \citep{yang2025qwen3}, add 18 curves of $\kappa$ in the three contexts and renderings (Figure~\ref{fig:pairk-qwen3}), and none of them follows a member of the family indexed by the prior. With demonstrations in the context, $\kappa$ with three anomalous inputs lies between 0.17 and 0.26. With the test episode alone it is 0.41 for the base model and 0.49 for the final model, the highest value of the 16 models, and the final model reaches 0.68 with five anomalous inputs, still off every member of the family. The response has no step at the boundary (0.05 to 0.32 in log-odds, every interval including zero) and extends well below $u$, at $x\le u-3$ (0.65 to 2.65, every interval excluding zero). As in the other language models, it is local smoothing with demonstrations in the context and a shift in the frequency of the labels with the test episode alone. Surprise leakage is 0.12 to 0.14.

\FloatBarrier
\section{Learned policies: regret at the default cell and results beyond the main text}\label{app:learned-policies}\label{app:table1}
This appendix collects the definitions behind the regret decomposition of Appendix~\ref{app:reference}, the regret of every policy at the default cell (Tables~\ref{tab:regret} and~\ref{tab:regret-rest}) and further results on the explicit learners of Section~\ref{sec:value}.

Structural actions leave the queries and labels that follow unchanged (Section~\ref{sec:sre}).
The utility of any action sequence on a given episode is therefore a deterministic function of that episode, and the regret of a learner is computed exactly, episode by episode.

\textbf{Latency and expansion errors.} In an episode whose structure lies outside $\mathcal F_0$, let $t_0$ be the first step at which an informative observation arrives; the latency is $\inf\{t\ge t_0:a_t=\mathrm{expand}\}-t_0$. Exact Bayesian inference itself switches late, because the richer family must first overcome its larger description length \citep{vanerven2012catching}. Latency is therefore measured as excess over the reference policy on the same episodes, the learner's Kaplan--Meier median latency minus the reference's, with episodes that never expand treated as censored. An expansion is an error when $\Delta Q^{\ast}(s_t)<0$, whatever the class of the episode. For implicit learners the lift takes the place of $a_t$.

\emph{The price response ends with the trained range.} At the two prices outside the range of training, price sensitivity falls to between 0.43 and 0.58 under every training regime. At a price below the range the learners expand less than at the lowest trained price, and the reference expands more. Episodes longer than those trained on keep the ratios.

\emph{Forming the evidence is not the limit.} Giving a learner the posterior of the reference as an input changes neither its regret nor its ratios.

\emph{Imitation places the boundary where the labels reach.} A learner that imitates the reference along the reference's own path sees no state beyond the reference's expansion. In those states it expands at half of the reference's rate, and the point at which it expands with probability one half lies 0.22 nats beyond the reference's boundary. With labels for those states the distance falls to 0.03 nats and the rate of expansion beyond the boundary rises to 0.34, against 0.41 for the reference.

\emph{Where the learners lose.} Their regret comes from \cls{exception} episodes, in which they patch less often than the reference: at the default cell the probability of patching without expanding is 0.09 for the learner trained on utility alone, 0.21 for utility after imitation and 0.30 to 0.31 for the imitation learners, against 0.41 for the reference (Table~\ref{tab:regret}). On matched failures the imitation learners expand at the rates of the reference, in 0.70 to 0.77 of the \cls{rule} episodes against 0.76 and in 0.03 to 0.05 of the threshold episodes against 0.03. The learners trained on utility expand in 0.03 to 0.09 of the threshold episodes. Given the same number of anomalous observations, every learner expands more when the anomalies are spread over inputs than when they sit on one, as the reference does.

\emph{Waiting is learned only in part.} On histories in which the reference waits for an informative query and acts when the announced query is a repeat of an anomaly, the learners stay more often under the informative announcement than on matched controls, by 0.10 to 0.22 in the probability of staying, against 1 for the reference, and by similar amounts whether the reference's alternative is a patch or an expansion. The price and the horizon enter these learners' decisions as they enter the reference's. The value of the next observation mostly does not.

% Table 2 takes the top of the next page.
\begin{table}[h]
\caption{Meta-decision regret against the reference at the default cell. $T=40$, $c_E=4$, $c_p=0.5$ (a held-out $(c_E,T)$ cell of the small Transformers' training split); nats per episode, prior-weighted over the four classes (\cls{noise} 0.40, \cls{exception} 0.20, \cls{drift} 0.15, \cls{rule} 0.25); regret $=\mathcal U(\text{reference})-\mathcal U(\text{policy})$ on the same 1,200 episodes (300 per class). The two e-process rows are one policy at this cell, where the fixed threshold was tuned, and differ only in the grid mean. Small Transformers: mean over three seeds; their SE is the per-run paired SE averaged over the seeds. BC (behavior cloning) imitates the reference's action, POLICY-ONLY is BC with the reference's posterior as input, and REWARD is trained on utility alone (Appendix~\ref{app:training}). Path labels: reference actions on the states beyond the expansion. Latency: Kaplan--Meier median of the expansion latency in \cls{rule} episodes (Appendix~\ref{app:statistics}) and the censored share. Patch only: the probability of patching without expanding in \cls{exception} episodes, which was not recorded for the bounded reference. Grid mean: regret averaged over the 120 price cells.}
\label{tab:regret}
\centering\tablestyle\setlength{\tabcolsep}{3.2pt}
\setcounter{tabcols}{10}
\begin{tabularx}{\textwidth}{>{\raggedright\arraybackslash}X
  S[table-format=1.3,table-space-text-post=\unc{0.031}]
  S[table-format=+1.3] S[table-format=+1.3] S[table-format=+1.3] S[table-format=+1.3]
  S[table-format=1.1] S[table-format=1.2]
  S[table-format=1.2]
  S[table-format=1.3]}
\toprule
 & & \multicolumn{4}{c}{\hd{regret by class}} & \multicolumn{2}{c}{\hd{expansion latency}} & & \\
\cmidrule(lr){3-6}\cmidrule(lr){7-8}
\hd{policy} & {\hd{regret\unc{SE}}} & {\hd{\cls{noise}}} & {\hd{\cls{exception}}} & {\hd{\cls{drift}}} & {\hd{\cls{rule}}} & {\hd{median}} & {\hd{censored}} & {\hdtwo*{patch}{only}} & {\hdtwo*{grid}{mean}} \\
\midrule
\grouprow{Reference policies}
Reference policy & 0.00 {\unc{0.00}} & +0.00 & +0.00 & +0.00 & +0.00 & 5 & 0.25 & 0.41 & 0.00 \\
\rowcolor{rowgray}Bounded reference & 0.49 {\unc{0.06}} & +0.05 & +0.45 & +0.08 & +1.48 & 6 & 0.27 & {\na} & 0.24 \\
\groupsep
\grouprow{Triggers and fixed policies}
e-process trigger, fixed threshold & 0.68 {\unc{0.06}} & -0.12 & +1.12 & +0.69 & +1.62 & 5 & 0.19 & 0.00 & 1.19 \\
\rowcolor{rowgray}e-process trigger, re-tuned per cell & 0.68 {\unc{0.06}} & -0.12 & +1.12 & +0.69 & +1.62 & 5 & 0.19 & 0.00 & 0.42 \\
never revise & 2.24 {\unc{0.14}} & -0.58 & +0.55 & -0.58 & +9.81 & {$\infty$} & 1.00 & 0.00 & 1.29 \\
\groupsep
\grouprow{Small Transformers (mean of three seeds)}
BC on the reference's path & 0.060 {\unc{0.031}} & -0.069 & +0.293 & -0.129 & +0.193 & 5.3 & 0.30 & 0.31 & 0.049 \\
\rowcolor{rowgray}BC with path labels & 0.039 {\unc{0.027}} & -0.033 & +0.334 & -0.084 & -0.008 & 4.7 & 0.26 & 0.30 & 0.039 \\
POLICY-ONLY (exact posterior as input) & 0.053 {\unc{0.029}} & -0.093 & +0.433 & -0.147 & +0.102 & 5.3 & 0.28 & 0.31 & 0.044 \\
\rowcolor{rowgray}REWARD (utility alone) & 0.210 {\unc{0.041}} & +0.089 & +0.800 & +0.068 & +0.015 & 3.7 & 0.26 & 0.09 & 0.089 \\
REWARD from BC (utility after imitation) & 0.081 {\unc{0.035}} & -0.037 & +0.392 & -0.061 & +0.108 & 6.0 & 0.31 & 0.21 & 0.038 \\
\bottomrule
\end{tabularx}
\end{table}

\begin{table}[h]
\caption{Meta-decision regret against the reference at the default cell: rows and columns not shown in Table~\ref{tab:regret}. Top: the baselines left out of Table~\ref{tab:regret}, with its columns; the grid mean is given with the threshold fixed and with the threshold re-tuned in every cell. Bottom: the remaining columns, for every row. Expands in: the share of the 300 \cls{noise}, \cls{drift} and \cls{exception} episodes in which the policy expands. Excess latency: the Kaplan--Meier median latency in \cls{rule} episodes minus the reference's, in steps. Type mismatch: the share of episodes, prior-weighted over the classes, in which the type of revision (none, patch only or expansion) differs from the reference's. Seed s.d.: the standard deviation of the regret over the three seeds, in nats per episode.}
\label{tab:regret-rest}
\centering\tablestyle\setlength{\tabcolsep}{3.2pt}
\setcounter{tabcols}{11}
\begin{tabularx}{\textwidth}{>{\raggedright\arraybackslash}X
  S[table-format=1.2,table-space-text-post=\unc{0.13}]
  S[table-format=+1.2] S[table-format=+1.2] S[table-format=+1.2] S[table-format=+1.2]
  S[table-format=2.0] S[table-format=1.2]
  S[table-format=1.2]
  S[table-format=1.2] S[table-format=1.2]}
\toprule
 & & \multicolumn{4}{c}{\hd{regret by class}} & \multicolumn{2}{c}{\hd{expansion latency}} & & \multicolumn{2}{c}{\hd{grid mean}} \\
\cmidrule(lr){3-6}\cmidrule(lr){7-8}\cmidrule(lr){10-11}
\hd{policy} & {\hd{regret\unc{SE}}} & {\hd{\cls{noise}}} & {\hd{\cls{exception}}} & {\hd{\cls{drift}}} & {\hd{\cls{rule}}} & {\hd{median}} & {\hd{censored}} & {\hdtwo*{patch}{only}} & {\hd{fixed}} & {\hd{re-tuned}} \\
\midrule
\grouprow{Triggers and fixed policies}
likelihood trigger, fixed threshold & 0.81 {\unc{0.08}} & -0.09 & +0.82 & +0.02 & +2.74 & 8 & 0.42 & 0.00 & 1.09 & 0.43 \\
\rowcolor{rowgray}count trigger, fixed threshold & 1.64 {\unc{0.08}} & -0.42 & +1.18 & -0.13 & +6.36 & 25 & 0.30 & 0.00 & 1.29 & 0.62 \\
surprise trigger, fixed threshold & 2.03 {\unc{0.08}} & +2.04 & +2.23 & +2.12 & +1.80 & 4 & 0.00 & 0.00 & 2.31 & 0.63 \\
\rowcolor{rowgray}always patch an anomalous input & 2.18 {\unc{0.13}} & +0.97 & -0.23 & +0.88 & +6.81 & {$\infty$} & 1.00 & 0.92 & {\na} & {\na} \\
\bottomrule
\end{tabularx}

\bigskip
\setlength{\tabcolsep}{4pt}
\setcounter{tabcols}{7}
\begin{tabularx}{\textwidth}{>{\raggedright\arraybackslash}X
  S[table-format=1.2,table-column-width=38pt] S[table-format=1.2,table-column-width=38pt]
  S[table-format=1.2,table-column-width=38pt]
  S[table-format=+2.1]
  S[table-format=1.2]
  S[table-format=1.3]}
\toprule
 & \multicolumn{3}{c}{\hd{expands in}} & & & \\
\cmidrule(lr){2-4}
\hd{policy} & {\hd{\cls{noise}}} & {\hd{\cls{drift}}} & {\hd{\cls{exception}}} & {\hdtwo*{excess}{latency}} & {\hdtwo*{type}{mismatch}} & {\hdtwo*{seed}{s.d.}} \\
\midrule
\grouprow{Reference policies}
Reference policy & 0.07 & 0.06 & 0.13 & +0 & 0.00 & {\na} \\
\rowcolor{rowgray}Bounded reference & 0.12 & 0.12 & 0.27 & +1 & 0.23 & {\na} \\
\groupsep
\grouprow{Triggers and fixed policies}
e-process trigger, fixed threshold & 0.10 & 0.29 & 0.35 & +0 & 0.28 & {\na} \\
\rowcolor{rowgray}e-process trigger, re-tuned per cell & 0.10 & 0.29 & 0.35 & +0 & 0.28 & {\na} \\
likelihood trigger, fixed threshold & 0.11 & 0.13 & 0.20 & +3 & 0.33 & {\na} \\
\rowcolor{rowgray}count trigger, fixed threshold & 0.04 & 0.11 & 0.23 & +20 & 0.31 & {\na} \\
surprise trigger, fixed threshold & 0.62 & 0.63 & 0.73 & -1 & 0.54 & {\na} \\
\rowcolor{rowgray}never revise & 0.00 & 0.00 & 0.00 & {\na} & 0.44 & {\na} \\
always patch an anomalous input & 0.00 & 0.00 & 0.00 & {\na} & 0.67 & {\na} \\
\groupsep
\grouprow{Small Transformers (mean of three seeds)}
BC on the reference's path & 0.05 & 0.05 & 0.10 & +0.3 & 0.10 & 0.032 \\
\rowcolor{rowgray}BC with path labels & 0.06 & 0.06 & 0.13 & -0.3 & 0.09 & 0.034 \\
POLICY-ONLY (exact posterior as input) & 0.05 & 0.04 & 0.12 & +0.3 & 0.09 & 0.008 \\
\rowcolor{rowgray}REWARD (utility alone) & 0.11 & 0.11 & 0.21 & -1.3 & 0.17 & 0.046 \\
REWARD from BC (utility after imitation) & 0.08 & 0.08 & 0.14 & +1.0 & 0.13 & 0.029 \\
\bottomrule
\end{tabularx}
\end{table}

\section{Post-training}\label{app:posttraining}
\begin{figure}[h]
  \centering
  \includegraphics[width=\textwidth]{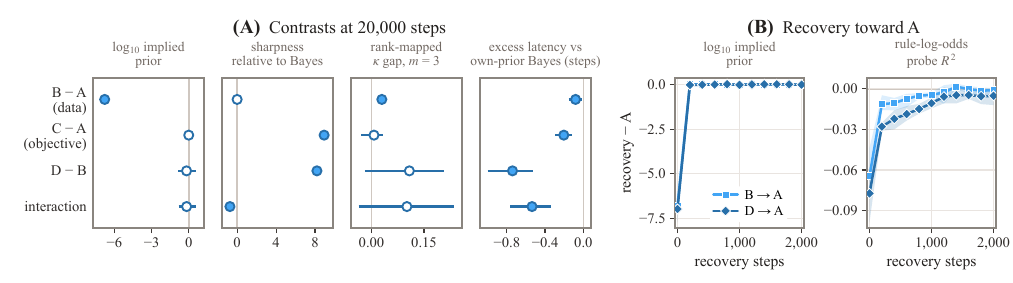}
  \caption{\textbf{Post-training separates the prior from predictive sharpness.} (A) Contrasts from the $2\times2$ post-training intervention after 20,000 steps. Points are means over three learners; bars are 95\% bootstrap intervals, with filled markers excluding zero. $\kappa$ is rank-mapped to remove sharpening. (B) Recovery of the implied prior and rule-log-odds decodability after restoring the original data and objective: means over the three learners, with bands for their 95\% bootstrap intervals.}
  \label{fig:post-training}
\end{figure}

\begin{figure}[h]
  \centering
  \includegraphics[width=\textwidth]{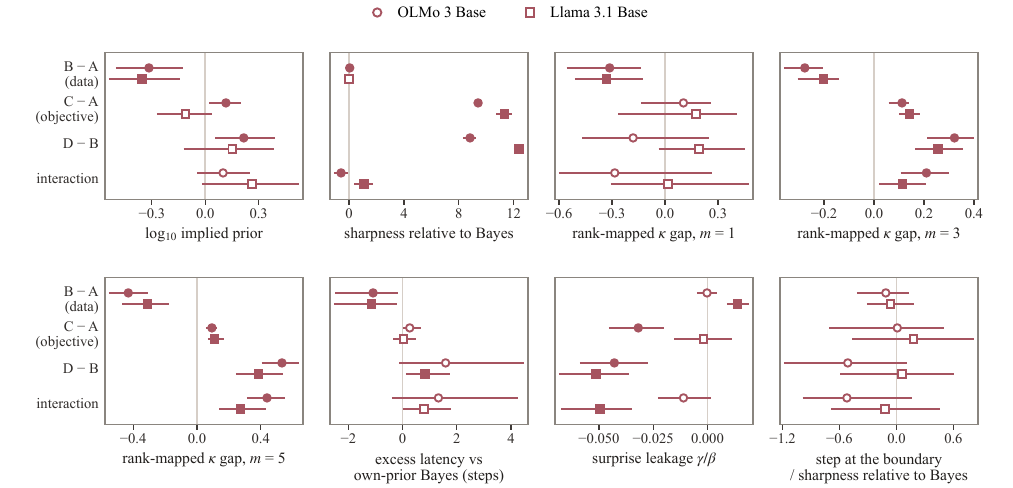}
  \caption{\textbf{Adapter branches on two base models.} Paired contrasts of the four branches on OLMo 3 Base (circles) and Llama 3.1 Base (squares) without demonstrations, with the readouts of Figure~\ref{fig:post-training}A, the $\kappa$ gap also at one and five anomalous inputs, surprise leakage and the step at the boundary divided by the sharpness. Filled markers are contrasts whose interval excludes zero; bars are 95\% intervals of one episode bootstrap shared by all models.}
  \label{fig:lora}
\end{figure}

This appendix collects the stage-wise differences among the language models, in their decisions (Section~\ref{sec:llm-decision}) and in their predictions (Section~\ref{sec:respond}), the controlled post-training contrasts of Section~\ref{sec:posttrain} (Figure~\ref{fig:post-training}) and their analogue with adapters on two language models (Figure~\ref{fig:lora}), and the related work on post-trained lineages and on scoring revisions.

\emph{Stages.} In both Instruct lineages preference tuning raises the propensity to defer and the response to the stated gain, and in OLMo 3 both changes hold across renderings and action orders. Reinforcement learning with verifiable rewards lowers the propensity to defer in OLMo 3 and leaves it nearly unchanged in T{\"u}lu 3. At the history level the stages differ only in their constants, and the order in which the actions are listed moves those constants more than the rendering does.

What the stages change is the size of the response and the fidelity in context, and which stage changes them depends on the recipe. In the OLMo 3 Instruct chain and in T{\"u}lu 3, preference tuning worsens the fit to threshold episodes, lowers the accuracy at unqueried inputs far from the boundary and delays the lift. In the Think chain these changes come with supervised fine-tuning, and preference tuning changes nothing. Two recipes on the same base show where the changes come from: Llama 3.1 Instruct differs from its base in five small readouts, and T{\"u}lu 3, built on that base, differs from Llama 3.1 Instruct in 29.

\textbf{Recipes.} Across public checkpoints, the change from the base model to the final model can be compared between four recipes: supervised fine-tuning, preference tuning and reinforcement learning with verifiable rewards in OLMo 3 Instruct, T{\"u}lu 3 and OLMo 3 Think, and distillation in Qwen 3. A change is counted when it has the same sign in the three renderings, exceeds the spread of the readout across them and has a 95\% interval that excludes zero in each of them. From base to final model, OLMo 3 Instruct shows 10 such changes, T{\"u}lu 3 28, OLMo 3 Think 9 and Qwen 3 7. The main change in Qwen 3 is a steeper calibration with the test episode alone, from 0.95 to 1.32 times the slope of the Bayes predictor, which T{\"u}lu 3 and OLMo 3 Think share and OLMo 3 Instruct does not. The degradation that the three recipes with preference tuning share is absent from the distilled model: the accuracy far from the boundary changes by $-0.02$ against $-0.09$ to $-0.15$, the generalization to unqueried inputs of the region under demonstrations by $+0.02$ against $-0.04$ to $-0.18$, and the false lift by $+0.05$, short of the criterion, against $+0.10$ and $+0.17$ in OLMo 3 Instruct and T{\"u}lu 3. None of the ten changes of OLMo 3 Instruct recurs in Qwen 3, against five in T{\"u}lu 3 and three in OLMo 3 Think. In the decisions, distillation lowers the probability of expanding at the history level by 0.14 and raises the response to the stated gain by 0.65, both changes holding across renderings and action orders, as the recipes of OLMo 3 Instruct and T{\"u}lu 3 do, while OLMo 3 Think moves both the other way. The rise in the probability of deferring that marks preference tuning in both Instruct lineages ($+0.46$ in OLMo 3, $+0.26$ in T{\"u}lu 3) does not appear under distillation ($-0.02$, with a sign that changes across the six combinations of rendering and action order), and the final model moves its mass to patching instead.

\emph{Readouts of the controlled branches.} The four branches are A (the generative data, log score), B (no \cls{rule} episodes, log score), C (the generative data, reward) and D (no \cls{rule} episodes, reward), and the contrasts are B $-$ A for the data, C $-$ A for the objective, D $-$ B for the objective on rule-free data, and their interaction. Every readout is compared with the Bayes predictor of the prior that the learner holds after post-training, estimated on matched pairs, and $\kappa$ is computed after the learner's log-odds are mapped onto that predictor's at equal quantiles, so that a sharpening of the predictions cannot enter it. The recovery runs continue B and D for 2,000 steps with the data and objective of A, and Figure~\ref{fig:post-training}B gives their readouts as differences from A at 20,000 steps.

\emph{The rule log-odds stay decodable.} In each branch a ridge probe reads the reference's rule log-odds from the residual stream at the positions that predict a label (Appendix~\ref{app:mechanism}). Its $R^2$ at the best layer falls from 0.90 in branch A to 0.84 without \cls{rule} episodes and to 0.88 under the reward objective.

\textbf{Adapter branches on two base models.} The two-by-two of the small Transformers was applied with rank-16 LoRA adapters \citep{hu2022lora} to OLMo 3 Base and Llama 3.1 Base: data with or without \cls{rule} episodes, and the log score or the expected 0/1 reward, with 16,384 episodes in 512 steps per branch and one seed (Appendix~\ref{app:training}). The branches are read in the implicit protocol without demonstrations, against the Bayes predictor of the prior each branch holds, with 95\% intervals from one episode bootstrap shared by all models, so that every contrast is paired (Figure~\ref{fig:lora}).

Removing \cls{rule} episodes halves the implied base rate on both bases, from 0.35 to 0.17 and 0.16, where the small Transformers lost nearly seven decades; at this dose that collapse does not recur. It leaves the sharpness unchanged (B $-$ A on the sharpness ratio $+0.04$ and $-0.01$). The reward objective multiplies the sharpness by about ten and twelve (C $-$ A $+9.4$ and $+11.3$ on the ratio) and moves the base rate by about a third of the data's effect, in opposite directions on the two bases ($+0.12$ and $-0.11$ in $\log_{10}$). The separation of the two axes therefore holds for Llama 3.1 under the criterion fixed in advance, which asks the objective's effect on the base rate to be below a fifth of the data's or its interval to include zero, and narrowly fails for OLMo 3 (interval \ci{0.02}{0.20}). Two readouts differ from the small Transformers. The rule-free branch falls further below the $\kappa$ curve of its own prior than the control at every level of evidence, by 0.2 to 0.4 in rank-mapped $\kappa$, while the scaled Bayes predictor fits it nearly as well as the control (0.032 and 0.031 against 0.024 and 0.030 nats per cell); the reward objective on rule-free data restores the response (D $-$ B about $+0.3$ at three and $+0.4$ to $+0.5$ at five anomalous inputs). And the reward objective lowers surprise leakage here, by 0.03 to 0.05 in three of the four contrasts (0.00 for C $-$ A on Llama 3.1), where it raised it in the small Transformers. Divided by the sharpness, the step at the boundary does not move under any contrast. No criterion shift appears in either base: every branch lifts no later than the Bayes predictor of its own prior (mean latency $-0.4$ to $-2.1$ steps, Kaplan--Meier median excess 0). With demonstrations in the context the implied base rate of every model, the bases included, falls to $10^{-3.6}$ or below, so the data contrast is not identified there, and the sharpening is about the same (C $-$ A $+8.0$ and $+11.3$).

Against the public lineages, the controlled rule-free data reproduce none of the markers of the preference-tuning stage in OLMo 3 Instruct and T{\"u}lu 3 (in-family excess log loss $+0.035$ and $+0.12$, accuracy far from the boundary $-0.07$ and $-0.12$, a later lift by 1.8 and 2.2 lines): B $-$ A changes the first two by at most 0.01 and delays the lift by at most 0.24 steps. The reward objective reproduces the drop in accuracy at a similar size ($-0.04$ and $-0.08$) and a rise in in-family log loss 26 to 58 times larger, which comes from the sharpening, and not the later lift.

\textbf{Post-training.} Stage-wise analyses of public lineages exist for OLMo 3 \citep{nguyen2026measuring} and T{\"u}lu 3 \citep{ziaei2026limits}, with a prescribed probability and a human response distribution as references. Against Bayesian references, instruction-tuned models forget faster after a change point in two of three model families \citep{gupta2025enough}, and instruction tuning of Llama 3.1 8B lowers the prior of a behavior and leaves its in-context learnability unchanged \citep{arora2025bayesian}. We take the Bayes-optimal revision policy as the reference and add controls matched for base rate, which separate a rational update of the prior from a shift of the decision criterion.

Where actions change the stream, a decision to keep or revise is scored by the outcome realized afterwards, through a forked run \citep{li2026measuring}, a hindsight reference \citep{kashanimotlagh2026return} or a search over intervention schedules \citep{chadha2026beyond}.

% Appendix J: detection and commitment.
\FloatBarrier
\section{Detection and commitment}\label{app:detection-commitment}

This appendix separates the decision to search the richer family, whose timing the latency readouts of Appendix~\ref{app:table1} and Section~\ref{sec:posttrain} measure, from the commitment to one of its rules, which a larger library can delay by itself.

\textbf{Invariance of the search decision.} The bounded reference decides on its count state (Appendix~\ref{app:reference}), which contains no index of a rule and no size of the library. Its backward induction depends on the library only through the laws of the count statistics and the expected gains of expanding. A larger library that leaves these unchanged, with the same base rate of the \cls{rule} class, the same distribution of regions and the same prices, leaves every step of the recursion and therefore every decision unchanged. Commitment to one rule depends on the library directly. Under a uniform prior over $|\mathcal R|$ rules, the posterior of the true rule $j$ exceeds $1-e$ only when $\sum_{j'\ne j}e^{-\Lambda_{jj'}}\le e/(1-e)$, with $\Lambda_{jj'}$ its log-likelihood ratio against rule $j'$. Commitment therefore costs about the logarithm of the number of rules that the data have not yet excluded, $\log|\mathcal R|$ nats when all of them remain plausible. A learner with a richer library should decide to search no later and may commit later, and a delay of commitment says nothing by itself about the criterion for search. The readouts of Section~\ref{sec:posttrain} concern search: $\kappa$ and the probe of the rule log-odds read the evidence for the \cls{rule} class, and neither identifies a rule.

\begin{figure}[htb]
  \centering
  \includegraphics[width=\textwidth]{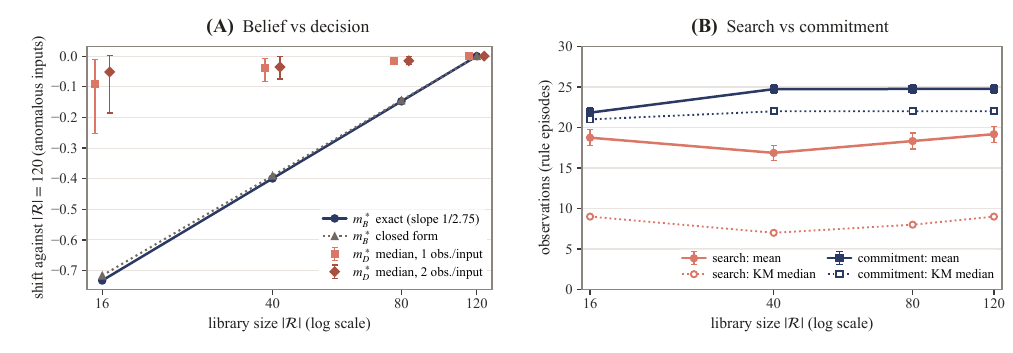}
  \caption{\textbf{Detection and commitment.} (A) Shift of $m^\ast_B$, exact and in closed form, and of $m^\ast_D$ (median and 10th to 90th percentile over prices, one or two observations per anomalous input) against the size of the rule library, relative to $|\mathcal R|=120$ (J1). (B) Search delay, the reference's expansion step, and commitment delay, the step at which its posterior puts more than half of the mass of the \cls{rule} class on the true rule, in 300 \cls{rule} episodes whose rules belong to every library: means with one bootstrap standard error, and Kaplan--Meier medians (J2).}
  \label{fig:app-J}
\end{figure}

\textbf{Experiment J1.} J1 measures how the size of the library moves $m^\ast_B$ and $m^\ast_D$ of the full reference (Figure~\ref{fig:app-J}A). It cuts the interval library to its $|\mathcal R|=16$, 40, 80 or 120 widest intervals, with the prior mass of the \cls{rule} class kept at 0.25, and solves designed histories whose matching rules belong to every library (base threshold 1, up to five anomalous top inputs, observed once or twice) in the one-shot protocol at $H=20$, $c_p=0.5$ and $c_o=0.1$. The exact $m^\ast_B$ is $-0.231$, 0.102, 0.354 and 0.501 at $|\mathcal R|=16$, 40, 80 and 120, a shift of exactly $-\Delta\log|\mathcal R|/2.7515$. By Equation~\eqref{eq:C-config} the library enters the configuration odds only through $\Delta L=\log(|\mathcal R|/16)$, and at $|\mathcal R|=16$ a rule costs no more than the threshold it replaces. By contrast, $m^\ast_D$ moves by a median over prices of $-0.09$, $-0.04$ and $-0.015$ at $|\mathcal R|=16$, 40 and 80 with one observation per input ($-0.05$, $-0.035$ and $-0.015$ with two), 7\% to 13\% of the shift of $m^\ast_B$. The library moves belief, and the decision stays where the region puts it: once the posterior of the \cls{rule} class has saturated, Equation~\eqref{eq:C-boundary} depends on the inputs that a rule predicts and a slot does not cover, and the size of the library does not change them.

\textbf{Experiment J2.} J2 runs the sequential reference at $c_E=4$, $c_p=0.5$ and $T=40$ on 300 \cls{rule} episodes whose rules are drawn from the 16-interval library, the same episodes under every library, and compares the search delay with the commitment delay (Figure~\ref{fig:app-J}B). The mean search delay shows no trend in $\log|\mathcal R|$ (slope $+0.25\pm0.30$ steps per unit, where $\pm$ is one bootstrap standard error over episodes, 95\% interval \ci{-0.34}{0.88}), and its differences between libraries, at most 2.3 steps, do not follow the size of the library. The commitment delay grows with slope $+1.42\pm0.19$, and all of its growth lies between 16 and 40 rules ($+3.2\pm0.4$ per unit, against $+0.02\pm0.01$ from 40 to 120). The narrower intervals added beyond 40 are never confused with the wide true rules: the commitment step changes in 4 of the 300 episodes between 40 and 120 rules and in none between 80 and 120. When both events occur, the reference expands a median of 13 to 17 steps before it commits to the rule. Detection and commitment are decoupled, and commitment grows with $\log|\mathcal R|$ only while the added rules remain close to the truth, as the count of rules not yet excluded predicts. The bounded reference is invariant by construction.

\FloatBarrier
\section{Architectures}\label{app:architectures}

This appendix reads the finding of Section~\ref{sec:respond} and Appendix~\ref{app:active-space}, that the small Transformers are Bayes predictors under their training prior, across architectures, and asks whether a learner with a state of fixed size must read the history again after it lifts.

Two fixed-state learners take the place of the attention in the implicit Transformer: a GRU \citep{cho2014learning}, and a selective state-space model in the Mamba-2 form \citep{dao2024transformers}, both written in plain PyTorch. They match the Transformer in parameters, about 4.8 million, in data stream, optimizer and number of steps, and are trained at the generative base rate with three seeds each (Appendix~\ref{app:training}). Recovery is read on \cls{rule} episodes aligned at each learner's own lift, as the excess log loss over the Bayes predictor at the $k$-th prediction after the lift, with $E_1$ aligned at its own lift as the floor.

\begin{table}[h]
\caption{Architectures at the generative base rate: mean ($\pm$~s.d.) over three seeds. The Bayes predictor is $E_1$, the Bayes predictor of the training prior of all three learners (intervals: 95\% over episodes). Class-balanced: the unweighted mean over the four classes. Slope inside the region: the coefficient of the distance to the nearest reproduced anomalous input in a regression of the shift of Section~\ref{sec:respond} over the unqueried inputs above $u$, in log-odds per input. KM: Kaplan--Meier; lift latencies count the observations after the first informative one.}
\label{tab:architectures}
\centering\tablestyle
\setcounter{tabcols}{5}
\begin{tabularx}{\textwidth}{>{\raggedright\arraybackslash}X
  S[table-format=-1.4,table-space-text-post=\unc{0.0012}]
  S[table-format=-1.4,table-space-text-post=\unc{0.0012}]
  S[table-format=-1.4,table-space-text-post=\unc{0.0012}]
  S[table-format=-1.3,table-space-text-post=\ci{-0.829}{-0.460}]}
\toprule
\hd{readout} & {\hd{Transformer}} & {\hd{GRU}} & {\hd{selective SSM}} & {\hd{Bayes predictor}} \\
\midrule
\grouprow{Excess log loss over $E_1$ ($10^{-3}$ nats per step)}
class-balanced & 0.47 {\unc{0.14}} & 0.48 {\unc{0.10}} & 0.46 {\unc{0.15}} & 0 \\
\rowcolor{rowgray}\cls{rule} episodes & 0.66 {\unc{0.38}} & 0.68 {\unc{0.50}} & 0.54 {\unc{0.43}} & 0 \\
\groupsep
\grouprow{Matched-pair $\kappa$}
$m=1$ & 0.994 {\unc{0.021}} & 0.990 {\unc{0.013}} & 0.995 {\unc{0.011}} & 1.000 \\
\rowcolor{rowgray}$m=3$ & 0.983 {\unc{0.004}} & 0.987 {\unc{0.001}} & 0.985 {\unc{0.007}} & 1.000 \\
$m=5$ & 0.986 {\unc{0.003}} & 0.984 {\unc{0.003}} & 0.984 {\unc{0.011}} & 1.000 \\
\groupsep
\grouprow{Response at the boundary}
step at the boundary (log-odds) & 1.110 {\unc{0.014}} & 1.105 {\unc{0.021}} & 1.114 {\unc{0.027}} & 1.122 {\ci{0.877}{1.369}} \\
\rowcolor{rowgray}slope inside the region & -0.623 {\unc{0.012}} & -0.624 {\unc{0.004}} & -0.621 {\unc{0.001}} & -0.616 {\ci{-0.829}{-0.460}} \\
\groupsep
\grouprow{Leakage and lift}
surprise leakage & 0.0012 {\unc{0.0008}} & 0.0013 {\unc{0.0012}} & 0.0010 {\unc{0.0009}} & 0 \\
\rowcolor{rowgray}lift latency, KM median (observations) & 0.0 {\unc{0.0}} & 0.0 {\unc{0.0}} & 0.0 {\unc{0.0}} & 0 \\
share without a lift where $E_1$ lifts & 0.007 {\unc{0.003}} & 0.006 {\unc{0.003}} & 0.006 {\unc{0.003}} & 0 \\
\bottomrule
\end{tabularx}
\end{table}

\begin{figure}[h]
  \centering
  \includegraphics[width=\textwidth]{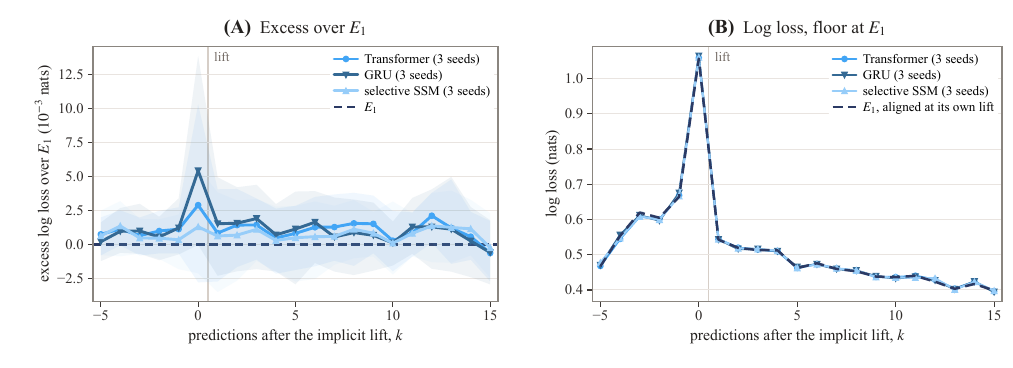}
  \caption{\textbf{Recovery after the lift.} (A) Excess log loss over $E_1$ of the Transformer, the GRU and the selective state-space learner at the $k$-th prediction after each learner's own lift on \cls{rule} episodes (the implicit lift of the axis: the first step at which its mean predicted probability of label 1 over the unqueried inputs above $u$ falls below one half), $k=0$ being the prediction of the anomalous observation that triggers it (mean over three seeds; bands span the seeds' 95\% intervals over episodes). (B) The log loss itself, against $E_1$ aligned at its own lift.}
  \label{fig:recovery}
\end{figure}

The three architectures agree within the spread over seeds on every readout (Table~\ref{tab:architectures}): the step at the boundary is 1.105 to 1.114 against 1.122 for the Bayes predictor, matched-pair $\kappa$ lies between 0.977 and 0.995 at every level of evidence, and surprise leakage is about 0.001. None of them recovers faster than the others after the lift, and none accumulates loss again: from the first prediction after the lift each stays within about $2\times10^{-3}$ nats of the Bayes predictor (Figure~\ref{fig:recovery}). At this scale a fixed state carries what the full model needs, the label counts at each input and the order in which they arrived, and no retroactive reinterpretation of the history appears. A test that could separate the architectures needs a state smaller than these statistics, or a fixed-state learner trained without \cls{rule} episodes.

\FloatBarrier
\section{Mechanism: probes and steering}\label{app:mechanism}

This appendix tests by intervention the reading of Section~\ref{sec:posttrain} that what post-training removes is suppressed rather than lost, and whether probes separate a family that is computed and gated out from one that is never computed, two cases that $\kappa$ cannot tell apart.

Ridge probes read the reference's rule log-odds and the structural evidence $\mathcal E_t$ from the residual stream at the positions that predict a label, in the three implicit learners trained at the generative base rate and in their four post-trained branches at 20,000 steps, with the probes refitted on each. Steering \citep{bigelow2025belief,bigelow2026stories} adds $\alpha u$ to the residual stream at all these positions after the layer at which the probe reads best, along four families of unit directions $u$: the probes' decoding directions $w/\lVert w\rVert$, the component of each orthogonal to the other, the encoding directions $Cw/\lVert Cw\rVert$, along which the activations vary with the target ($C$ the covariance of the residual stream) \citep{haufe2014interpretation}, and random directions at the same norms. The dose $k$ is the shift of the probe's own read-out in nats, $\alpha=k/(w\cdot u)$, from $\pm0.5$ to $\pm32$ in factors of two ($\pm1$, $\pm8$ and $\pm32$ on the branches), and the criterion was fixed before the run: a direction moves the lift if at $k=\pm1$ the lift time shifts by at least one step with a 95\% interval that excludes zero.

\begin{figure}[h]
  \centering
  \includegraphics[width=\textwidth]{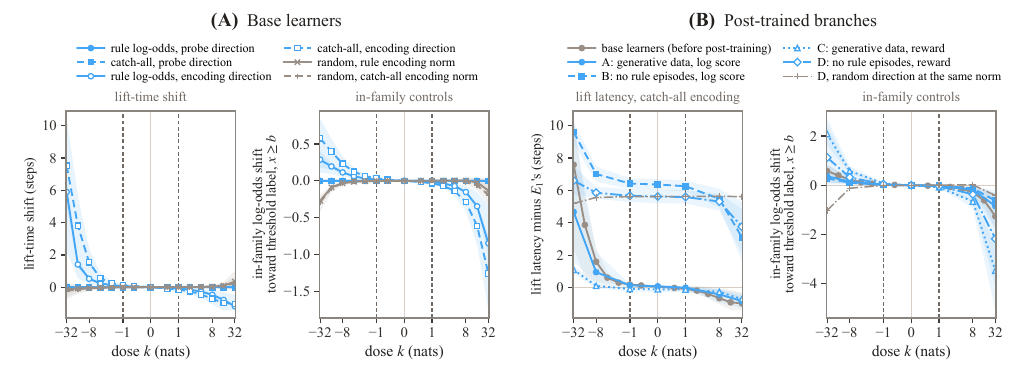}
  \caption{\textbf{Steering along the probe and encoding directions.} Dashed lines mark $k=\pm1$, where the pre-specified criterion is read; bands are 95\% intervals over models, then episodes. (A) The base learners: the lift-time shift, and the in-family controls, where negative values are false lifts. (B) The post-trained branches: the lift latency, and the in-family controls.}
  \label{fig:steering}
\end{figure}

Four facts follow (Figure~\ref{fig:steering}). (i) The probe directions move nothing: they are nearly orthogonal to the directions along which the residual stream varies with the target (cosine 0.03 to 0.06). (ii) The encoding directions move the lift monotonically in $k$, and random directions at the same norm do not. The rule and catch-all encoding directions nearly coincide (cosine 0.86 to 0.97), and their pushes move the in-family controls the same way, so no separate catch-all channel is found. (iii) The pre-specified criterion, a shift of one step at $k=\pm1$, is met by no direction on any model. (iv) On the branches trained without \cls{rule} episodes, B and D (the latter with the reward objective), a push along the catch-all encoding direction brings the lift latency, the excess over $E_1$ on all \cls{rule} episodes with an informative observation (7.0 and 6.2 steps on those in which $E_1$ lifts), back from $+6.4$ and $+5.6$ steps to $+3.1$ and $+3.8$ at $k=+32$, while a random direction at the same norm leaves D at $+5.6$, and the push produces false lifts on the in-family controls. The suppressed response can be partly re-evoked, and not selectively.

\FloatBarrier
\section{Prompts and read-outs}\label{app:prompts}
This appendix collects the input formats and read-outs of the language models (Section~\ref{sec:learners}), the e-process trigger (Section~\ref{sec:sre}), and notes on the decisions of the language models (Section~\ref{sec:llm-decision}).

In the implicit protocol every stage reads the same raw text. An episode is written one observation per line and the model assigns probabilities to the two labels at each of the 16 inputs, which gives the same table of predictions as for the small Transformers. The context holds the test episode alone, or the test episode after eight demonstration episodes drawn from threshold rules, or after eight drawn from the generative mixture. A base model reads the prompt as raw text and a chat-tuned stage reads the same prompt inside the template of its lineage, and stages are compared within one format.

\textbf{Implicit protocol.} An episode is written as the header line \verb|Each line gives a position and its label.| followed by one line per observation. The three renderings differ in how a line is written; the first five lines of the episode of Appendix~\ref{app:sre} read
\begin{promptbox}[width=0.62\linewidth,title={implicit protocol: the first five lines in R1, R2 and R3}]
\begin{minipage}[t]{0.22\linewidth}{\color{promptink}R1}\par\vspace{2pt}
\begin{Verbatim}[fontsize=\footnotesize]
1 -> V
1 -> V
14 -> V
7 -> X
2 -> V
\end{Verbatim}
\end{minipage}\hfill
\begin{minipage}[t]{0.22\linewidth}{\color{promptink}R2}\par\vspace{2pt}
\begin{Verbatim}[fontsize=\footnotesize]
a -> @
a -> @
n -> @
g -> ~
b -> @
\end{Verbatim}
\end{minipage}\hfill
\begin{minipage}[t]{0.22\linewidth}{\color{promptink}R3}\par\vspace{2pt}
\begin{Verbatim}[fontsize=\footnotesize]
1: V
1: V
14: V
7: X
2: V
\end{Verbatim}
\end{minipage}
\end{promptbox}
Rendering R1 writes the positions as 1 to 16 and joins position and label by an arrow, R2 writes the positions as the letters a to p, and R3 joins position and label by a colon. The two label symbols of an episode are drawn from Q, Z, K, X, V and J in R1 and R3. In R2 they are \verb|@| and \texttt{\fontencoding{T1}\selectfont\char126}, the only ASCII punctuation marks that stay a single token at the end of a line under the OLMo tokenizer, and the draw decides which of them stands for label 1. Every symbol is a single token with its leading space, and no word that names a class occurs in any text. The context condition C0 is the header and the evaluated episode. In CF0 the header is followed by eight demonstration episodes from the current family, each introduced by a line \texttt{Sequence}~$k$ after a blank line, and then by \verb|Sequence 9| and the evaluated episode; CMIX draws the demonstrations from the generative mixture. The demonstrations carry the evaluated episode's two symbols with the same assignment and are the same for every model and rendering, and the episodes of a group, one of each class, share them and the draw of symbols. Two variants of CF0 give each demonstration its own pair of symbols, or shift its positions to 17 to 32 (R1 and R3). The prediction of a label is read at the separator before it as $q=P(\xi_1)/(P(\xi_0)+P(\xi_1))$, the probability of the symbol $\xi_1$ of label 1 renormalized over the two symbols. The table of predictions at all 16 inputs after $t$ observations is read by appending, after line $t$, a line with the input and its separator, at $t\in\{8,16,24,32,40\}$, and at every $t$ on the dense subset of Appendix~\ref{app:training}.

\textbf{Explicit protocol.} A prompt describes the game, the default predictor, the four actions, the announcement and the scoring rule, followed by the log of observations; the prices, the number of scored rounds, the announced position and the cue \verb|Decision:| come last. The observations are rendered as in the implicit protocol, and under R2 the description of the default predictor reads \emph{before} and \emph{at or after} in place of \emph{below} and \emph{at or above}. In action order A the actions and their prices are listed as continue, patch, expand, observe, and in order B in reverse. The prompts name stay and defer as continue and observe, and the read-out maps these two surface forms back to the actions. The history-level prompt of one decision point of the same-history design in rendering R1 and order A reads as follows, with its 16 observation lines set here in four columns:
\begin{promptbox}[title={explicit protocol: history-level prompt of one decision point (R1, order A)}]
\setlength{\topsep}{1pt}\setlength{\partopsep}{0pt}\setlength{\multicolsep}{1pt}%
\begin{Verbatim}[fontsize=\scriptsize,breaklines=true]
A hidden machine gives each of the positions 1 to 16 one of two labels, J or K. In every round one position is queried and its label is observed. With probability 0.08 an observed label is misreported (the other label is shown), independently in every round. Each round queries a position at random: with probability 0.35 one that was queried before, otherwise one that was not queried before.

In every scored round a predictor gives the probability of each label at the queried position. The default predictor assumes a single boundary: positions below the boundary have label J and positions at or above it have label K. It infers the boundary from the observed labels, allows for misreported labels, and allows the boundary to move once during the game.

Before the scored rounds you take exactly one action:
- continue: keep the default predictor.
- patch: the position of the most recent observation that disagrees with the default predictor (if none disagrees, the most recently queried position) gets a separate record: from then on its label is predicted from the labels observed at that position alone, and the default predictor no longer uses that position.
- expand: replace the default predictor by the full predictor, which considers every kind of machine this game can produce, including patterns that a single boundary cannot express, and re-reads the whole log.
- observe: wait until the label of the announced position is shown, then take one of continue, patch or expand (and pay its cost as well).

One position is announced in advance. Its label is shown to every predictor before the scored rounds and is not scored.

Score: the sum over the scored rounds of the natural logarithm of the probability that the predictor in use gave to the observed label, minus the cost of the actions taken. For example, a probability of 0.9 on the observed label adds -0.11 to the score and a probability of 0.5 adds -0.69. Choose the action with the highest expected score and answer with its name.

Observations so far (position -> label):
\end{Verbatim}
\begin{multicols}{4}
\begin{Verbatim}[fontsize=\scriptsize]
3 -> J
7 -> K
1 -> J
9 -> K
5 -> K
15 -> J
11 -> K
2 -> J
13 -> K
6 -> K
15 -> J
10 -> K
4 -> K
12 -> K
8 -> K
14 -> K
\end{Verbatim}
\end{multicols}
\begin{Verbatim}[fontsize=\scriptsize,breaklines=true]

Costs: continue 0, patch 1, expand 4, observe 0.1.
Scored rounds: 30.
Announced position: 16.
Decision:
\end{Verbatim}
\end{promptbox}
The failure-level prompt of the same decision point adds a summary of the default predictor's fit between the log and the costs, set off by blank lines; an observation disagrees with the predictor when its current fit gives the observed label a probability below one half:
\begin{promptbox}[title={failure-level prompt: the summary added between the log and the costs}]
\begin{Verbatim}[fontsize=\scriptsize,breaklines=true]
Summary of the default predictor's fit: 2 of the 16 observations disagree with it.
Position 15: 2 of 2 observations disagree (repeated).
The most recent disagreement is at position 15.
\end{Verbatim}
\end{promptbox}
The gain-level prompt adds, after the announced position, the reference's expected gain per scored round from expanding now, before costs:
\begin{promptbox}[title={gain-level prompt: the line added after the announced position}]
\begin{Verbatim}[fontsize=\scriptsize,breaklines=true]
Expected gain in score per scored round if you expand now instead of continuing, not counting costs: 0.21.
\end{Verbatim}
\end{promptbox}

\textbf{Formats.} In raw text a model reads the prompt as it stands. In the chat format the prompt without its cue line is the user message of the model's chat template, and the assistant turn is pre-filled with the cue. A model without a chat template is wrapped in the template of its lineage's instruct model, and both Qwen 3 models in the template of Qwen 3 8B with thinking disabled, which opens the assistant turn with an empty thinking block. Every prompt starts with the start-of-text token where the tokenizer prepends one by default, as in the Llama 3.1 lineage. The templates of OLMo 3 Instruct, OLMo 3 Think and T{\"u}lu 3 wrap the prompt above as follows.

\begin{promptbox}[title={chat format: the OLMo 3 Instruct template}]
\begin{Verbatim}[fontsize=\scriptsize,breaklines=true,commandchars=\\\{\}]
<|im_start|>system
You are a helpful function-calling AI assistant. You do not currently have access to any functions. <functions></functions><|im_end|>
<|im_start|>user
\textrm{\textit{(the prompt above without its cue line)}}<|im_end|>
<|im_start|>assistant
Decision:
\end{Verbatim}
\end{promptbox}
\begin{promptbox}[title={chat format: the OLMo 3 Think template}]
\begin{Verbatim}[fontsize=\scriptsize,breaklines=true,commandchars=\\\{\}]
<|im_start|>system
You are OLMo, a helpful function-calling AI assistant built by Ai2. Your date cutoff is November 2024, and your model weights are available at https://huggingface.co/allenai. You do not currently have access to any functions. <functions></functions><|im_end|>
<|im_start|>user
\textrm{\textit{(the prompt above without its cue line)}}<|im_end|>
<|im_start|>assistant
<think></think>
Decision:
\end{Verbatim}
\end{promptbox}
\begin{promptbox}[title={chat format: the T{\"u}lu 3 template}]
\begin{Verbatim}[fontsize=\scriptsize,breaklines=true,commandchars=\\\{\}]
<|user|>
\textrm{\textit{(the prompt above without its cue line)}}
<|assistant|>
Decision:
\end{Verbatim}
\end{promptbox}
Llama 3.1 Instruct is wrapped in the same way, in the chat template that its own tokenizer provides.

\textbf{Action read-out.} The probability of an action is read at the position after the cue, from the action's name with a leading space and, in the summed read-out, from its four single-token surface forms, with and without the leading space, capitalized or not. In the chat format the answer may open with two asterisks after a space. The path read-out also counts the answers that open this way: it adds for each action the probability of this opening $o$ times the probability of the action's forms after it,
\[
P(a)=\sum_{v\in V_a}P(v)+P(o)\sum_{v\in V'_a}P(v\mid o),
\]
where $V_a$ and $V'_a$ are the forms of $a$ that are canonical single tokens at the cue and after $o$. A read-out is valid for a model in a format if the actions hold at least half of the probability mass on average and the most probable first token is an action or the opening at 90\% of the decision points or more. In the chat format with the path read-out every model is valid in every combination of rendering, action order and input level, except OLMo 3 Think, whose five combinations in action order B are excluded, and OLMo 3 Think DPO, which is valid in no format and was not run. In raw text the read-out is valid for OLMo 3 Base, OLMo 3 stage 1, Llama 3.1 Base and T{\"u}lu 3.1.

In the one-shot protocol deferring first is almost always affordable, since an immediate action gains at most $c_o=0.1$ over deferring, so the language models' decisions are read through regression coefficients and class-wise rates rather than regret.

The e-process trigger expands when $\mathcal E_t\ge\log\tfrac1\alpha$. It controls the probability of expanding under a correct $\mathcal F_0$ at level $\alpha$ and is blind to price and horizon.

\emph{Each learner is read on its own decision points.} In panel A of Figure~\ref{fig:model-map} the reference, the bounded reference and the triggers are read on the common states (Appendix~\ref{app:sre}), and the small Transformers on the common states within the trained range of prices. The language models are read on their own decision points, the 5,400 natural points in rendering R1 and action order A (cell R1-A): one prefix from each of 300 test episodes, under the 18 combinations of six prices $c_E$ and three horizons, with the episode's next query announced. The three models with reasoning are read on the 1,098 same-history points of Appendix~\ref{app:reasoning-e11}, on which the reference lies at (0.59, 0.57). In panel B the small Transformer is read on the common states and the language models on the natural points, at rule log-odds between 0 and 2. At the history level the language models do not respond to the benefit, one weakly excepted, and their direction is undefined.

\emph{The history-level decision is constant, with small exceptions.} A model decides constantly at the history level when, on the 5,400 natural points of cell R1-A, the lower bound of the 95\% cluster-bootstrap interval of its benefit coefficient is at or below zero and its probability of expanding is flat across the source classes, within 0.05 between \cls{rule} and threshold episodes and within 0.02 between a \cls{rule} history and its boundary-matched threshold history, the history of the threshold episode whose boundary is the lower end of the rule (Appendix~\ref{app:training}). The robustness subset reads 900 of these points again in the other five combinations of rendering and action order, which name its cells. Among its 92 valid cells, one per model, format and combination, 78 are null in each of 20 redrawn bootstraps, 6 show a response in each and 8 are mixed. OLMo 3 Instruct shows a small, poorly determined response in R2-B and R3-A (0.35 \ci{0.10}{1.9} and 0.80 \ci{0.10}{35}) and none that holds across draws in R1-B. From preference tuning on, the T{\"u}lu stages show a small response in R2-A and R3-A, under the first action order (0.06 to 0.12). In R3-A most post-trained models share it, and it is absent in every base model, in OLMo 3 Instruct SFT and DPO, in both Think stages and in T{\"u}lu 3 SFT. The rise of the benefit coefficient in cell R1-A from OLMo 3 Instruct DPO to OLMo 3 Instruct ($+0.06$ \ci{0.00}{0.21}) is marginal. At the failure level the summary of the failures adds no response for OLMo 3 Instruct SFT, OLMo 3 Instruct DPO and OLMo 3 Instruct and small ones for OLMo 3 Base (0.014 \ci{0.002}{0.027}), OLMo 3 Think SFT (0.034 \ci{0.019}{0.051}) and OLMo 3 Think (0.053 \ci{0.027}{0.081}), and T{\"u}lu 3 DPO, T{\"u}lu 3 and T{\"u}lu 3.1 expand less where the benefit is larger ($-0.10$ to $-0.17$).

\emph{Predictions that have moved do not reach the decision.} In ten of the thirteen models of Section~\ref{sec:llm-decision} the decision almost never follows predictions that have moved toward the rule.
The other three expand often whether or not their predictions have moved, and OLMo 3 Think less often where they have (Appendix~\ref{app:controls}).

\emph{Without reasoning the OLMo 3 models do not weigh a stated gain.} In the lowest fifth of the stated gains the four stages of OLMo 3 Instruct expand with probability 0.34 to 0.73, and the reference with 0.02. With thinking off the two OLMo 3 Think models put expand first on nearly every gain-level point, and their probability of expanding is nearly flat in the stated gain, the price and the horizon: for OLMo 3 Think it falls from 0.79 to 0.73 as $c_E$ rises from 0.5 to 16, where the reference's falls from 0.77 to 0.07, and the slopes on the gain, the price and the horizon are 0.05 to 0.13 per $e$-fold on the natural points. The ratio of two such slopes is not identified, and its fitted value follows the price grid of the design (Appendix~\ref{app:reasoning-olmo}). With room to reason the slopes grow tenfold or more and the ratios settle near one.

\textbf{The Qwen 3 pair.} Both models are read in the chat template with thinking turned off and with the path read-out, and the values below are those of the 5,400 natural decision points. At the history level neither model responds to the benefit of expanding (coefficients $-0.01$ for the base model and 0.03 for the final model, both intervals including zero), and the probability of expanding differs between the classes by at most 0.01. The final model patches almost always, with probability 0.92. Both models know without doing: at 0.60 and 0.48 of the points where the reference expands their predictions, read on the decision-context prefix, have moved toward the rule, and at none of these points is expanding their most probable action. At the gain level the final model's response to the stated gain is 0.76, with price sensitivity 0.37 and horizon sensitivity 0.14 ($-0.14$ on the 1,098 same-history points of Appendix~\ref{app:reasoning-e11}), and these ratios hold across the six combinations of rendering and action order (price sensitivity 0.22 to 0.37, horizon sensitivity $-0.11$ to 0.15). It weighs the price in part and the horizon barely, beyond the final Instruct models, which respond to the gain alone. The base model's response to the gain is 0.12, and its ratios change sign across the six combinations. At rule log-odds between 0 and 2, the final model's probability of expanding falls only from 0.65 to 0.58 across the range of prices, where the reference's falls from 1 to 0, and in the coordinates of Figure~\ref{fig:model-map}A the final model lies on the price side of the cluster of the Instruct and T{\"u}lu models, with a share of 0.35 on the price. At the failure level the final model shows a small response to the benefit (0.08) whose direction follows the price and the horizon rather than the benefit.

\subsection{Controls on task specification and read-out context}\label{app:controls}

\textbf{The generative process stated in the prompt.} The history-level prompt of rendering R1 describes the full predictor only as one that considers every kind of machine the game can produce, while the reference knows the four kinds and their base rates. Rendering R1S adds a paragraph between the description of the default predictor and the list of actions. It states the four kinds of machine with their probabilities, how their parameters are drawn and which kinds each of the two predictors assumes, in the words of the prompt and without the name of any class. For the decision point above it reads
\begin{promptbox}[title={R1S: the paragraph added between the default predictor and the actions}]
\begin{Verbatim}[fontsize=\scriptsize,breaklines=true]
The hidden machine is one of four kinds, drawn before the game with these probabilities. With probability 0.40 it is a single boundary and nothing else, so that every disagreement with it comes from a misreported label. With probability 0.20 it is a single boundary with one or two positions whose true label is the opposite of what the boundary says. With probability 0.15 it is a single boundary that moves once during the game to a new place. With probability 0.25 it gives label K to a block of positions and label J to all other positions: the positions from a lower end to an upper end have label K, so that the positions above the upper end have label J although they lie above the lower end. In the first three kinds the boundary (in the third kind, its place before the move) is drawn uniformly from the positions 1 to 16. In the second kind the number of positions with the opposite label is one with probability 0.676 and two with probability 0.324, and these positions are drawn uniformly from the 16 positions. In the third kind the new place is drawn uniformly from the other 15 positions, and the round from which the boundary is at the new place is drawn uniformly from the rounds 11 to 31, counting the observations so far as rounds 1, 2, 3 and so on, then the announced position and then the scored rounds. In the fourth kind the two ends are drawn uniformly from the 120 pairs in which the lower end is at or below the upper end and the upper end is at most 15. The kind does not affect which positions are queried, and in every kind an observed label is misreported with probability 0.08, as described above. The default predictor performs Bayesian prediction under the first and third kinds, with their probabilities renormalized. The full predictor performs Bayesian prediction under all four kinds with the probabilities given above and re-reads the whole log.
\end{Verbatim}
\end{promptbox}
Rendering R1L adds in the same place a paragraph of about the same number of tokens that restates what R1 already says about the queries, the log, the announced position and the score:
\begin{promptbox}[title={R1L: the paragraph of the length control, in the same place}]
\begin{Verbatim}[fontsize=\scriptsize,breaklines=true]
The game proceeds in rounds. In every round one position is queried and its label is observed. The position of a round is chosen at random: with probability 0.35 it is one of the positions that were queried in earlier rounds, chosen at random among them, and otherwise it is one of the positions that were not queried in earlier rounds, chosen at random among them. In every round, with probability 0.08 the observed label is misreported, so that the other label is shown, and this happens independently in every round. The observations so far are listed below, after the actions and the score, one line per round, in the order in which the rounds took place: each line gives the position that was queried in that round, then an arrow, then the label that was observed in that round. After the observations so far, one position is announced in advance; it is given below, after the costs and the number of scored rounds. Its label is shown to every predictor before the scored rounds, whichever action is taken, and it is not scored. Then the scored rounds follow, and their number is given below, after the costs. In every scored round one position is queried, the predictor in use gives a probability to each of the two labels at that position, the two probabilities add up to one, and then the observed label is shown. The scored round adds to the score the natural logarithm of the probability that the predictor in use gave to the observed label. This number is never above zero: it is close to zero when the probability given to the observed label is close to one, and it is far below zero when that probability is small. For example, a probability of 0.9 on the observed label adds -0.11 to the score, and a probability of 0.5 adds -0.69. The score of the game is the sum of these numbers over all scored rounds, minus the cost of the actions taken; the costs are listed below with the number of scored rounds and the announced position.
\end{Verbatim}
\end{promptbox}

\begin{table}[h]
\caption{The generative process stated in the prompt (R1S) against R1 and the length control R1L, at the history level in order A: the probability of expanding on the \cls{noise} histories of the matched groups, from which the other classes differ by at most 0.05, its gap between \cls{rule} histories and their boundary-matched threshold histories, and the benefit coefficient $\beta_b$ on the same-history points, with 95\% intervals from the paired cluster bootstrap. No cell meets the criterion, in either action order.}
\label{tab:controls-spec}
\centering\tablestyle
\setcounter{tabcols}{7}
\begin{tabularx}{\textwidth}{>{\raggedright\arraybackslash}X
  S[table-format=1.3,table-column-width=33.5pt] S[table-format=1.3,table-column-width=33.5pt] S[table-format=1.3,table-column-width=33.5pt]
  S[table-format=-1.3,table-space-text-post=\ci{-0.070}{-0.016}]
  S[table-format=-1.3,table-space-text-post=\ci{-0.060}{-0.004}]
  S[table-format=-1.3,table-space-text-post=\ci{-0.128}{-0.028}]}
\toprule
 & \multicolumn{3}{c}{\hd{probability of expanding, \cls{noise}}} & \multicolumn{2}{c}{\hd{\cls{rule} $-$ threshold}} & {\hd{$\beta_b$}} \\
\cmidrule(lr){2-4}\cmidrule(lr){5-6}\cmidrule(lr){7-7}
\hd{model} & {\hd{R1}} & {\hd{R1L}} & {\hd{R1S}} & {\hd{R1}} & {\hd{R1S}} & {\hd{R1S}} \\
\midrule
\grouprow{OLMo 3}
OLMo 3 stage 1 & 0.352 & 0.237 & 0.290 & 0.001 {\ci{-0.001}{0.004}} & 0.000 {\ci{-0.001}{0.001}} & -0.033 {\ci{-0.094}{0.018}} \\
\rowcolor{rowgray}OLMo 3 Base & 0.122 & 0.136 & 0.159 & 0.000 {\ci{-0.001}{0.001}} & -0.001 {\ci{-0.002}{0.000}} & -0.039 {\ci{-0.088}{0.011}} \\
OLMo 3 Instruct SFT & 0.054 & 0.247 & 0.176 & -0.002 {\ci{-0.007}{0.004}} & -0.009 {\ci{-0.023}{0.003}} & -0.030 {\ci{-0.552}{0.753}} \\
\rowcolor{rowgray}OLMo 3 Instruct DPO & 0.086 & 0.435 & 0.365 & -0.002 {\ci{-0.017}{0.010}} & -0.031 {\ci{-0.060}{-0.004}} & 0.127 {\ci{-0.441}{0.708}} \\
OLMo 3 Instruct & 0.050 & 0.395 & 0.446 & 0.002 {\ci{-0.011}{0.015}} & -0.033 {\ci{-0.084}{0.014}} & 0.191 {\ci{-0.402}{1.293}} \\
\rowcolor{rowgray}OLMo 3 Think SFT & 0.381 & 0.279 & 0.468 & -0.005 {\ci{-0.008}{-0.002}} & -0.004 {\ci{-0.007}{-0.001}} & -0.077 {\ci{-0.128}{-0.028}} \\
OLMo 3 Think & 0.531 & 0.416 & 0.715 & -0.007 {\ci{-0.013}{-0.001}} & -0.007 {\ci{-0.012}{-0.002}} & -0.066 {\ci{-0.174}{0.028}} \\
\groupsep
\grouprow{Llama 3.1 and T{\"u}lu 3}
Llama 3.1 Base & 0.283 & 0.201 & 0.283 & -0.003 {\ci{-0.006}{0.000}} & -0.002 {\ci{-0.005}{0.000}} & -0.020 {\ci{-0.062}{0.037}} \\
\rowcolor{rowgray}T{\"u}lu 3 SFT & 0.116 & 0.104 & 0.123 & -0.003 {\ci{-0.005}{0.000}} & -0.004 {\ci{-0.007}{-0.002}} & 0.014 {\ci{-0.086}{0.097}} \\
T{\"u}lu 3 DPO & 0.073 & 0.042 & 0.318 & -0.006 {\ci{-0.009}{-0.002}} & -0.027 {\ci{-0.038}{-0.016}} & 0.232 {\ci{-0.012}{0.758}} \\
\rowcolor{rowgray}T{\"u}lu 3 & 0.064 & 0.034 & 0.372 & -0.004 {\ci{-0.009}{-0.001}} & -0.025 {\ci{-0.037}{-0.014}} & 0.287 {\ci{0.029}{0.804}} \\
T{\"u}lu 3.1 & 0.130 & 0.071 & 0.344 & -0.010 {\ci{-0.018}{-0.004}} & -0.025 {\ci{-0.037}{-0.015}} & 0.134 {\ci{-0.021}{0.283}} \\
\rowcolor{rowgray}Llama 3.1 Instruct & 0.382 & 0.336 & 0.659 & -0.042 {\ci{-0.070}{-0.016}} & -0.014 {\ci{-0.027}{-0.001}} & 0.188 {\ci{-0.029}{0.470}} \\
\groupsep
\grouprow{Qwen 3}
Qwen 3 8B Base & 0.178 & 0.197 & 0.364 & -0.010 {\ci{-0.014}{-0.005}} & -0.012 {\ci{-0.018}{-0.006}} & 0.007 {\ci{-0.132}{0.207}} \\
\rowcolor{rowgray}Qwen 3 8B & 0.015 & 0.021 & 0.712 & -0.001 {\ci{-0.002}{0.000}} & -0.017 {\ci{-0.025}{-0.007}} & 0.163 {\ci{-0.031}{0.319}} \\
\bottomrule
\end{tabularx}
\end{table}

The two renderings are read on the points of the natural-finish arm of Appendix~\ref{app:reasoning-e11}, in which a model reasons until it ends by itself, within a cap of tokens: at the history level 600 matched points (40 matched groups and the boundary-matched threshold history of each \cls{rule} history, at $c_E=1$, 4 and 16) and 366 same-history points, in both action orders, and at the gain level 1,098 same-history points, in order A. The models are the thirteen of the explicit protocol and the Qwen 3 pair, in the chat format with the path read-out and thinking off. R1 is scored again on the same points, so that every comparison is paired. The read-out is valid in every cell except, as above, the order-B cells of OLMo 3 Think under R1 and R1L. The criterion is that of Appendix~\ref{app:reasoning-e11} with a positive coefficient: the decision uses the evidence when the benefit coefficient is positive with an interval that excludes zero and the probability of expanding on \cls{rule} histories exceeds that on their boundary-matched threshold histories by at least 0.10.

No model meets the criterion in any rendering or action order (Table~\ref{tab:controls-spec}). What the statement changes is the probability of expanding itself. It raises it in most models, in Qwen 3 8B from 0.015 to 0.712, and as much on boundary-matched threshold histories as on \cls{rule} histories: under R1S the gap between the two lies between $-0.064$ and 0.000 in the two orders. The length control moves the probability of expanding as well, up in some models and down in others, and leaves the gap below 0.05 in every model. At the gain level the statement leaves the weighing far from the reference's. Where the response to the stated gain is identified, in ten models, price sensitivity under R1S lies between $-0.12$ and 0.35 and horizon sensitivity between $-0.27$ and 0.42, against 1 for the reference.

\begin{table}[h]
\caption{The prediction read on the decision-context prefix, under R1 in order A: on the null-equivalent pairs the median matched-pair $\kappa$ and the difference in the probability of expanding between $D$ and $D'$, and at the 1,603 points of Figure~\ref{fig:model-map}C the share at which the prediction has moved toward the rule, on the decision-context prefix and in the implicit protocol, the share at which expand is the most probable action, at all points and where the prediction has or has not moved, and the probability mass on the two label symbols. The intervals are 95\%, over pairs for $\kappa$ and from the cluster bootstrap elsewhere.}
\label{tab:controls-knowing}
\centering\tablestyle
\setcounter{tabcols}{9}
\begin{tabularx}{\textwidth}{>{\raggedright\arraybackslash}X
  S[table-format=1.2,table-space-text-post=\ci{0.00}{0.00}]
  S[table-format=-1.3]
  S[table-format=1.2,table-space-text-post=\ci{0.00}{0.00}]
  S[table-format=1.2]
  S[table-format=1.2,table-space-text-post=\ci{0.00}{0.00}]
  S[table-format=1.2] S[table-format=1.2] S[table-format=1.2]}
\toprule
 & \multicolumn{2}{c}{\hd{null-equivalent pairs}} & \multicolumn{2}{c}{\hd{moved toward the rule}} & \multicolumn{3}{c}{\hd{expand most probable}} & \\
\cmidrule(lr){2-3}\cmidrule(lr){4-5}\cmidrule(lr){6-8}
\hd{model} & {\hd{median $\kappa$}} & {\hdtwo{expand,}{$D-D'$}} & {\hdtwo{decision}{prefix}} & {\hd{implicit}} & {\hd{all}} & {\hd{moved}} & {\hdtwo{not}{moved}} & {\hdtwo{label}{mass}} \\
\midrule
\grouprow{OLMo 3}
OLMo 3 stage 1 & 0.02 {\ci{0.01}{0.03}} & 0.000 & 0.61 {\ci{0.53}{0.69}} & 0.65 & 0.56 {\ci{0.49}{0.63}} & 0.55 & 0.58 & 0.88 \\
\rowcolor{rowgray}OLMo 3 Base & 0.08 {\ci{0.06}{0.12}} & -0.001 & 0.62 {\ci{0.54}{0.69}} & 0.69 & 0.00 {\ci{0.00}{0.00}} & 0.00 & 0.00 & 0.92 \\
OLMo 3 Instruct SFT & 0.25 {\ci{0.22}{0.29}} & -0.004 & 0.54 {\ci{0.46}{0.62}} & 0.65 & 0.00 {\ci{0.00}{0.00}} & 0.00 & 0.00 & 0.98 \\
\rowcolor{rowgray}OLMo 3 Instruct DPO & 0.40 {\ci{0.33}{0.47}} & -0.006 & 0.53 {\ci{0.45}{0.60}} & 0.64 & 0.00 {\ci{0.00}{0.00}} & 0.00 & 0.00 & 1.00 \\
OLMo 3 Instruct & 0.65 {\ci{0.56}{0.79}} & -0.003 & 0.54 {\ci{0.46}{0.62}} & 0.63 & 0.00 {\ci{0.00}{0.00}} & 0.00 & 0.00 & 1.00 \\
\rowcolor{rowgray}OLMo 3 Think SFT & 0.15 {\ci{0.12}{0.16}} & -0.006 & 0.57 {\ci{0.50}{0.65}} & 0.61 & 0.01 {\ci{0.00}{0.02}} & 0.00 & 0.01 & 0.97 \\
OLMo 3 Think & 0.22 {\ci{0.19}{0.26}} & -0.010 & 0.54 {\ci{0.46}{0.61}} & 0.60 & 0.54 {\ci{0.47}{0.61}} & 0.42 & 0.69 & 1.00 \\
\groupsep
\grouprow{Llama 3.1 and T{\"u}lu 3}
Llama 3.1 Base & 0.07 {\ci{0.06}{0.08}} & -0.004 & 0.65 {\ci{0.56}{0.72}} & 0.64 & 0.00 {\ci{0.00}{0.00}} & 0.00 & 0.00 & 0.80 \\
\rowcolor{rowgray}T{\"u}lu 3 SFT & 0.09 {\ci{0.07}{0.10}} & -0.003 & 0.65 {\ci{0.57}{0.72}} & 0.65 & 0.00 {\ci{0.00}{0.00}} & 0.00 & 0.00 & 1.00 \\
T{\"u}lu 3 DPO & 0.21 {\ci{0.17}{0.25}} & -0.003 & 0.59 {\ci{0.50}{0.66}} & 0.66 & 0.00 {\ci{0.00}{0.00}} & 0.00 & 0.00 & 1.00 \\
\rowcolor{rowgray}T{\"u}lu 3 & 0.22 {\ci{0.18}{0.24}} & -0.002 & 0.59 {\ci{0.51}{0.66}} & 0.67 & 0.00 {\ci{0.00}{0.00}} & 0.00 & 0.00 & 1.00 \\
T{\"u}lu 3.1 & 0.19 {\ci{0.16}{0.23}} & -0.005 & 0.54 {\ci{0.46}{0.62}} & 0.66 & 0.00 {\ci{0.00}{0.00}} & 0.00 & 0.00 & 0.99 \\
\rowcolor{rowgray}Llama 3.1 Instruct & 0.11 {\ci{0.08}{0.12}} & -0.026 & 0.60 {\ci{0.52}{0.67}} & 0.62 & 0.59 {\ci{0.52}{0.66}} & 0.61 & 0.56 & 0.89 \\
\groupsep
\grouprow{Qwen 3}
Qwen 3 8B Base & 0.08 {\ci{0.07}{0.10}} & -0.007 & 0.60 {\ci{0.52}{0.67}} & 0.66 & 0.00 {\ci{0.00}{0.00}} & 0.00 & 0.00 & 0.84 \\
\rowcolor{rowgray}Qwen 3 8B & 0.31 {\ci{0.28}{0.35}} & -0.001 & 0.48 {\ci{0.41}{0.56}} & 0.67 & 0.00 {\ci{0.00}{0.00}} & 0.00 & 0.00 & 1.00 \\
\bottomrule
\end{tabularx}
\end{table}

\textbf{The prediction read on the decision-context prefix.} The predictions behind the knowing without doing of Section~\ref{sec:llm-decision} and Figure~\ref{fig:model-map}C are read on the prefix of the decision prompt. The history-level prompt is kept byte for byte up to its cue, and the cue is replaced by the question \texttt{Before you act: which label do you expect to observe when position $x$ is queried? Answer with J or K.} With the assistant turn pre-filled with \verb|Label:| the probability of a label is read over the forms of its symbol with and without the leading space, at the cue and after the opening of the path read-out, and renormalized over the two labels as in the implicit protocol. The question is asked at every unqueried position at which $E_1$ and $E_0$ differ by more than 0.1 in log-odds, the positions of the implicit read-out, and the decision is read on the identical prefix. The prediction has moved toward the rule when its shift in log-odds from $E_0$, regressed through the origin on that of $E_1$ over these positions, has a slope above one half, and the decision is expand when expand is the most probable of the four actions. The two symbols hold 0.80 to 1.00 of the probability mass, and every cell passes the validity condition of the implicit protocol.

Two sets of points are read. The first is the 300 null-equivalent pairs of the implicit protocol, 60 at each level $m$. Each history is written as a history-level decision prompt at $c_E=4$, $c_p=0.5$ and $H=20$, with the last queried position announced, the question asks for the probe $x^\ast$, and the two members of a pair give a matched-pair $\kappa$ on the decision-context prefix, under R1 and R1S. The second is the 1,603 decision points of Figure~\ref{fig:model-map}C, the natural and matched points of cell R1-A at which the reference expands, read under R1.

On the pairs the matched-pair $\kappa$ read on the decision-context prefix is positive in every model, with medians from 0.02 to 0.65 under R1, and the probability of expanding on the same prefixes is no higher on $D$ than on $D'$ (Table~\ref{tab:controls-knowing}). Stating the generative process raises $\kappa$ in the three stages of OLMo 3 Instruct, changes it by at most 0.05 in the other models, and leaves the decision no higher on $D$ than on $D'$. At the points of Figure~\ref{fig:model-map}C the prediction has moved toward the rule at 0.48 to 0.65 of the points, against 0.60 to 0.69 in the implicit protocol on the same points. In twelve models expand is the most probable action at none of the points where the prediction has moved. In the other three, OLMo 3 stage 1, OLMo 3 Think and Llama 3.1 Instruct, it is the most probable action at about half of the points whether or not the prediction has moved, and in OLMo 3 Think less often where it has.

\FloatBarrier
\section{Training and evaluation details}\label{app:training}

\textbf{Small Transformers.} The small Transformers share one decoder-only architecture, with layer normalization before each sublayer: six blocks of width 256 with eight attention heads and GELU feed-forward layers of width 1,024. The implicit learner reads two tokens per step, the input and its label, has learned position embeddings and a two-way label head, and has 4.8 million parameters. It is trained on episodes of 41 steps, whose first 40 follow the law of SRE at $T=40$ and whose last step brings the prediction after the fortieth observation into training. % lead-in (new sentence); the next sentence moved from Section 4 (round 11)
The three implicit learners, MIX, RARE and F0ONLY, differ only in the base rate of \cls{rule} episodes, 0.25, 0.02 and 0 (Table~\ref{tab:training}). With a base rate of 0 the learner has never seen an interval rule, and it shows what adaptation inside the current family can and cannot imitate. For Appendix~\ref{app:architectures} the attention of the learner trained at the generative base rate is replaced by a single-layer GRU, in five blocks of width 260 (4.77 million parameters), or by a selective state-space layer in the Mamba-2 form, in five blocks of width 256 with inner width 512 and state size 64 (4.81 million), against 4.78 million for the Transformer; both are trained like it, with three seeds each. The explicit learner uses the same blocks with two tokens per step: a decision token that carries the input, the announced query, the patch slots in use and Fourier features of $\log c_E$, $\log c_p$, $t$, $H_t$ and $\log H_t$, and an outcome token that carries the observed label and the action taken. A three-way head on the decision token chooses among stay, patch and expand. Time enters only as a feature, so episodes longer than those trained on are defined. All runs use AdamW \citep{loshchilov2019decoupled} with $\beta=(0.9,0.95)$, weight decay 0.1 on the weight matrices and gradient clipping at norm 1 (Table~\ref{tab:training}).

\begin{table}[h]
\caption{Training runs of the small Transformers and of the two fixed-state learners of Appendix~\ref{app:architectures}. MIX, RARE and F0ONLY are the implicit learners trained with a \cls{rule} base rate of 0.25 (that of the generative process), 0.02 and 0, and the names also stand for these episode streams; BC, POLICY-ONLY and REWARD are the regimes of Table~\ref{tab:regret}. Learning rates rise linearly over the warm-up and then follow a cosine to 10\% of the peak, except in post-training, where they stay constant. Every run draws fresh episodes at each step except the three BC regimes, which read the label set.}
\label{tab:training}
\centering\tablestyle
\setcounter{tabcols}{7}
\begin{tabularx}{\textwidth}{l >{\raggedright\arraybackslash}X S[table-format=6.0] S[table-format=3.0] c S[table-format=4.0] l}
\toprule
 & & & & \multicolumn{2}{c}{\hd{learning rate}} & \\
\cmidrule(lr){5-6}
\hd{learner} & \hd{data} & {\hd{steps}} & {\hd{batch}} & \hd{peak} & {\hd{warm-up steps}} & \hd{runs} \\
\midrule
\grouprow{Implicit learners}
MIX & \cls{rule} base rate 0.25 & 60000 & 512 & $6\times10^{-4}$ & 2000 & 3 seeds \\
\rowcolor{rowgray}RARE & \cls{rule} base rate 0.02 & 60000 & 512 & $6\times10^{-4}$ & 2000 & 3 seeds$^{a}$ \\
F0ONLY & \cls{rule} base rate 0 & 60000 & 512 & $6\times10^{-4}$ & 2000 & 3 seeds \\
\rowcolor{rowgray}MIX, GRU & \cls{rule} base rate 0.25 & 60000 & 512 & $6\times10^{-4}$ & 2000 & 3 seeds \\
MIX, selective SSM & \cls{rule} base rate 0.25 & 60000 & 512 & $6\times10^{-4}$ & 2000 & 3 seeds \\
\groupsep
\grouprow{Explicit learners}
BC on the reference's path & label set & 20000 & 384 & $6\times10^{-4}$ & 1000 & 3 seeds \\
\rowcolor{rowgray}BC with path labels & label set, paths beyond the expansion & 20000 & 384 & $6\times10^{-4}$ & 1000 & 3 seeds \\
POLICY-ONLY & label set, posterior as input & 20000 & 384 & $6\times10^{-4}$ & 1000 & 3 seeds \\
\rowcolor{rowgray}REWARD & \cls{rule} base rate 0.25 & 20000 & 192 & $3\times10^{-4}$ & 1000 & 3 seeds \\
REWARD from BC & \cls{rule} base rate 0.25 & 20000 & 192 & $10^{-4}$ & 1000 & 3 seeds \\
\groupsep
\grouprow{Post-training}
Branches A--D & \cls{rule} base rate 0.25 or 0 & 20000 & 512 & $10^{-4}$ & 200 & 3 base learners$^{b}$ \\
\rowcolor{rowgray}Recovery from B and D & \cls{rule} base rate 0.25 & 2000 & 512 & $10^{-4}$ & {none} & 3 base learners \\
\bottomrule
\multicolumn{7}{l}{\scriptsize $^{a}$\,And one run of 120,000 steps.\qquad $^{b}$\,On the first base learner, 3 data seeds.}
\end{tabularx}
\end{table}

The imitation learners (BC) are trained on 20,000 episodes, 4,000 at each $T\in\{8,16,24,32,40\}$, labelled with the reference's action at the training price cells; \cls{rule} episodes with the held-out widths are removed, and the last 200 episodes of each $T$ select the checkpoint by validation cross-entropy. BC on the reference's path labels the states of the reference's own path up to its expansion. BC with path labels adds the states beyond the expansion, along the path of the reference's best choice between staying and patching and along the never-revise path. POLICY-ONLY is BC on the reference's path with the exact class posterior and rule log-odds added to the input. REWARD regresses, for staying and for patching, the value relative to expanding now, which is known exactly on the episode because the observations do not depend on the actions. The target of the action taken is its one-step reward plus the realized utility of the learner's own greedy continuation, fitted with a Huber loss ($\delta=2$) on fresh episodes, with $T$ drawn in proportion to its value and one stay-or-patch decision per episode flipped for exploration; the checkpoint is selected by regret on the validation episodes. REWARD from BC starts from BC on the reference's path and trains the head alone for its first 2,000 steps.

\textbf{Post-training.} Each of the three MIX learners is post-trained from its final checkpoint, with a fresh optimizer, in four branches that cross the data, MIX or F0ONLY episodes, with the objective, the log loss $-\log q_t(y_t)$ or the expected 0/1 reward $-q_t(y_t)$ of answering with a label sampled from the prediction, whose gradient is the exact policy gradient. The branches are read after 500, 1,000, 2,000, 5,000, 10,000 and 20,000 steps, and branches with the same data share the stream of episodes. On the first learner every branch is repeated with two further data seeds, and C and D also with a REINFORCE estimate \citep{williams1992simple} of the reward gradient with a leave-one-out baseline. The recovery runs continue B and D from 20,000 steps with branch A's data and objective and the optimizer state of that checkpoint, and are read every 200 steps.

\textbf{Language models.} Table~\ref{tab:llms} lists the models, with the six 32B models of Appendix~\ref{app:reasoning}. The fourteen models of the OLMo 3, OLMo 3 Think and Llama 3.1 / T{\"u}lu 3 lineages are read in the implicit protocol, and thirteen of them in the explicit protocol, all but OLMo 3 Think DPO (Appendix~\ref{app:prompts}). The Qwen 3 pair is read in the implicit protocol in raw text, which is the non-thinking use of the final model, in 36 cells of rendering, context and evaluation set per model. Every model is scored in bfloat16 with its own tokenizer. In the explicit protocol both models of the pair are scored in the chat template with thinking turned off and with the path read-out, and all 14 cells are valid for each: the actions hold 0.61 to 0.78 of the probability mass for the base model and 0.95 to 0.995 for the final model, and the most probable first token leads to an action at 0.993 of the points or more. The base model's raw-text read-out fails the validity condition on the history-level subset (mass 0.27, first token 0.35) and was not run, so the pair has no raw-text row and its comparison of base and final model stays within one format.

\newcommand{\hfrepo}[2]{\href{https://huggingface.co/#1/tree/#2}{\texttt{#1}}}
\begin{table}[h]
\caption{Sources of the language models: OLMo 3 and OLMo 3.1 \citep{olmo2025olmo3}, Llama 3.1 \citep{grattafiori2024llama3}, T{\"u}lu 3 and T{\"u}lu 3.1 \citep{lambert2024tulu3} and Qwen 3 \citep{yang2025qwen3}; each repository name links to the revision that was read.}
\label{tab:llms}
\centering\tablestyle\setlength{\tabcolsep}{6pt}
\setcounter{tabcols}{3}
\begin{tabular}{lll}
\toprule
\hd{model} & \hd{stage} & \hd{repository} \\
\midrule
\grouprow{OLMo 3, 7 billion parameters}
OLMo 3 stage 1 & pre-training, stage 1 & \hfrepo{allenai/Olmo-3-1025-7B}{373bad25002f} \\
\rowcolor{rowgray}OLMo 3 Base & base & \hfrepo{allenai/Olmo-3-1025-7B}{a81bae42db39} \\
OLMo 3 Instruct SFT & SFT & \hfrepo{allenai/Olmo-3-7B-Instruct-SFT}{e1452fc572d5} \\
\rowcolor{rowgray}OLMo 3 Instruct DPO & DPO & \hfrepo{allenai/Olmo-3-7B-Instruct-DPO}{b33130b7de49} \\
OLMo 3 Instruct & RLVR & \hfrepo{allenai/Olmo-3-7B-Instruct}{6e5971d9eba4} \\
\rowcolor{rowgray}OLMo 3 Think SFT & SFT & \hfrepo{allenai/Olmo-3-7B-Think-SFT}{6ff857587e04} \\
OLMo 3 Think DPO & DPO & \hfrepo{allenai/Olmo-3-7B-Think-DPO}{7b18bf927b43} \\
\rowcolor{rowgray}OLMo 3 Think & RLVR & \hfrepo{allenai/Olmo-3-7B-Think}{d97e442d7cc6} \\
\groupsep
\grouprow{Llama 3.1 and T{\"u}lu 3, 8 billion parameters}
Llama 3.1 Base & base & \hfrepo{meta-llama/Llama-3.1-8B}{d04e592bb4f6} \\
\rowcolor{rowgray}T{\"u}lu 3 SFT & SFT & \hfrepo{allenai/Llama-3.1-Tulu-3-8B-SFT}{f2a0b46b0cfd} \\
T{\"u}lu 3 DPO & DPO & \hfrepo{allenai/Llama-3.1-Tulu-3-8B-DPO}{a7beb67e33ff} \\
\rowcolor{rowgray}T{\"u}lu 3 & RLVR (PPO) & \hfrepo{allenai/Llama-3.1-Tulu-3-8B}{666943798adb} \\
T{\"u}lu 3.1 & RLVR (GRPO) & \hfrepo{allenai/Llama-3.1-Tulu-3.1-8B}{46239c2d07db} \\
\rowcolor{rowgray}Llama 3.1 Instruct & Meta's post-training & \hfrepo{meta-llama/Llama-3.1-8B-Instruct}{0e9e39f249a1} \\
\groupsep
\grouprow{Qwen 3, 8 billion parameters}
Qwen 3 8B Base & base & \hfrepo{Qwen/Qwen3-8B-Base}{49e3418fbbbc} \\
\rowcolor{rowgray}Qwen 3 8B & final post-trained & \hfrepo{Qwen/Qwen3-8B}{b968826d9c46} \\
\groupsep
\grouprow{OLMo 3 and OLMo 3.1, 32 billion parameters}
OLMo 3 32B Base & base & \hfrepo{allenai/Olmo-3-1125-32B}{c2b61dae89a1} \\
\rowcolor{rowgray}OLMo 3 32B Think SFT & SFT & \hfrepo{allenai/Olmo-3-32B-Think-SFT}{a6d7f3cf497c} \\
OLMo 3 32B Think DPO & DPO & \hfrepo{allenai/Olmo-3-32B-Think-DPO}{97604023f260} \\
\rowcolor{rowgray}OLMo 3 32B Think & RLVR & \hfrepo{allenai/Olmo-3-32B-Think}{f2edda15216e} \\
OLMo 3.1 32B Think & RLVR, continued & \hfrepo{allenai/Olmo-3.1-32B-Think}{832c3f543499} \\
\rowcolor{rowgray}OLMo 3.1 32B Instruct & RLVR & \hfrepo{allenai/Olmo-3.1-32B-Instruct}{ac0587e4a774} \\
\bottomrule
\end{tabular}
\end{table}

\textbf{Adapter branches.} The two-by-two design of post-training is applied to OLMo 3 Base and Llama 3.1 Base with low-rank adapters of rank 16 ($\alpha=32$, no dropout) on the query, key, value, output, gate, up and down projections of all 32 layers (40.0 million trainable parameters for OLMo 3 and 41.9 million for Llama 3.1), the base weights in bfloat16 and the adapters in float32, initialized identically for the four branches of a base. Each branch reads 16,384 episodes of 41 steps, rendered in R1 and context C0, about 2.8 million tokens, 32 per step for 512 steps, with AdamW ($\beta=(0.9,0.95)$, no weight decay), a learning rate of $2\times10^{-4}$ after 16 warm-up steps and gradient clipping at norm 1. The objectives are the log loss of the label tokens and the expected 0/1 reward of the two-symbol read-out. Branches A and C read the MIX stream in the same order, and B and D an F0ONLY stream that differs from it only where a \cls{rule} episode is replaced. Adapters are kept after 128, 256 and 512 steps, and those after 512 steps are read in rendering R1 under C0 and CF0, on the dense subset and on the null-equivalent pairs.

\textbf{Evaluation sets.} The default test set has 2,000 episodes per class at $T=40$; the sets at $T=16$, 24 and 32 share its seeds, so that a shorter episode is a prefix of the longer one with the same schedule, latent rule and noise, and the comparison across horizons is paired. The implicit small Transformers are read on all of it, and the reference, the triggers and the explicit learners on its first 300 episodes per class at each of the 120 price cells. The matched set has 1,000 groups of four episodes, one per class, with a shared query schedule and shared label flips; the implicit small Transformers are read on all of it, and the explicit learners and the reference on its first 150 groups. The null-equivalent pairs number 400 at each evidence level $m=1,\dots,5$, the designed states 270, each at 40 price cells, and the waiting set 182 state-cells with as many matched controls. The language models read the first 250 episodes per class of the test set, the first 40 per class also with a table of predictions after every step (the dense subset), the first 150 matched groups, each with a boundary-matched threshold episode added, whose boundary is the lower end of the rule and which differs from the \cls{rule} episode only above $u$, and 60 null-equivalent pairs per level, four pairs sharing demonstrations and symbols. In the explicit protocol they read 19,488 decision points on 1,137 histories: 300 of the test set, 750 of the matched set and 87 designed. The feature and shift worlds have test sets of 300 episodes per class and 300 matched groups each.

\textbf{Compute.} All training and the scoring of the 7B and 8B language models ran on one consumer GPU with 24 GB of memory; the 32B models and the runs with reasoning (Appendix~\ref{app:reasoning}) ran on one datacenter GPU with 80 GB; the reference policies, the exact Bayes predictors and the analyses ran on CPUs.

\FloatBarrier
\section{Statistical procedures}\label{app:statistics}

This appendix states what the decision regression of Section~\ref{sec:learners} identifies and how intervals and latencies are computed.

\textbf{Decision regression.} Let $B$ be the reference's expected gain in score from expanding now over never revising, summed over the remaining horizon, and $O$ the value of its best alternative to expanding, patching now or waiting, over never revising. The reference expands when $B>c_E+O$, that is, when $\log b+\log H-\log(c_E+O)>0$ with $b=B/H$, so the regression
\begin{equation}\label{eq:O-model}
P(\mathrm{expand})=\lambda+(1-2\lambda)\operatorname{logistic}\big(\beta_0+\beta_b\log b+\beta_H\log H-\beta_c\log(c_E+O)+\gamma S_{t-1}+\beta_t\log(1+t)\big),\qquad\lambda\le0.1,
\end{equation}
returns $(\beta_c,\beta_H,\gamma)/\beta_b=(1,1,0)$ for it, the time term absorbing statistics that accumulate with $t$. For a policy that expands when a statistic $g$ of $z=\log b$ exceeds a threshold $\theta$ of $u=\log(c_E+O)$, a local fit at its boundary gives $\beta_c/\beta_b=\partial_u\theta/\partial_z g$, the elasticity of the benefit it requires with respect to the effective price $c_E+O$, which is 0 for a fixed threshold, and $\gamma/\beta_b$ is the change in log benefit worth one nat of surprise. Over all states the fit is the projection solving $\mathbb E[X\{A-\operatorname{logistic}(X^\top\beta)\}]=0$ \citep{white1982maximum}, which depends on the states, so every policy is fitted on the same ones. The reference's decisions are separable and its unpenalized likelihood has no finite maximum \citep{albert1984existence}. A ridge of $10^{-4}$ on the standardized slopes sets the length of its coefficient vector and the data set its direction, which changes by less than $4\times10^{-4}$ over four orders of magnitude of the ridge, and Figure~\ref{fig:model-map} reports directions, while the lapse $\lambda$ absorbs decisions that do not depend on the predictors. The bounded reference has price sensitivity 1.00, horizon sensitivity 1.03 and surprise leakage 0.05, since its backward induction solves for price and horizon and what it lacks is evidence. The e-process trigger has $-0.02$, $-0.14$ and 0.31 with a fixed threshold and 0.72, 0.62 and 0.12 when re-tuned in every cell.

\textbf{Intervals and latency.} Intervals are 95\% percentile bootstrap intervals over evaluation episodes, each a cluster with its prefixes, horizons and price cells (1,000 resamples for the decision regression), and, for the post-training contrasts of Figure~\ref{fig:post-training}, over learners and then episodes. Regret is paired, every policy being scored on the reference's episodes with the class prior as weights. Latency in \cls{rule} episodes is counted from the first query of an input above $u$, episodes without expansion are right-censored at their end, and medians and restricted means come from the Kaplan--Meier estimator \citep{kaplan1958nonparametric}.

% Appendix P: on the framing.
\FloatBarrier
\section{On the framing}\label{app:framing}

This appendix gives the computational reading of the distinction recalled in Section~\ref{sec:related} between inference within a framework, local repair of the framework and replacement of its basis. In SRE the three differ in what they compute and what they cost. Normal inference, in Kuhn's sense \citep{kuhn1962structure}, moves probability within $\mathcal F_0$, including the change points that $E_0$ tracks, at no price. Local repair adds a slot for one input at price $c_p$. An exception list pays $\log(1/\rho)+\log(K-1)$ in description length for every anomalous input (Equation~\eqref{eq:C-mdl}), and the gain of each slot stays at its own input, as the repairs of a degenerating programme grow with the anomalies and predict nothing new \citep{lakatos1970falsification}. Expansion pays $c_E$ once for a rule that predicts the anomalies at inputs not yet queried, as a progressive programme predicts facts not yet seen, and over a patch its net gain is $H\sum_{x\in\mathcal N}g_x-(c_E-c_p)$ (Equation~\eqref{eq:C-boundary}). The three are scored by their future predictions against their prices, a prequential criterion \citep{dawid1984present}, so the boundary between repair and revision is a price, and repair is optimal with a small region, a high price of expansion or a short horizon. The correspondence holds for a single learner that decides when to pay for access to a family the environment already contains, and not for the communities and revolutions of the historical accounts.

\FloatBarrier
\section{Scale and test-time reasoning}\label{app:reasoning}

This appendix reads the two protocols of Section~\ref{sec:learners} at a larger scale and with room to reason. Six OLMo 3 models of 32 billion parameters are read under the implicit and the explicit protocol exactly as the 7B models (Appendix~\ref{app:reasoning-32b}), and three models, Qwen 3 8B, OLMo 3 Think and OLMo 3 Think SFT, are allowed to think before they answer, under a thinking budget and until they finish (Appendix~\ref{app:reasoning-e11}). The second part asks what the constant decision of Section~\ref{sec:llm-decision} leaves open: whether a decision that does not use the evidence at the history level begins to use it once the model has reasoned, and whether the weighing of a stated gain comes closer to the reference's.

\subsection{The 32B end points}\label{app:reasoning-32b}

The models are OLMo 3 32B Base, the SFT, DPO and RLVR end points of its Think chain, OLMo 3.1 32B Think, which continues the RLVR of the Think end point, and OLMo 3.1 32B Instruct. They read the three renderings and three contexts of the implicit protocol and the chat format with the path read-out of the explicit protocol (Appendix~\ref{app:prompts}), and every implicit cell passes the validity condition.

\textbf{Implicit protocol.} Scale leaves the shape of the response unchanged (Table~\ref{tab:scale-implicit}). As for the 7B models, no curve of the matched-pair $\kappa$ of any 32B model lies on a member of the family indexed by the prior, the step at the boundary has an interval that includes zero in every model, and the response extends well below $u$ ($x\le u-3$), with and without demonstrations in the context. What scale changes is the calibration. At every stage of the Think chain the calibration slope $s_0$ with the test episode alone is lower at 32B, by 0.27 to 0.39, and with demonstrations in the context the lift comes earlier, by 3.1 to 4.3 steps.

\begin{table}[h]
\caption{The 32B end points against their 7B counterparts in the implicit protocol, with the test episode alone. Matched-pair $\kappa$ at $m=1$, 3 and 5 anomalous inputs; surprise leakage; the step at the boundary in log-odds with its 95\% interval over episodes; the response well below $u$ (at $l\le x\le u-3$, Appendix~\ref{app:active-space}); the false lift on the boundary-matched threshold episode (Appendix~\ref{app:training}); the calibration slope $s_0$. The 7B counterpart of OLMo 3.1 32B Think is OLMo 3 Think. The Bayes predictor $E_1$ has $\kappa=1$ at every $m$, leakage 0, a step of 1.12\,\ci{0.88}{1.37}, a response of $-0.12$ well below $u$ and a false lift of 0.02.}
\label{tab:scale-implicit}
\centering\tablestyle
\setcounter{tabcols}{9}
\begin{tabularx}{\textwidth}{>{\raggedright\arraybackslash}X S[table-format=1.2] S[table-format=1.2] S[table-format=1.2] S[table-format=1.3] S[table-format=1.2,table-space-text-post=\ci{-0.82}{1.08}] S[table-format=1.2] S[table-format=1.2] S[table-format=1.3]}
\toprule
 & \multicolumn{3}{c}{\hd{matched-pair $\kappa$}} & & & & & \\
\cmidrule(lr){2-4}
\hd{model} & {\hd{$m=1$}} & {\hd{$m=3$}} & {\hd{$m=5$}} & {\hd{leakage}} & {\hd{boundary step}} & {\hdtwo*{well}{below $u$}} & {\hdtwo*{false}{lift}} & {\hd{$s_0$}} \\
\midrule
\grouprow{Base}
OLMo 3 32B Base & 0.30 & 0.29 & 0.44 & 0.137 & 0.25 {\ci{-0.51}{1.07}} & 2.32 & 0.28 & 0.614 \\
\rowcolor{rowgray}OLMo 3 Base (7B) & 0.41 & 0.40 & 0.58 & 0.110 & 0.24 {\ci{-0.52}{1.08}} & 2.25 & 0.26 & 0.717 \\
\groupsep
\grouprow{Think SFT}
OLMo 3 32B Think SFT & 0.28 & 0.33 & 0.49 & 0.149 & 0.22 {\ci{-0.63}{1.10}} & 2.28 & 0.32 & 0.629 \\
\rowcolor{rowgray}OLMo 3 Think SFT (7B) & 0.27 & 0.28 & 0.37 & 0.136 & 0.11 {\ci{-0.81}{1.05}} & 2.26 & 0.32 & 0.903 \\
\groupsep
\grouprow{Think DPO}
OLMo 3 32B Think DPO & 0.25 & 0.31 & 0.49 & 0.156 & 0.21 {\ci{-0.62}{1.07}} & 2.14 & 0.33 & 0.608 \\
\rowcolor{rowgray}OLMo 3 Think DPO (7B) & 0.28 & 0.28 & 0.37 & 0.137 & 0.09 {\ci{-0.80}{1.01}} & 2.14 & 0.34 & 0.920 \\
\groupsep
\grouprow{Think}
OLMo 3 32B Think & 0.25 & 0.32 & 0.49 & 0.158 & 0.18 {\ci{-0.64}{1.05}} & 2.16 & 0.34 & 0.607 \\
\rowcolor{rowgray}OLMo 3.1 32B Think & 0.16 & 0.27 & 0.45 & 0.169 & 0.14 {\ci{-0.72}{1.03}} & 2.07 & 0.37 & 0.562 \\
OLMo 3 Think (7B) & 0.27 & 0.29 & 0.40 & 0.135 & 0.09 {\ci{-0.82}{1.05}} & 2.20 & 0.33 & 0.951 \\
\groupsep
\grouprow{Instruct}
OLMo 3.1 32B Instruct & 0.11 & 0.23 & 0.46 & 0.181 & 0.13 {\ci{-0.69}{0.96}} & 1.86 & 0.37 & 0.502 \\
\rowcolor{rowgray}OLMo 3 Instruct (7B) & 0.17 & 0.24 & 0.37 & 0.193 & 0.01 {\ci{-0.71}{0.75}} & 1.79 & 0.36 & 0.624 \\
\bottomrule
\end{tabularx}
\end{table}

\textbf{Explicit protocol.} At the history level the decision of the base model stays constant at 32B, and the four post-trained end points are partial (Table~\ref{tab:scale-explicit}): their probability of expanding stays flat across the classes, and their benefit coefficient is small, with an interval that excludes zero, 0.022 to 0.029 for the three Think-chain end points, as for OLMo 3 Think at 7B, and 0.11 for OLMo 3.1 32B Instruct, where the coefficient of the base model is zero. Within each model the probability of expanding differs by at most 0.005 between the four classes and the boundary-matched threshold histories of Appendix~\ref{app:training} (0.035 for OLMo 3.1 32B Instruct) and between a \cls{rule} history and its boundary-matched threshold history. At the gain level scale changes the valuation of the stated gain: the Think chain weighs the price nearly in the reference's proportion and the horizon in part, the base model weighs both in part where its 7B counterpart responds to the gain alone, and OLMo 3.1 32B Instruct weighs the price in part and not the horizon. These ratios are identified by the rule of Appendix~\ref{app:reasoning-e11}, the price sensitivity of the Think chain staying within 0.65 to 0.95 across three subsets of the price levels, against 0.81 to 0.99 on all six, and the responses behind them are small: as $c_E$ rises from 0.5 to 16, the probability of expanding of the Think chain falls by 0.05 to 0.07 and the reference's from 0.77 to 0.07. OLMo 3.1 32B Think has no valid decision read-out, since its first token at the cue opens a sentence rather than naming an action, and it enters only the implicit results.

\begin{table}[h]
\caption{The 32B end points against their 7B counterparts in the explicit protocol (chat format with the path read-out, rendering R1, order A, natural points; 95\% intervals from the paired cluster bootstrap). History level: the benefit coefficient $\beta_b$, the spread of the probability of expanding over the four classes and the boundary-matched threshold histories, its gap between \cls{rule} histories and their boundary-matched threshold histories, and the change in the probability of deferring under an informative announcement; the failure-level column gives $\beta_b$ at that level, in grey where its 95\% interval from the cluster bootstrap includes zero. Gain level: price and horizon sensitivity, n.i.\ where not identified. The verdicts are those of Section~\ref{sec:llm-decision} at the history level (constant when both conditions of Appendix~\ref{app:prompts} hold, partial when one of them holds) and at the gain level (responds alone when price and horizon sensitivity are both below 0.3, weighed in the reference's proportion when both are above 0.7, mixed otherwise, or n.i.; the horizon sensitivity of OLMo 3 32B Think, printed as 0.70, is 0.6997). OLMo 3 Think DPO was not read at 7B.}
\label{tab:scale-explicit}
\centering\tablestyle\setlength{\tabcolsep}{2.8pt}
\setcounter{tabcols}{10}
\begin{tabularx}{\textwidth}{>{\raggedright\arraybackslash}X
  S[table-format=-1.3,table-space-text-post=\ci{-0.053}{-0.259}]
  S[table-format=1.3] S[table-format=-1.3] S[table-format=+1.3]
  S[table-format=-1.3]
  S[table-format=-1.2,table-space-text-post=\ci{-0.11}{-0.00}]
  S[table-format=-1.2,table-space-text-post=\ci{-0.35}{-0.18}]
  l l}
\toprule
 & \multicolumn{4}{c}{\hd{history level}} & {\hd{failure level}} & \multicolumn{2}{c}{\hd{gain level}} & \multicolumn{2}{c}{\hd{verdict}} \\
\cmidrule(lr){2-5}\cmidrule(lr){6-6}\cmidrule(lr){7-8}\cmidrule(lr){9-10}
\hd{model} & {\hd{$\beta_b$}} & {\hdtwo{class}{spread}} & {\hdtwo{\cls{rule} $-$}{threshold}} & {\hd{defer $\Delta$}} & {\hd{$\beta_b$}} & {\hd{price}} & {\hd{horizon}} & \hd{history} & \hd{gain} \\
\midrule
\grouprow{OLMo 3 Base}
32B & 0.006 {\ci{-0.002}{0.015}} & 0.005 & -0.005 & +0.003 & 0.052 & 0.63 {\ci{0.58}{0.68}} & 0.50 {\ci{0.42}{0.59}} & constant & mixed \\
\rowcolor{rowgray}7B & -0.015 {\ci{-0.023}{-0.008}} & 0.001 & 0.000 & -0.002 & 0.014 & -0.05 {\ci{-0.11}{-0.00}} & -0.26 {\ci{-0.35}{-0.18}} & constant & responds alone \\
\groupsep
\grouprow{OLMo 3 Think SFT}
32B & 0.025 {\ci{0.014}{0.038}} & 0.004 & -0.004 & +0.003 & \color{uncink}0.009 & 0.81 {\ci{0.75}{0.88}} & 0.63 {\ci{0.53}{0.76}} & partial & mixed \\
\rowcolor{rowgray}7B & 0.001 {\ci{-0.008}{0.010}} & 0.002 & -0.003 & -0.001 & 0.034 & {\nicell} & {\nicell} & constant & n.i. \\
\groupsep
\grouprow{OLMo 3 Think DPO}
32B & 0.029 {\ci{0.016}{0.045}} & 0.005 & -0.004 & +0.005 & \color{uncink}-0.008 & 0.99 {\ci{0.91}{1.09}} & 0.75 {\ci{0.64}{0.89}} & partial & weighed \\
\groupsep
\grouprow{OLMo 3 Think}
32B & 0.022 {\ci{0.009}{0.037}} & 0.005 & -0.003 & +0.005 & \color{uncink}0.010 & 0.98 {\ci{0.89}{1.09}} & 0.70 {\ci{0.57}{0.86}} & partial & mixed \\
\rowcolor{rowgray}7B & 0.021 {\ci{0.004}{0.037}} & 0.004 & -0.004 & -0.001 & 0.053 & {\nicell} & {\nicell} & partial & n.i. \\
\groupsep
\grouprow{Instruct (OLMo 3.1 at 32B, OLMo 3 at 7B)}
32B & 0.108 {\ci{0.014}{0.218}} & 0.035 & 0.004 & -0.001 & 5.054 & 0.54 {\ci{0.51}{0.59}} & -0.06 {\ci{-0.26}{0.16}} & partial & mixed \\
\rowcolor{rowgray}7B & 0.065 {\ci{-0.053}{0.259}} & 0.007 & 0.002 & +0.034 & \color{uncink}0.369 & 0.06 {\ci{0.03}{0.09}} & -0.15 {\ci{-0.35}{0.07}} & constant & responds alone \\
\bottomrule
\end{tabularx}
\end{table}

\subsection{Test-time reasoning}\label{app:reasoning-e11}

\textbf{Design.} A sensitivity ratio is interpreted only where the response to the stated gain is identified, that is, where the interval of the benefit coefficient excludes zero and the ratio is stable across subsets of the price levels. Elsewhere we report the raw slopes and the probability of expanding, and the tables mark the ratio n.i. The decision points, the three levels and the read-out are those of Appendix~\ref{app:prompts}, with the four action words read at the cue after the model has reasoned. Thinking is switched on through the chat template of each model, and the chains are sampled at the settings of the model cards (temperature 0.6, top-$p$ 0.95, and top-$k$ 20 for Qwen 3 8B). On these prompts a model writes about ten thousand tokens before it finishes its reasoning, and two arms follow. In the budget arm the thinking is closed after $k=256$, 512, 1,024 or 2,048 tokens with the sentence that Qwen's documentation prescribes for a thinking budget \citep[Section~4.3]{yang2025qwen3}, the closing tag and the cue; the budgets of a chain are prefixes of one chain, and thinking off is $k=0$. In the natural-finish arm the chain runs until the model writes its own closing tag, and a chain that reaches the cap is closed with the sentence. Two criteria were fixed before the runs. At the history level, a benefit coefficient whose interval excludes zero together with a gap of at least 0.10 between the probability of expanding on \cls{rule} histories and on their boundary-matched threshold histories counts as \emph{reasoning uses the evidence}. At the gain level, price and horizon sensitivity both within $[0.7,1.3]$ count as \emph{weighing in the reference's proportion}.

\subsubsection{Qwen 3 8B under a thinking budget}

The budget arm reads Qwen 3 8B on 1,984 decision points at the three levels and in both action orders, with four sampled chains and one greedy chain per prompt. No chain finishes its reasoning within 2,048 tokens, so every read-out of this arm follows a forced close, and the read-out is valid in every cell.

\emph{At the history level no budget brings the evidence into the decision.} At every budget and in both orders the probability of expanding sits at its floor, 0.016 to 0.038, and is the same in the four classes and the boundary-matched threshold histories, and the gap between \cls{rule} histories and their boundary-matched threshold histories is below 0.01 (Table~\ref{tab:qwen-budget}). At this floor the benefit coefficient of the lapse model is unreliable, and the raw slope of the probability of expanding on the benefit carries the verdict: 0.001 to 0.011 per standard deviation, with every interval including zero.

\emph{At the gain level the budget moves the weighing to the reference's proportion.} Price and horizon sensitivity rise with the budget and lie within $[0.7,1.3]$ in both orders at 2,048 tokens, in one order at 1,024 and in neither below (Table~\ref{tab:qwen-budget}).

\begin{table}[h]
\caption{Qwen 3 8B under a thinking budget: the gap in the probability of expanding between \cls{rule} histories and their boundary-matched threshold histories at the history level, and price and horizon sensitivity at the gain level, by budget and action order. Thinking off is the budget 0. Every interval of the gap lies within $[-0.006,0.007]$; the sensitivities carry 95\% intervals from the cluster bootstrap.}
\label{tab:qwen-budget}
\centering\tablestyle
\setcounter{tabcols}{7}
\begin{tabular}{l S[table-format=-1.3,table-column-width=52pt] S[table-format=-1.3,table-column-width=52pt] S[table-format=-1.2,table-space-text-post=\ci{-0.34}{0.00}] S[table-format=-1.2,table-space-text-post=\ci{-0.34}{0.00}] S[table-format=-1.2,table-space-text-post=\ci{-0.52}{-0.22}] S[table-format=-1.2,table-space-text-post=\ci{-0.52}{-0.22}]}
\toprule
 & \multicolumn{2}{c}{\hd{history level: \cls{rule} $-$ threshold}} & \multicolumn{2}{c}{\hd{gain level, order A}} & \multicolumn{2}{c}{\hd{gain level, order B}} \\
\cmidrule(lr){2-3}\cmidrule(lr){4-5}\cmidrule(lr){6-7}
\hdtwo*[l]{thinking}{budget} & {\hd{order A}} & {\hd{order B}} & {\hd{price}} & {\hd{horizon}} & {\hd{price}} & {\hd{horizon}} \\
\midrule
off & -0.001 & 0.000 & 0.37 {\ci{0.23}{0.67}} & -0.14 {\ci{-0.34}{0.00}} & 0.46 {\ci{0.33}{0.72}} & -0.34 {\ci{-0.52}{-0.22}} \\
\rowcolor{rowgray}256 & -0.002 & -0.001 & 0.14 {\ci{0.03}{0.34}} & -0.08 {\ci{-0.20}{0.07}} & 0.43 {\ci{0.33}{0.58}} & 0.08 {\ci{-0.02}{0.20}} \\
512 & 0.000 & 0.000 & 0.46 {\ci{0.22}{0.84}} & 0.32 {\ci{0.11}{0.67}} & 0.80 {\ci{0.66}{0.97}} & 0.59 {\ci{0.41}{0.77}} \\
\rowcolor{rowgray}1,024 & 0.000 & 0.001 & 0.79 {\ci{0.56}{1.09}} & 0.65 {\ci{0.43}{0.93}} & 0.88 {\ci{0.74}{1.04}} & 0.86 {\ci{0.72}{1.00}} \\
2,048 & 0.003 & 0.000 & 0.92 {\ci{0.70}{1.20}} & 0.84 {\ci{0.60}{1.14}} & 0.84 {\ci{0.76}{0.95}} & 0.83 {\ci{0.74}{0.98}} \\
\bottomrule
\end{tabular}
\end{table}

\subsubsection{Qwen 3 8B reasoning until it finishes}

The natural-finish arm generates one sampled chain per prompt, up to a cap of 24,576 tokens, on the 600 matched points in each action order and on the same-history points, 366 at the history level and 1,098 at the gain level. Almost all chains finish by themselves (Table~\ref{tab:qwen-closing}). The first choice of the read-out is the action that the chain names last in 99.2\% of the chains, so the read-out reports the conclusion the model states.

\begin{table}[h]
\caption{How Qwen 3 8B finishes its reasoning in the natural-finish arm, by set. Share of chains that finish within the cap of 24,576 tokens, with its 95\% interval, and the median and 90th percentile of the number of thinking tokens at the finish.}
\label{tab:qwen-closing}
\centering\tablestyle
\setcounter{tabcols}{5}
\begin{tabular}{l S[table-format=4.0] S[table-format=1.3,table-space-text-post=\ci{0.989}{0.998}] S[table-format=5.0,table-column-width=37pt] S[table-format=5.0]}
\toprule
 & & & \multicolumn{2}{c}{\hd{thinking tokens at the finish}} \\
\cmidrule(lr){4-5}
\hd{level, action order} & {\hd{chains}} & {\hd{share finished}} & {\hd{median}} & {\hd{90th percentile}} \\
\midrule
\grouprow{Matched groups}
history level, order A & 600 & 0.930 {\ci{0.913}{0.947}} & 14748 & 21848 \\
\rowcolor{rowgray}history level, order B & 600 & 0.963 {\ci{0.947}{0.979}} & 13452 & 19993 \\
\groupsep
\grouprow{Same-history design}
history level, order A & 366 & 0.970 {\ci{0.952}{0.985}} & 14124 & 20656 \\
\rowcolor{rowgray}gain level, order A & 1098 & 0.994 {\ci{0.989}{0.998}} & 8035 & 12325 \\
\bottomrule
\end{tabular}
\end{table}

\emph{The history level does not meet the criterion.} The gap between \cls{rule} histories and their boundary-matched threshold histories is 0.060 \ci{0.008}{0.115} in order A and 0.028 \ci{-0.036}{0.090} in order B, against 0.10 (Table~\ref{tab:dose}). The \cls{rule} history moves the model locally: in order A it takes 0.26 of the probability off stay and gives 0.10 to patch, 0.10 to defer and 0.06 to expand (Table~\ref{tab:qwen-natural} gives the probabilities pooled over the two orders). On \cls{rule} histories the probability of expanding barely moves with the price, 0.10, 0.09 and 0.05 at $c_E=1$, 4 and 16, where the reference's falls from 0.93 to 0.03.

\begin{table}[h]
\caption{Qwen 3 8B reasoning until it finishes, history level, matched groups pooled over the two action orders (chains that finish by themselves): the probability of each action by class, the number of chains whose first choice is expand, and the share of points at which the reference's best action is expand. The last row holds the boundary-matched threshold histories of the \cls{rule} histories. The probability of expanding carries its 95\% interval.}
\label{tab:qwen-natural}
\centering\tablestyle
\setcounter{tabcols}{8}
\begin{tabular}{l S[table-format=3.0] S[table-format=1.3] S[table-format=1.3] S[table-format=1.3,table-space-text-post=\ci{0.043}{0.120}] S[table-format=1.3] r S[table-format=1.2]}
\toprule
 & & \multicolumn{4}{c}{\hd{probability of the action}} & & \\
\cmidrule(lr){3-6}
\hd{class} & {\hd{chains}} & {\hd{stay}} & {\hd{patch}} & {\hd{expand}} & {\hd{defer}} & \hdtwo*{first choice}{expand} & {\hdtwo*{reference}{expands}} \\
\midrule
\cls{noise} (in-family) & 228 & 0.569 & 0.360 & 0.022 {\ci{0.004}{0.045}} & 0.049 & 5 of 228 & 0.02 \\
\rowcolor{rowgray}\cls{exception} & 226 & 0.489 & 0.421 & 0.071 {\ci{0.043}{0.098}} & 0.019 & 16 of 226 & 0.12 \\
\cls{drift} & 225 & 0.514 & 0.396 & 0.044 {\ci{0.021}{0.069}} & 0.045 & 10 of 225 & 0.01 \\
\rowcolor{rowgray}\cls{rule} & 228 & 0.414 & 0.404 & 0.079 {\ci{0.043}{0.120}} & 0.102 & 18 of 228 & 0.53 \\
\clsring{rule} matched threshold & 229 & 0.588 & 0.346 & 0.035 {\ci{0.014}{0.060}} & 0.032 & 8 of 229 & 0.02 \\
\bottomrule
\end{tabular}
\end{table}

\emph{The gain level meets its criterion.} Price sensitivity is 1.12 \ci{1.02}{1.20} and horizon sensitivity 1.24 \ci{1.08}{1.36}, against 0.37 and $-0.14$ with thinking off on these points ($+0.14$ on the 5,400 natural points of Appendix~\ref{app:prompts}).

\begin{figure}[t]
  \centering
  \includegraphics[width=\textwidth]{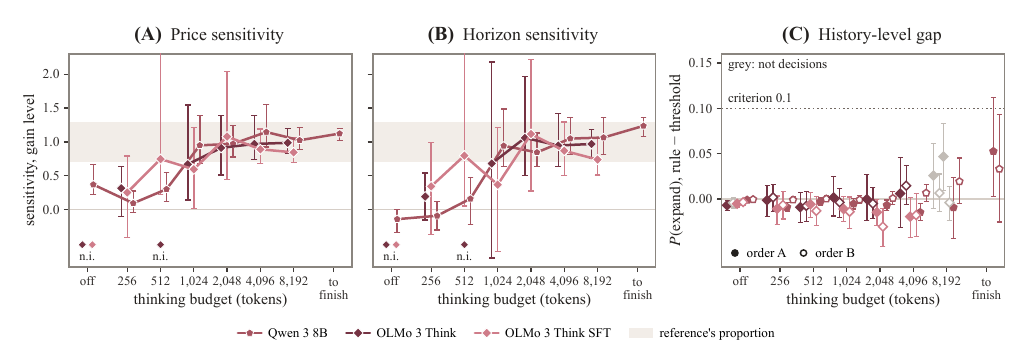}
  \caption{\textbf{Reasoning recovers the valuation of a stated gain, but not structural evidence from the history.} (A) Price and (B) horizon sensitivity at the gain level against the thinking budget, with 95\% intervals and the reference's proportion, 0.7 to 1.3, as a band; n.i.\ marks a budget at which the ratio is not identified. (C) The history-level gap in the probability of expanding between \cls{rule} histories and their boundary-matched threshold histories, in orders A (filled) and B (hollow), against the criterion of 0.10. Grey points are not decisions: at the cap of 8,192 tokens the read-out of the two OLMo 3 Think models follows the order in which the actions are listed, and one thinking-off cell fails the validity condition. The rightmost point of each panel is Qwen 3 8B reasoning until it finishes.}
  \label{fig:app-R-dose}
\end{figure}

\emph{Dose.} Read at prefixes of the same chains, the gain-level sensitivities enter the reference's proportion from 1,024 tokens and stay there, while the history-level gap stays between $-0.015$ and $+0.019$ at every forced close up to 8,192 tokens (Table~\ref{tab:dose} and Figure~\ref{fig:app-R-dose}). The local response at the history level appears only when the model finishes by itself.

\begin{table}[t!]
\caption{The dose of reasoning for the three models. History level: the gap in the probability of expanding between \cls{rule} histories and their boundary-matched threshold histories (matched groups, one chain per point), in each action order. Gain level: price and horizon sensitivity on the 1,098 same-history points, order A; n.i.: not identified. Qwen 3 8B is read at prefixes of the chains of its natural-finish arm and after it finishes by itself, the OLMo 3 Think models at prefixes of their chains up to their cap of 8,192 tokens. 95\% intervals from the cluster bootstrap; the thinking-off order-B history-level cell of OLMo 3 Think fails the validity condition.}
\label{tab:dose}
\centering\tablestyle
\setcounter{tabcols}{5}
\begin{tabular}{l S[table-format=-1.3,table-space-text-post=\ci{-0.046}{-0.027}] S[table-format=-1.3,table-space-text-post=\ci{-0.052}{-0.011}] S[table-format=1.2,table-space-text-post=\ci{-0.42}{0.79}] S[table-format=-1.2,table-space-text-post=\ci{-0.72}{2.18}]}
\toprule
 & \multicolumn{2}{c}{\hd{history level: \cls{rule} $-$ threshold}} & \multicolumn{2}{c}{\hd{gain level: sensitivity}} \\
\cmidrule(lr){2-3}\cmidrule(lr){4-5}
\hdtwo*[l]{thinking}{budget} & {\hd{order A}} & {\hd{order B}} & {\hd{price}} & {\hd{horizon}} \\
\midrule
\grouprow{Qwen 3 8B}
off & -0.001 {\ci{-0.002}{0.000}} & 0.000 {\ci{-0.002}{0.001}} & 0.37 {\ci{0.23}{0.67}} & -0.14 {\ci{-0.34}{0.00}} \\
\rowcolor{rowgray}256 & -0.009 {\ci{-0.014}{-0.004}} & -0.001 {\ci{-0.004}{0.002}} & 0.09 {\ci{-0.04}{0.27}} & -0.09 {\ci{-0.30}{0.12}} \\
512 & -0.003 {\ci{-0.007}{0.001}} & 0.000 {\ci{-0.003}{0.003}} & 0.30 {\ci{0.12}{0.55}} & 0.16 {\ci{-0.22}{0.47}} \\
\rowcolor{rowgray}1,024 & -0.005 {\ci{-0.011}{0.000}} & -0.001 {\ci{-0.005}{0.004}} & 0.95 {\ci{0.68}{1.39}} & 0.94 {\ci{0.64}{1.49}} \\
2,048 & -0.007 {\ci{-0.013}{-0.001}} & 0.001 {\ci{-0.007}{0.008}} & 0.97 {\ci{0.78}{1.24}} & 0.85 {\ci{0.63}{1.13}} \\
\rowcolor{rowgray}4,096 & -0.015 {\ci{-0.024}{-0.004}} & 0.007 {\ci{-0.002}{0.015}} & 1.15 {\ci{0.93}{1.56}} & 1.05 {\ci{0.81}{1.37}} \\
8,192 & -0.009 {\ci{-0.046}{0.027}} & 0.019 {\ci{-0.004}{0.044}} & 1.02 {\ci{0.89}{1.21}} & 1.06 {\ci{0.82}{1.36}} \\
\rowcolor{rowgray}until it finishes & 0.060 {\ci{0.008}{0.115}} & 0.028 {\ci{-0.036}{0.090}} & 1.12 {\ci{1.02}{1.20}} & 1.24 {\ci{1.08}{1.36}} \\
\groupsep
\grouprow{OLMo 3 Think}
off & -0.007 {\ci{-0.013}{-0.001}} & -0.004 {\ci{-0.009}{0.001}} & {\nicell} & {\nicell} \\
\rowcolor{rowgray}256 & -0.001 {\ci{-0.019}{0.017}} & 0.002 {\ci{-0.012}{0.016}} & 0.31 {\ci{-0.11}{0.63}} & 0.19 {\ci{-0.16}{0.54}} \\
512 & -0.009 {\ci{-0.027}{0.008}} & -0.008 {\ci{-0.024}{0.009}} & {\nicell} & {\nicell} \\
\rowcolor{rowgray}1,024 & 0.002 {\ci{-0.019}{0.024}} & -0.004 {\ci{-0.020}{0.013}} & 0.67 {\ci{0.14}{1.55}} & 0.68 {\ci{-0.72}{2.18}} \\
2,048 & 0.000 {\ci{-0.025}{0.026}} & -0.005 {\ci{-0.021}{0.011}} & 0.91 {\ci{0.51}{1.39}} & 1.06 {\ci{0.50}{1.97}} \\
\rowcolor{rowgray}4,096 & 0.006 {\ci{-0.033}{0.042}} & 0.015 {\ci{-0.006}{0.035}} & 0.97 {\ci{0.74}{1.39}} & 0.95 {\ci{0.64}{1.42}} \\
8,192 & 0.026 {\ci{-0.011}{0.061}} & 0.007 {\ci{-0.013}{0.028}} & 0.99 {\ci{0.83}{1.19}} & 0.97 {\ci{0.75}{1.18}} \\
\groupsep
\grouprow{OLMo 3 Think SFT}
off & -0.005 {\ci{-0.008}{-0.002}} & -0.003 {\ci{-0.007}{0.000}} & {\nicell} & {\nicell} \\
\rowcolor{rowgray}256 & -0.011 {\ci{-0.026}{0.003}} & -0.007 {\ci{-0.022}{0.008}} & 0.25 {\ci{-0.42}{0.79}} & 0.34 {\ci{-0.37}{0.99}} \\
512 & -0.006 {\ci{-0.020}{0.007}} & -0.013 {\ci{-0.028}{0.002}} & 0.75 {\ci{0.22}{4.03}} & 0.80 {\ci{0.05}{4.13}} \\
\rowcolor{rowgray}1,024 & -0.011 {\ci{-0.026}{0.004}} & -0.014 {\ci{-0.032}{0.003}} & 0.60 {\ci{0.01}{1.22}} & 0.37 {\ci{-0.62}{1.22}} \\
2,048 & -0.015 {\ci{-0.030}{0.001}} & -0.030 {\ci{-0.052}{-0.011}} & 1.08 {\ci{0.45}{2.04}} & 1.12 {\ci{0.27}{2.22}} \\
\rowcolor{rowgray}4,096 & -0.019 {\ci{-0.042}{0.003}} & -0.018 {\ci{-0.044}{0.005}} & 0.89 {\ci{0.68}{1.19}} & 0.87 {\ci{0.50}{1.29}} \\
8,192 & 0.047 {\ci{0.009}{0.086}} & -0.004 {\ci{-0.023}{0.016}} & 0.85 {\ci{0.70}{1.06}} & 0.74 {\ci{0.51}{1.05}} \\
\bottomrule
\end{tabular}
\end{table}

\emph{What the reasoning says.} The words that could name a structure, \emph{interval}, \emph{region}, \emph{range}, \emph{cluster}, \emph{block}, \emph{between positions} and \emph{exceptions}, occur in 54\% to 73\% of the chains of every class and describe the single boundary: 93\% of the mentions of \emph{between positions} follow \emph{boundary} within 80 characters, and \emph{interval} and \emph{block} occur in under 1\% of the chains. \cls[Rule]{rule} chains name expand near their end more often than the chains of the boundary-matched threshold histories, 0.74 against 0.57, and rarely conclude it, 0.09 against 0.04. The two endings below, from \cls{rule} histories at $c_E=1$, show the two conclusions.

\begin{outputbox}[unbreakable,title={end of a chain of Qwen 3 8B that finishes by itself: a rule history ($c_E=1$) that concludes expand}]
\begin{Verbatim}[fontsize=\scriptsize,breaklines=true]
[...] know the label of position 4 is Z, and it's already part of the observed data, the observe action might not help. Therefore, the best action is expand.

But wait, the patch action already patches position 14, which is a position that disagrees with the default predictor. If we choose to patch, we get a better score than continue, but expand is better.

Therefore, the final answer is expand.
\end{Verbatim}
\end{outputbox}

\begin{outputbox}[unbreakable,title={end of a chain of Qwen 3 8B that finishes by itself: a rule history ($c_E=1$) that concludes patch}]
\begin{Verbatim}[fontsize=\scriptsize,breaklines=true]
[...] can predict Q for it, which is correct, and the default predictor ignores it. This would improve the score for future rounds involving position 16.

The cost is 0.5, which is low.

Alternatively, expanding might be better, but with a cost of 1.

Given the uncertainty, I think the best action is to patch, as it addresses a specific disagreement with a low cost.

Therefore, the answer is patch.
\end{Verbatim}
\end{outputbox}

\subsubsection{OLMo 3 Think under a long budget}\label{app:reasoning-olmo}

OLMo 3 Think and OLMo 3 Think SFT are read on the same sets, with one sampled chain per prompt up to a cap of 8,192 tokens. No history-level chain and few gain-level chains (5 and 66 of 1,098) finish within the cap, so their read-outs follow a forced close.

\emph{At the history level the read-out at the cap is not a decision.} At the cut, 95\% of the chains name no action in their last 300 characters, and the read-out of such an unfinished chain follows the order in which the actions are listed: expand is the first choice in 72\% to 98\% of the chains of every class in order A, and stay in 94\% to 99\% in order B, where the list is reversed. Since these chains do not finish within the cap, the evidence criterion cannot be evaluated for these models.

\emph{At the gain level the budget brings the ratios into the reference's proportion.} In order A, the only order with a gain-level cell, price and horizon sensitivity at 8,192 tokens are 0.99 \ci{0.83}{1.19} and 0.97 \ci{0.75}{1.18} for OLMo 3 Think and 0.85 \ci{0.70}{1.06} and 0.74 \ci{0.51}{1.05} for OLMo 3 Think SFT (Table~\ref{tab:dose}).

With thinking off the ratios of the two models are not identified. Their slopes on the gain, the price and the horizon are 0.05 to 0.13 per $e$-fold, and the probability of expanding is nearly flat: for OLMo 3 Think it falls from 0.79 to 0.73 as $c_E$ rises from 0.5 to 16 on the natural points, where the reference's falls from 0.77 to 0.07. The fitted price sensitivity then follows the price levels of the grid, 0.38 with $c_E\in\{0.5,2,8\}$, $-0.04$ with $\{0.5,1,2,4\}$ and 1.66 with $\{2,4,8,16\}$. With reasoning the slope on the benefit rises tenfold or more, and the ratios are stable across price subsets, 0.72 and 1.07 for $c_E\in\{0.5,2\}$ and $\{2,8\}$. On the 1,098 same-history points, in order A, the slope per $e$-fold of the benefit rises from 0.09 \ci{0.02}{0.16} with thinking off to 1.06 \ci{0.88}{1.31} at 8,192 tokens for OLMo 3 Think and from 0.039 \ci{-0.003}{0.073} to 0.79 \ci{0.59}{1.06} for OLMo 3 Think SFT, and for Qwen 3 8B from 0.27 \ci{0.16}{0.42} to 19.5 \ci{15.0}{60.4} when it finishes by itself (0.84 \ci{0.69}{1.06} at a prefix of 2,048 tokens). In the coordinates of Figure~\ref{fig:model-map}A the three models with reasoning lie on the reference's direction, OLMo 3 Think at (0.59, 0.58), OLMo 3 Think SFT at (0.67, 0.56) and Qwen 3 8B at (0.51, 0.58), against (0.59, 0.57) for the reference.

\subsection{Conclusions}

Scale leaves the shape of the predictive response and the history-level decision failure unchanged, and changes the calibration and the valuation of a stated gain. The 32B models respond to structural evidence with the shape of the 7B models, and their decisions at the history level do not separate a \cls{rule} history from its boundary-matched threshold history, while their calibration slope is lower, their lift with demonstrations earlier and their weighing of a stated gain closer to the reference's. Reasoning responds to the history locally and does not convert that response into an estimate of what expansion would buy. When it finishes by itself, Qwen 3 8B moves probability on \cls{rule} histories toward patch and defer, with a gap between \cls{rule} histories and their boundary-matched threshold histories of 0.06 and 0.03 in the two orders, below the criterion of 0.10, while its probability of expanding on \cls{rule} histories does not follow the price and its chains explain the anomalies as exceptions to a single boundary. Given the gain, the three models value it after reasoning in the reference's proportions.

\end{document}